\documentclass[letterpaper, 10 pt, journal, twoside]{IEEEtran}
\IEEEoverridecommandlockouts % Provides \thanks commands
\let\labelindent\relax
\usepackage{hyperref, url, amsmath, amssymb, booktabs, graphicx, bm, color, algorithm, tabularx, enumitem, flushend}
\usepackage[noend]{algpseudocode}
\graphicspath{{images/}}

\newcommand{\new}[1]{#1}

\DeclareMathOperator*{\argmin}{\arg\min}
\newcolumntype{Y}{>{\centering\arraybackslash}X}

\title{\Large \bf Bodies Uncovered: Learning to Manipulate Real Blankets \\ Around People via Physics Simulations}

\author{Kavya Puthuveetil$^{1}$, Charles C. Kemp$^{1}$, and Zackory Erickson$^{2}$%
\thanks{Manuscript received: September, 9, 2021; Revised December, 19, 2021; Accepted January, 3, 2022.}
\thanks{This paper was recommended for publication by Editor Stephen J. Guy upon evaluation of the Associate Editor and Reviewers' comments.
This work was supported by NSF award IIS-2024444 and Google Cloud Credits for Research. Dr. Kemp owns equity in and works for Hello Robot, a company commercializing robotic assistance technologies.}
\thanks{$^{1}$Kavya Puthuveetil, and Charles C. Kemp are with the Healthcare Robotics Lab, Georgia Institute of Technology, Atlanta, GA, USA.
    {\tt\footnotesize kpputhuveetil@vcu.edu}}%
\thanks{$^{2}$Zackory Erickson is with the Robotics Institute, Carnegie Mellon University, Pittsburgh, PA, USA.
        {\tt\footnotesize zackory@cmu.edu}}%
\thanks{$^{3}$\url{https://github.com/RCHI-Lab/bodies-uncovered}}
\thanks{Digital Object Identifier (DOI): see top of this page.}
}
\begin{document}

\maketitle

\begin{abstract}
While robots present an opportunity to provide physical assistance to older adults and people with mobility impairments in bed, people frequently rest in bed with blankets that cover the majority of their body. To provide assistance for many daily self-care tasks, such as bathing, dressing, or ambulating, a caregiver must first uncover blankets from part of a person's body. In this work, we introduce a formulation for robotic bedding manipulation around people in which a robot uncovers a blanket from a target body part while ensuring the rest of the human body remains covered. \new{We compare two approaches for optimizing policies which provide a robot with grasp and release points that uncover a target part of the body: 1) reinforcement learning and 2) self-supervised learning with optimization to generate training data.} We trained and conducted evaluations of these policies in physics simulation environments that consist of a deformable cloth mesh covering a simulated human lying supine on a bed. In addition, we transfer simulation-trained policies to a real mobile manipulator and demonstrate that it can uncover a blanket from target body parts of a manikin lying in bed. Source code is available online$^{3}$.
\end{abstract}


% Although an essential task to consider in pursuing the development of robust robotic caretakers, autonomous robotic manipulation of bedding around a person in bed is still largely unexplored. In this work, we formulate a bedding manipulation task where the robot aims to select grasp and release points on a blanket over a person that uncover a given target body part without exposing non-target parts of the body. We trained policies in simulation to uncover six different target body parts given information about the person’s pose using both reinforcement learning and self-supervised learning approaches. For most target limbs, the resulting policies successfully choose reasonable grasp and release points to complete the task. In an evaluation of how well policies generalized to scenarios outside the training set, we observed significant degradation in precision when handling more complex target limb cases. Finally, we demonstrate promising transfer of simulation trained policies to a real-world mobile manipulator for manipulating a blanket over a manikin in bed.

% Anatomical positions in bed: https://www.thoughtco.com/anatomical-position-definitions-illustrations-4175376
% Supine, prone, right lateral recumbent, Fowler's

\section{Introduction}
\label{sec:intro}

	\IEEEPARstart{R}{obots} have the potential to improve the quality of life and independence of millions of individuals who need assistance with activities of daily living (ADLs). For many individuals with substantial mobility impairments, such as quadriplegics, physical assistance often occurs while they rest on a bed with blankets and sheets covering the majority of their body. For daily activities like hygiene (e.g. a bed bath), dressing, itch scratching, ambulation (e.g. limb repositioning or getting out of bed), and injury/wound care, a caregiver must first manipulate bedding to uncover part of a person's body. In the context of robotic assistance, bedding can prove to be a significant obstacle, not only because it can act as a visual occlusion but also because it can prevent access to the person's body. Although the most simplistic solution may be to remove a blanket in its entirety, a robotic caretaker should not indiscriminately remove blankets as human caregivers are often trained to minimize unnecessary exposure of the person they are assisting with a given task, bed bathing for example, to maintain privacy and comfort~\cite{lynn2011bedbath}.
    
	In this work, we introduce a method for robots to manipulate blankets over a person lying in bed. Specifically, we show how models trained entirely in simulation can enable a real robot to uncover a blanket from target body parts of a manikin person lying in bed (e.g. right forearm or both feet), while ensuring that the rest of the manikin's body remains covered. Finding solutions to this problem is non-trivial since selectively uncovering a target body part relies on a robot modeling the relationship between the blanket, whose state is constantly changing during manipulation, and a person's pose. Our approach considers solutions to be grasp and release locations over the bed that optimize for moving a blanket to uncover select human body parts while minimizing the amount that non-target body parts get uncovered.
    
\begin{figure}
\centering
\includegraphics[width=0.24\textwidth, trim={11.5cm 5.29cm 12.2cm 3cm}, clip]{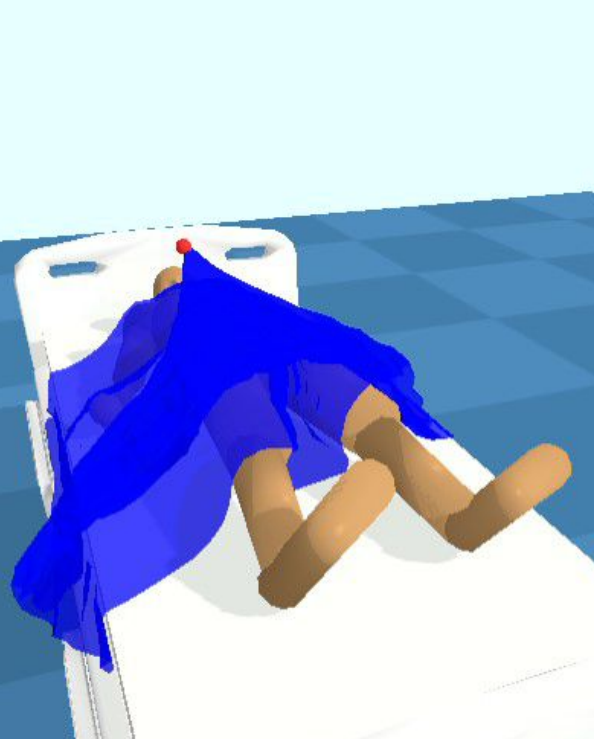}
\hspace{-0.25cm}
\includegraphics[width=0.24\textwidth, trim={8.5cm 0cm 9cm 1.5cm}, clip]{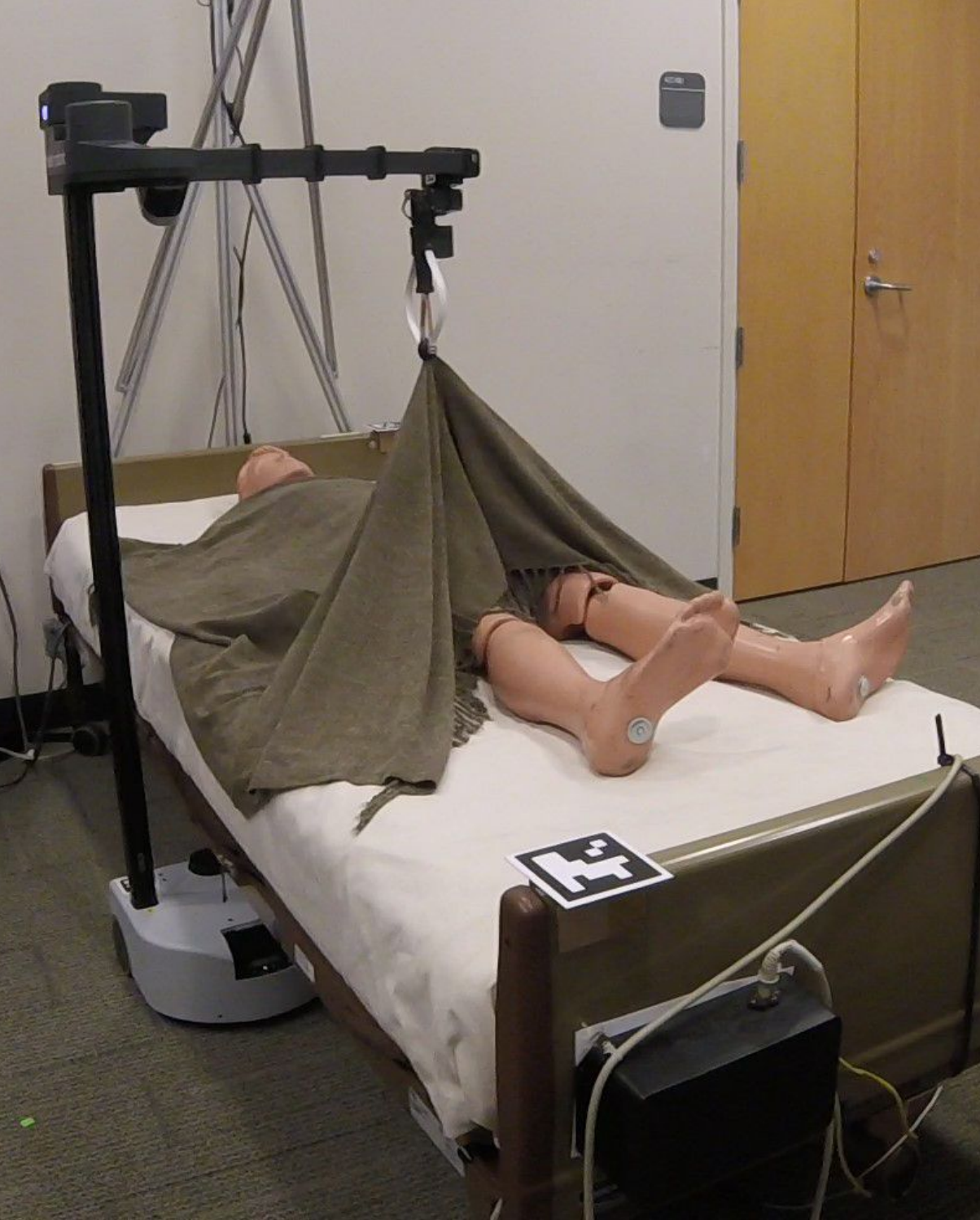}
\vspace{-0.2cm}
\caption{\label{fig:intro_image} A bedding manipulation policy trained in simulation that we have transferred to the real world for manipulating blankets over a person. (Left) The physics simulation environment used to train models for manipulating deformable blankets over a person. (Right) A real mobile manipulator (Stretch RE1) executing the simulation-trained policy to grasp a blanket and uncover the legs of a medical manikin.}
\vspace{-0.48cm}
\end{figure}

    \new{The large distribution of human body shapes and resting poses present significant challenges to developing strategies for targeted bedding manipulation around people. To overcome this, we use simulation and data-driven learning to inform bedding manipulation.}
    
    \new{We present two approaches for bedding manipulation around people: 1) reinforcement learning and 2) self-supervised learning with gradient-free optimization to generate training data without human intervention.} For the reinforcement learning formulation, the manipulator is rewarded for uncovering target body parts and is penalized for uncovering non-target body parts or for covering a person's head. For the self-supervised learning formulation, we used covariance matrix adaptation evolution strategy (CMA-ES) to optimize for bedding manipulation solutions that achieve high reward for a fixed environment setup and train neural network models that map state observations to optimized manipulation trajectories.
    
    We first evaluate trained bedding manipulation models in simulation with deformable cloth covering simulated humans. We compare performance between states drawn from the training distribution of human and blanket poses and states drawn from a more varied distribution. We then demonstrate these simulation-trained models on a real mobile manipulator as it uncovers a blanket over a medical manikin lying in a bed, as shown in Fig.~\ref{fig:intro_image}. For several randomized human poses in bed, the trained models achieved strong performance in uncovering target body parts while ensuring the rest of the human body remains covered, both in simulation and in the real world. These results provide a promising signal towards the goal of bedding manipulation around people, a valuable skill for robotic caregivers that wish to assist people in bed.

Through this work, we make the following contributions:
\begin{itemize}
\item We introduce a formulation for autonomous bedding manipulation around people lying supine in bed. Our approach seeks to uncover target body parts while ensuring the rest of the body remains covered.
\item We present, evaluate, and compare both reinforcement learning and \new{self-supervised} learning approaches to bedding manipulation in simulation.
\item We transfer simulation-trained bedding manipulation models to a real-world mobile manipulator and demonstrate this approach for manipulating a blanket over a manikin in bed.
\end{itemize}

\section{Related Work}
\label{sec:related_work}

\subsection{Bedding and Cloth Manipulation}
Work investigating robotic cloth manipulation has considered tasks that involve smoothing, folding, or unfolding cloth using a variety of task-oriented and robot learning-based approaches~\cite{tanaka2018emd, ganapathi2021learning, tsurumine2019drl}. Task-oriented approaches generally use traditional perception-based algorithms to identify manually selected cloth features, like hems, corners, or wrinkles~\cite{sun2015wrinkles, ramisa2012wrinkles, yamazaki2014hems, seita2019bedmaking, yuba2017stateest, qian2020segmentation, shepard2010towel}. These features are used to build an estimate of the cloth state that can inform cloth manipulation, often the selection of grasp points on the cloth. 

Robot learning and model-based approaches instead learn the relationship between the shape of the cloth and the task specific manipulation. For example, Matsubara et al.~\cite{matsubara2013topology} proposed a reinforcement learning framework that allows a dual-arm robot to learn motor skills for a self-dressing task based on the topological relationship between the robot and the cloth. Hoque et al.~\cite{hoque2021visuospatial} introduce VisuoSpatial Foresight, which learns visual dynamics from RGB-D images in simulation instead of performing state estimation for a multi-step fabric smoothing task. Garcia-Camacho et al.~\cite{garcia2020benchmarking} has benchmarked bimanual cloth manipulation for pulling a tablecloth onto a table, which shares some similarities to the problem of manipulating blankets on a bed.

While learning-based cloth manipulation approaches in the real world have yielded promising results for a variety of tasks, the potential for generalization is limited due to the excessive cost associated with collecting a training data set of sufficiently large size~\cite{ganapathi2021learning, ebert2018visual}. Several recent methods have leveraged simulation as a way to learn general manipulation skills of cloth, such as folding, yet do not consider the manipulation of cloth around people~\cite{lin2020softgym, matas2018sim, wu2020learning}.

With respect to robotic manipulation of bedding, existing literature is currently sparse and does not explore the use of physics simulation~\cite{jimenez2020perception}. Seita et al.~\cite{seita2019bedmaking}, has introduced a method for autonomous bed-making that leverages depth images to find and pull the corners of a blanket to corresponding corners of a bed. The authors tested their trained models on both HSR and Fetch robots for covering the surface of a bed with a blanket. While prior work has explored making an empty bed, in this paper we explore manipulating blankets to uncover a human body in bed.
% Similarly, the authors introduce a self-supervised learning approach for selecting grasp points along a blanket by training on human pick point data~\cite{seita2019bedmaking}. The robot decides whether to attempt multiple grasps and pulls the blanket based on a bed coverage heuristic. 

Despite the frequent need for uncovering blankets from a person in bed before providing physical assistance, autonomous robotic assistance for this task remains largely unexplored. 

\subsection{Robotic Assistance}

Beyond the need for uncovering the human body in bed to provide caregiving, robots have also manipulated cloth as a means to directly provide assistance.

One application of cloth manipulation within caregiving has been robot-assisted dressing~\cite{koganti2015cloth, twardon2018learning, yamazaki2014bottom, canal2018joining, pignat2017learning, gao2016iterative}. Recent work has considered how haptic~\cite{erickson2018hapticpredictive, zhang2017personalized, kapusta2016data, yu2017haptic} or capacitive~\cite{erickson2019multidimensional, kapusta2019personalized} sensing techniques can help estimate human pose and overcome some of the visual occlusions that occur when dressing various garments. While much of the prior work in robot-assisted dressing focuses on methods to accomplish a dressing task given a fixed grasp position on a garment, Zhang et al.~\cite{zhang2020learningGP} and Saxena et al.~\cite{saxena2019garment} introduce approaches for identifying robot grasp points along garments for dressing assistance. 

When providing care to an individual who is in a bed, a common self-care task is bathing. Autonomous bed bathing by an assistive mobile manipulator was demonstrated by King et al.~\cite{king2010bedbath}. In their work, the robot executed a wiping behavior sequence over an operator-selected area in order to clean debris off of a person lying in bed. Erickson et al.~\cite{erickson2019multidimensional} presented a capacitive sensing technique for traversing a robot's end effector along the contours of a human limb and demonstrated this technique for performing an assistive bathing task with a wet washcloth. Kapusta et al.~\cite{kapusta2019system} introduced a teleoperation interface for a mobile manipulator that enabled an individual with severe quadriplegia to manually control the robot to help with cleaning his face and pulling a blanket down from the individual's knees to his feet.

\begin{figure}
\centering
\includegraphics[width=0.096\textwidth, trim={0.5cm 0.5cm 0.4cm 0cm}, clip]{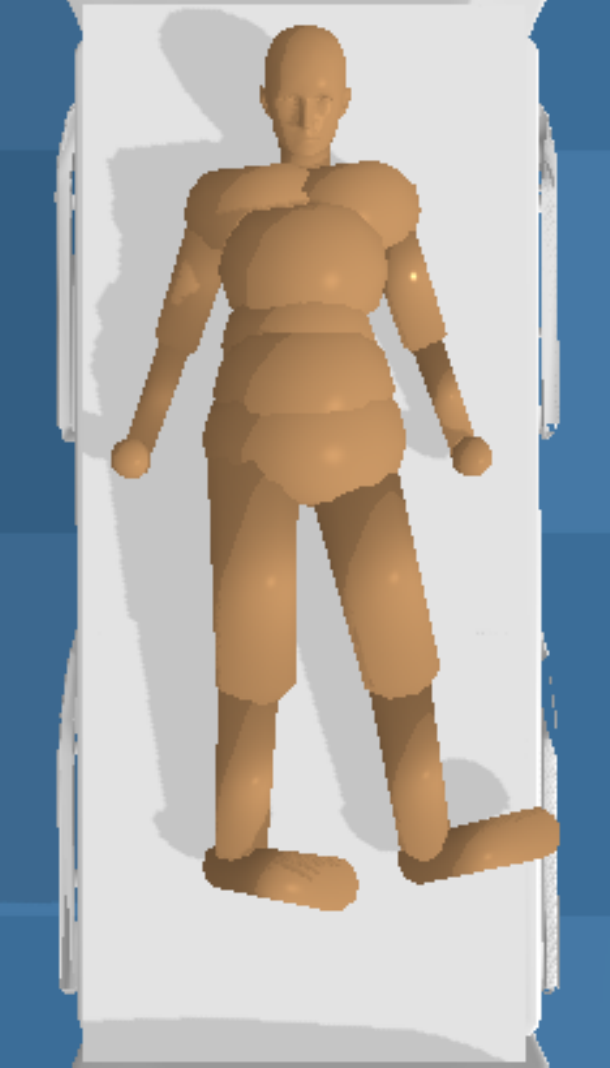} \hspace{-0.2cm}
 \includegraphics[width=0.096\textwidth, trim={0.5cm 0.5cm 0.4cm 0cm}, clip]{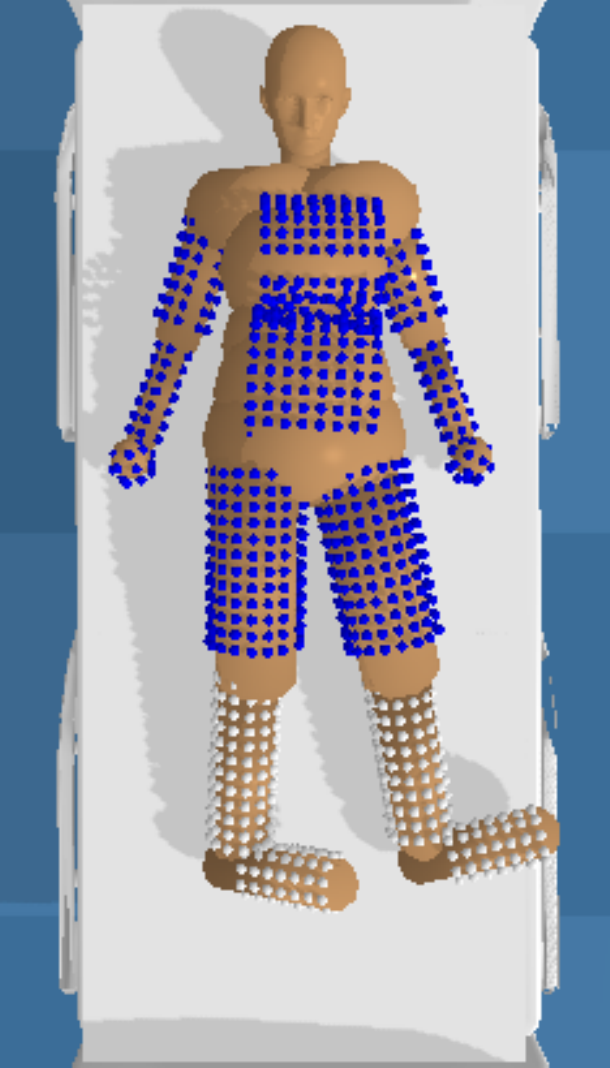} \hspace{-0.2cm}
 \includegraphics[width=0.096\textwidth, trim={0.5cm 0.5cm 0.4cm 0cm}, clip]{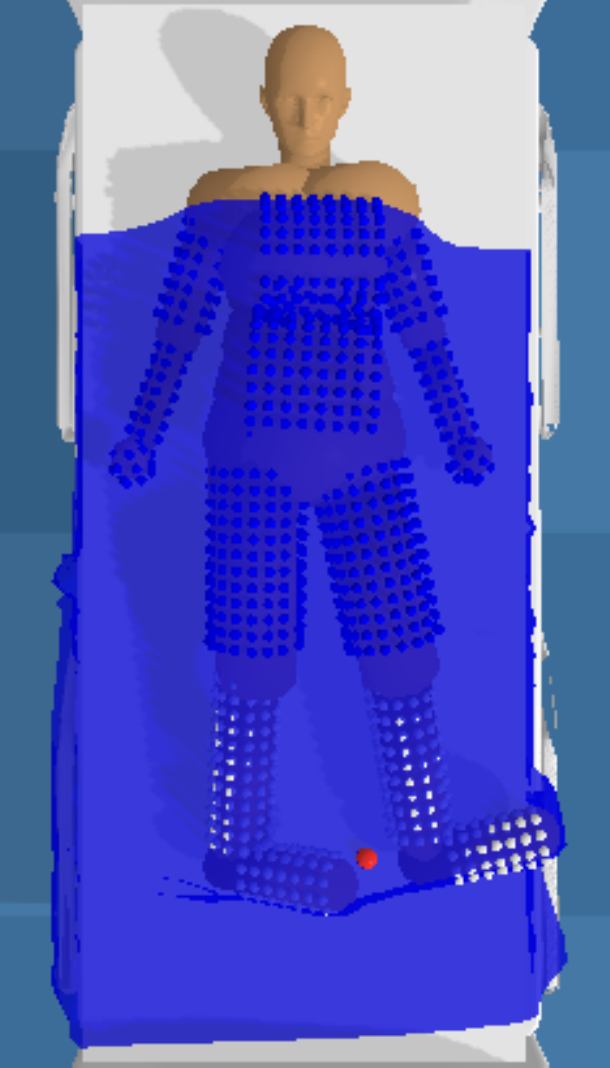} \hspace{-0.2cm}
 \includegraphics[width=0.096\textwidth, trim={0.5cm 0.5cm 0.4cm 0cm}, clip]{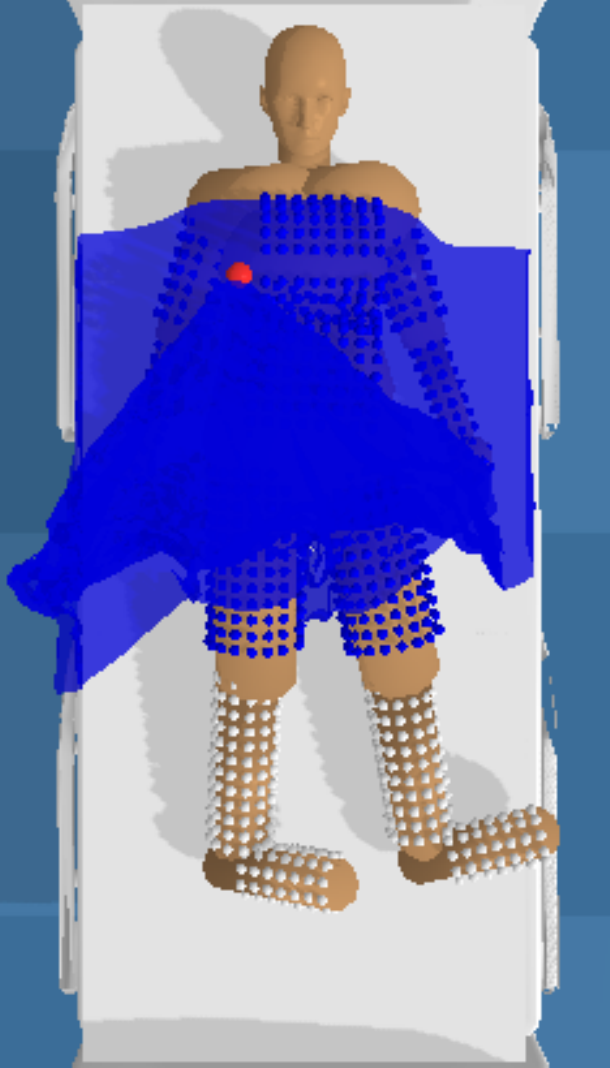} \hspace{-0.2cm}
 \includegraphics[width=0.096\textwidth, trim={0.5cm 0.5cm 0.4cm 0cm}, clip]{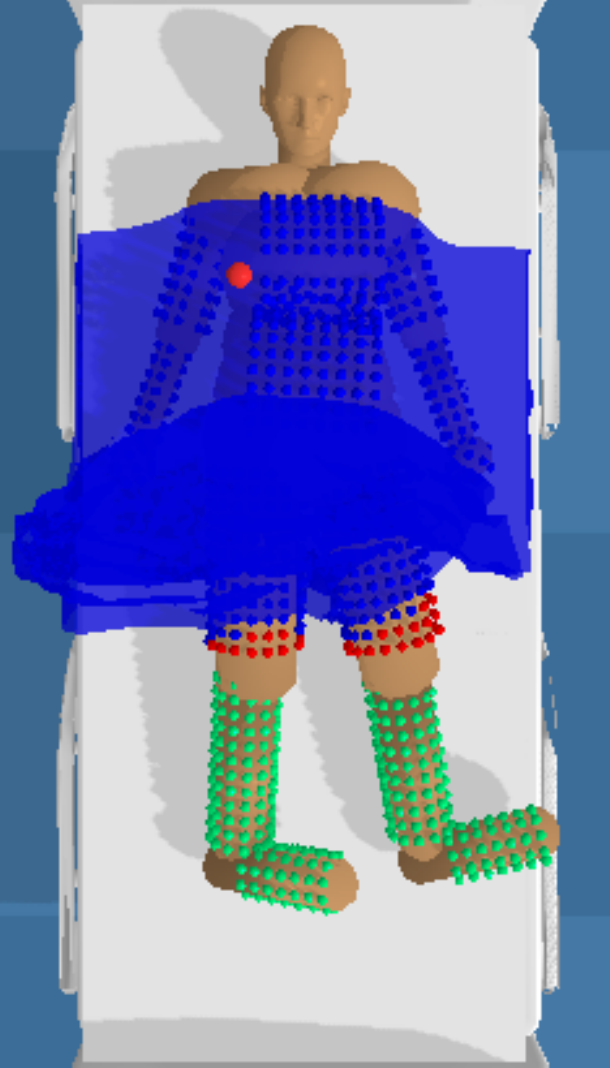} \\
 \vspace{0.05cm}
 \includegraphics[width=0.096\textwidth, trim={0.6cm 2cm 0.6cm 2.9cm}, clip]{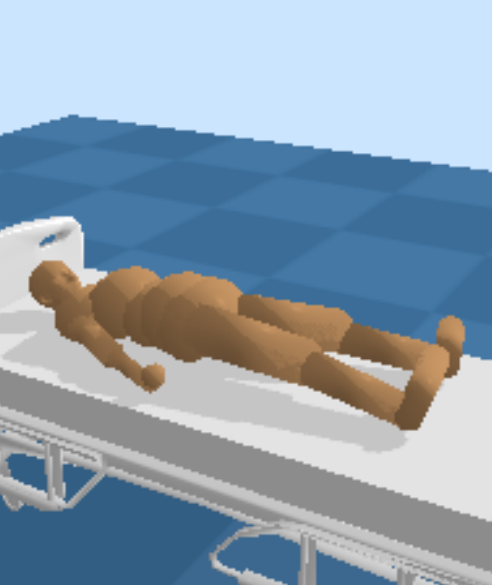} \hspace{-0.2cm}
 \includegraphics[width=0.096\textwidth, trim={0.6cm 2cm 0.6cm 2.9cm}, clip]{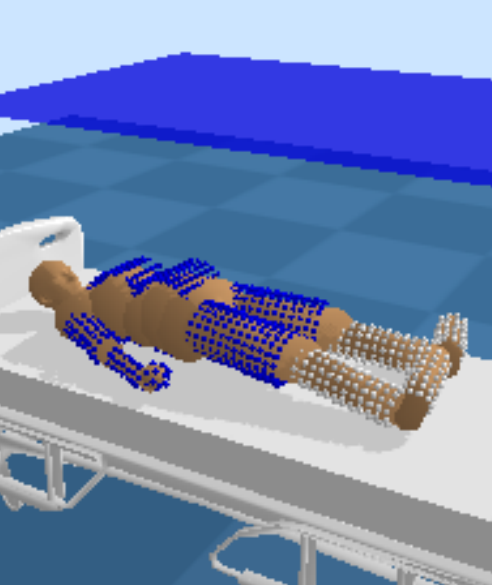} \hspace{-0.2cm}
 \includegraphics[width=0.096\textwidth, trim={0.6cm 2cm 0.6cm 2.9cm}, clip]{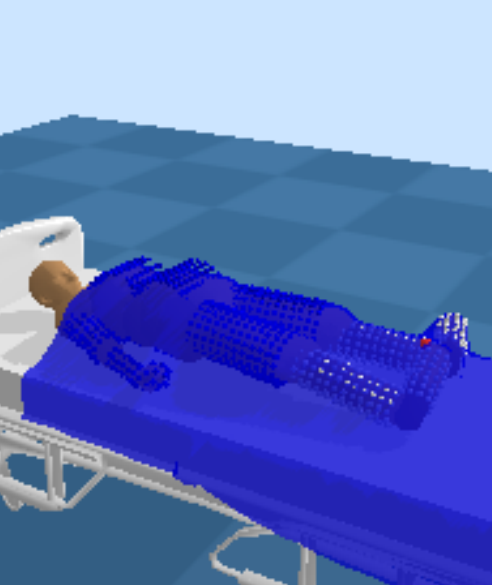} \hspace{-0.2cm}
 \includegraphics[width=0.096\textwidth, trim={0.6cm 2cm 0.6cm 2.9cm}, clip]{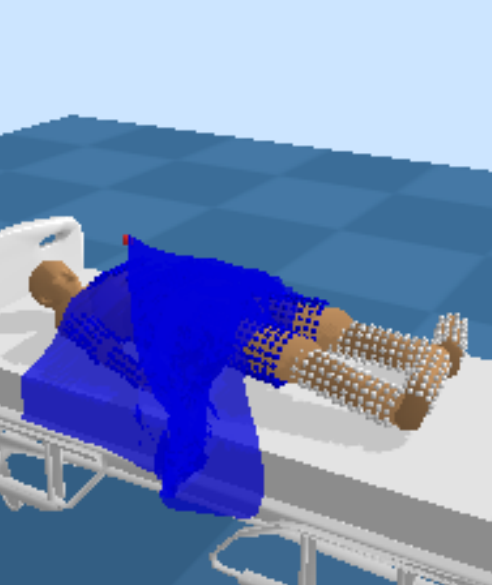} \hspace{-0.2cm}
 \includegraphics[width=0.096\textwidth, trim={0.6cm 2cm 0.6cm 2.9cm}, clip]{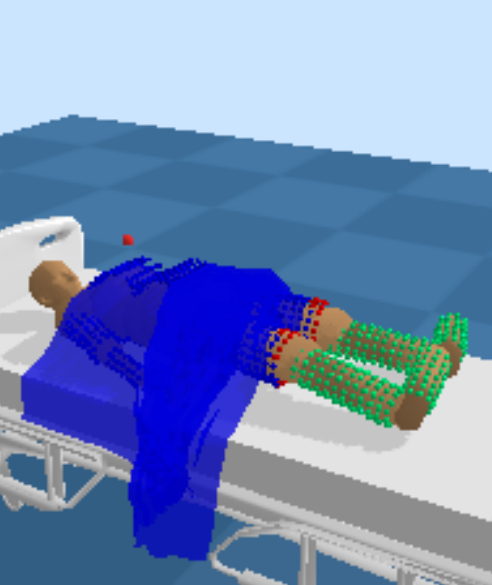}\\
\caption{\label{fig:simenv} Environment setup for the bedding manipulation task in simulation. 1) We allow a human model to settle on the bed, 2) we discretize the body into target and non-target points, 3) we drop the blanket onto the person and the spherical manipulator grasps the blanket at a grasp location, 4) the blanket is raised to 40cm above the bed, the spherical manipulator moves to the release location 5) the manipulator drops the blanket and we assess whether the points on the body are correctly (green) or incorrectly (red) uncovered. Non-target points on the head are not visualized.}
\vspace{-0.5cm}
% \vspace{0.2cm}
\end{figure}

\section{Bedding Manipulation Around People}
\label{sec:methods}

In this section, we describe our formulation and implementation of the bedding manipulation task in simulation using Assistive Gym~\cite{erickson2020assistivegym}. We establish the action space, observation features, and reward function employed in the bedding manipulation task and detail reinforcement learning and self-supervised learning approaches for task completion. Lastly, we describe implementation of simulation trained policies in the real world using the Stretch RE1 mobile manipulator from Hello Robot.

\subsection{Simulation Environment}
\label{sec:sim_env}
    In order to train models for bedding manipulation around people, we begin by implementing a deformable bedding manipulation environment in Assistive Gym~\cite{erickson2020assistivegym}, a physics-based simulation framework for physically assistive robotics. Simulation of planar deformables like cloth is implemented in PyBullet as a mass-spring deformable system.
    
	We represent a human in this environment using a configurable capsulized female model, shown in Fig.~\ref{fig:simenv}, with fixed body size based on 50th percentile values as offered by Assistive Gym~\cite{erickson2020assistivegym}. The model is initially positioned supine above a hospital bed with dimensions of 0.88m x 2.1m with arms and legs spread from its sides by 20 degrees and 10 degrees respectively. Variation uniformly sampled between \(\pm\) 0.2 radians is introduced to the model’s joint positions before dropping the model from 1.5m above the ground onto the bed. The model is allowed to reach a static resting pose such that the velocity of all joints is under 0.01 m/s.

	The simulated blanket is represented as a deformable grid mesh with 2089 vertices and dimensions of 1.25m $\times$ 1.7m. The cloth parameters, including mass, friction, spring damping, and spring elasticity, were manually tuned to result in cloth dynamics similar to a real-world blanket. Once the human model has settled on the bed, the blanket is positioned above the person and dropped from a height of 1.5m such that the entire human body is covered except the head and neck. Interactions with the blanket are performed by a sphere manipulator representing the robot’s end effector.
    We discretize the outer surface of the human body into 1775 points equally spaced 3~cm apart from each other. Let $\bm{\chi}$ be this set of discrete body points. We project these points onto a 2D plane along the surface of the bed, defined as $\bm{\chi'} = \bm{\chi} \odot (1, 1, 0)$, where $\odot$ represents the Hadamard product. We also project the set of cloth mesh points, $\bm{V}$, onto the same 2D plane via $\bm{V'} = \bm{V} \odot (1, 1, 0)$. We define a function $\bm{C}(\bm{x'}, \bm{V'})$ to classify whether a given projected body point $\bm{x'} \in \bm{\chi'}$ is covered or uncovered by the cloth based on whether the Euclidean distance between $\bm{x'}$ and any of the projected cloth points is within a threshold distance $\lambda = 2.8 \text{cm}$. $\bm{C}(\bm{x'}, \bm{V'})$ is expressed as:
% \[\bm{C}(\bm{\chi'_{*}}, \bm{V}) = \begin{cases} 1 & \text{if }\exists\bm{v} \in \bm{V'} :  ||\bm{\chi'}_{*} - \bm{v}||_2 < \lambda\\0 & \text{otherwise}\\\end{cases}\] 
\begin{equation}
\label{eq:C}
\bm{C}(\bm{x'}, \bm{V'}) = \begin{cases} 1 & \text{if }\exists\bm{v'} \in \bm{V'} :  ||\bm{x'} - \bm{v'}||_2 < \lambda\\0 & \text{otherwise}\\\end{cases}
\end{equation}

    We compute $\bm{C}(\bm{x'}, \bm{V'})$ for all projected body points in  $\bm{\chi'}$. A point is considered covered if $\bm{C}(\bm{x'}, \bm{V'})=1$ and is considered uncovered if $\bm{C}(\bm{x',\bm{V'})}=0$. The threshold distance $\lambda$ was determined via manual tuning to minimize false identification of the state of projected body points.
    
% \subsection{Policy Learning and Control \ck{This section is really big - i recommend breaking it up - for example you could add subsections}}
% \label{sec:policy_learning}

\subsection{Actions and Observations}
   We define an action space $A$ for the bedding manipulation task wherein the robot moves linearly between a grasp and release point for the cloth blanket. Grasp $\bm{a}_g = (a_{g,x}, a_{g,y})$ and release $\bm{a}_r = (a_{r,x}, a_{r,y})$ points are represented as 2D points along a plane parallel to the surface of the bed. These points are sampled from a bounded region along the bed, i.e. $a_{g,x},a_{r,x} \in$ (-0.44~m, 0.44~m) and $a_{g,y},a_{r,y} \in $ (-1.05~m, 1.05~m). Actions for the robot are defined as $\bm{a} = (a_{g,x}, a_{g,y}, a_{r,x}, a_{r,y}) \in A$. Given a grasp and release point, the robot grasps the closest, measured by Euclidean distance, cloth mesh vertex $\bm{v'} \in \bm{V'}$ to the defined grasp point, lifts the cloth upwards above the bed, and follows a linear Cartesian trajectory to the release point after which the cloth is released and falls downwards onto the person.

\begin{figure}
\centering
 \includegraphics[width=0.161\textwidth, trim={0cm 1cm 0cm 0cm}, clip]{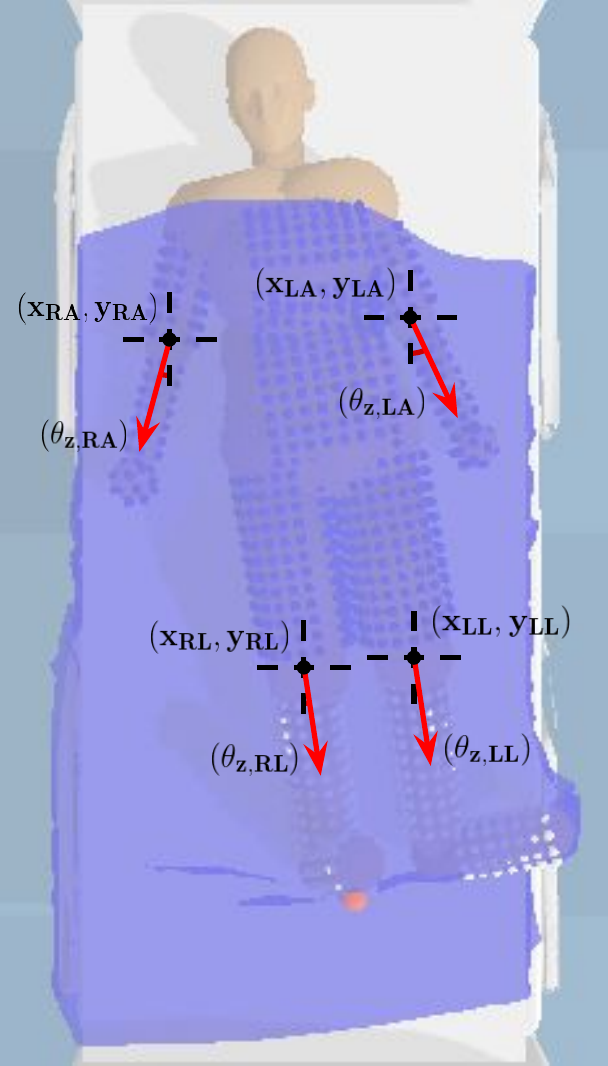}
 \includegraphics[width=0.13\textwidth, trim={1cm 2.5cm 2.5cm 1.4cm}, clip]{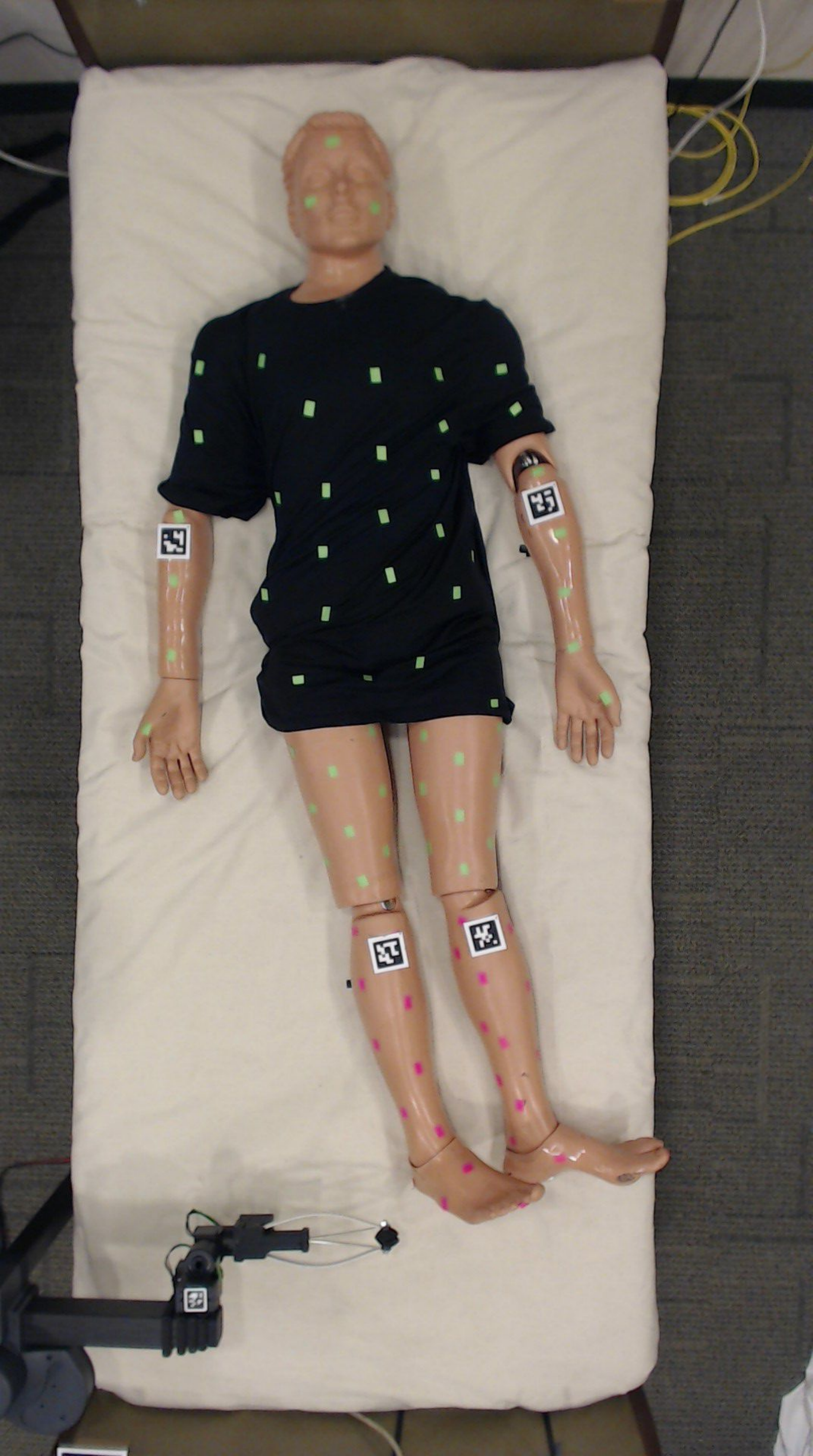}
\vspace{-0.2cm}
\caption{\label{fig:obs} The observation is defined as the 2D position of the elbows and knees, as well as the yaw orientation of the forearms and knees. We capture these observations using fiducial tags in the real world.}
\vspace{-0.5cm}
\end{figure}

	At the beginning of a trial, the robot captures an observation, $\bm{s} = (\bm{s}_{RL}, \bm{s}_{LL}, \bm{s}_{RA}, \bm{s}_{LA})$, of the human pose in bed, where $\bm{s}_{\bullet} = (\bm{s}_{\bullet, x}, \bm{s}_{\bullet, y}, \bm{s}_{\bullet, \theta_{z}})$. As shown in Fig.~\ref{fig:obs}, the observation represents the pose of the human's right leg $\bm{s}_{RL}$, left leg $\bm{s}_{LL}$, right arm $\bm{s}_{RA}$, and left arm $\bm{s}_{LA}$ using the 2D position of elbow and knee joints, as well as the yaw orientation of the forearms and shins.
    
    This observation $\bm{s}$ is then provided to a policy $\pi(\bm{s})$, which outputs a grasp and release point $\bm{a}$ for uncovering a target body part. We start with the assumption that the blanket is dropped on a person from the same starting position and orientation, and hence we do not provide blanket state information as part of the observation. This has the added benefit of simplifying the observation, which reduces the training time needed to learn policies for targeted bedding manipulation above people. In Section~\ref{sec:generalize}, we break this assumption and evaluate how well our learned policies generalize to varying blanket states above a person.
    
    In order to manipulate the cloth in simulation when given an action $\bm{a} = (a_{g,x}, a_{g,y}, a_{r,x}, a_{r,y})$, we begin by anchoring the spherical manipulator in simulation to the cloth vertex $\bm{v}^* \in \bm{V'}$ that is nearest to the grasp point $\bm{a}_g$ in the bed plane, $\bm{v}^* = \argmin\limits_{\bm{v'}\in\bm{V'}} ||\bm{v'} - \bm{a}_g||_2$. Once anchored, the manipulator translates upwards along the $Z-axis$ 40cm above the bed, lifting the cloth off of the person, and then follows a linear Cartesian trajectory from $\bm{a}_g$ to $\bm{a}_r$. \new{We select this 40cm fixed lift distance to ensure the robot's end effector does not collide with the human body. This height also corresponds to the maximum lift height of the mobile manipulator used in our real-world evaluation, described in Section~\ref{sec:real_world_setup}}. Once the spherical manipulator reaches the release point, the anchor is released, the cloth drops back down onto the person, and a reward value is computed. This process is depicted in Fig.~\ref{fig:simenv}.

\subsection{Reward Definition}
In order to compute the reward value used to train policies, we first define a reward function $R(\bm{S}, \bm{a})$ for bedding manipulation around people given the global state of the simulation environment $\bm{S}$. Relevant elements from $\bm{S}$ include the set of target points along the body to uncover $\bm{P_{t}}$ (shown as white points in Fig.~\ref{fig:simenv}), the set of non-target points that should remain covered $\bm{P_{n}}$ (shown as blue points in Fig.~\ref{fig:simenv}), and the set of head points $\bm{P_{h}}$, where $\bm{P_{t}}, \bm{P_{n}}, \bm{P_{h}} \subseteq \bm{\chi}$. We can similarly project these points onto the bed plane such that $\bm{P'_{t}}, \bm{P'_{n}}, \bm{P'_{h}} \subseteq \bm{\chi'}$. Once cloth manipulation is completed, we also have the following elements of $\bm{S}$:
\begin{itemize}
\item $\rho_{t} = |\bm{P'_{t}}| - \sum\limits_{\bm{p'}\in \bm{P'_{t}}} \bm{C}(\bm{p'}, \bm{V'}) $
\item $\rho_{n} = |\bm{P'_{n}}| - \sum\limits_{\bm{p'}\in \bm{P'_{n}}} \bm{C}(\bm{p'}, \bm{V'}) $
\item $\rho_{h} = \sum\limits_{\bm{p'}\in \bm{P'_{h}}} \bm{C}(\bm{p'}, \bm{V'}) $.
\end{itemize}
$\rho_{t}$, $\rho_{n}$, $\rho_{h}$ describe the number of target points uncovered, the number of non-target points uncovered, and the number of head points covered, respectively. $\bm{C}(p', \bm{V'})$ is defined according to (\ref{eq:C}) and $|\bm{P'_{t}}|$ is the cardinality of set $\bm{P'_{t}}$.

We define a reward function $R(\bm{S}, \bm{a}) = R_{t}(\bm{S}) + R_{n}(\bm{S}) + R_{h}(\bm{S}) + R_{d}(\bm{S}, \bm{a})$  used when training robot policies where:
\begin{itemize}
\item \(R_{t}(\bm{S}) = 100(\rho_{t}/|\bm{P_{t}}|)\):  uncover target reward
\item \(R_{n}(\bm{S}) = -100\:(\rho_{n}/|\bm{P_{n}}|)\): uncover  non-target penalty.
\item \(R_{h}(\bm{S}) = -200\:(\rho_{h}/|\bm{P_{h}}|)\): cover head penalty.
\item \(R_{d}(\bm{S}, \bm{a}) = \begin{cases} -150 & ||\bm{a}_r - \bm{a}_g||_2 \geq  1.5m\\0 & otherwise\\\end{cases}\): penalty for large distance between the grasp and release points.
% \item \(R_{d}(\bm{S}) = \begin{cases} -150 & d \geq  1.5m\\0 & d < 1.5m\\\end{cases}, \: d= ||\bm{a}_r - \bm{a}_g||_2\): penalty for excessive distance between the grasp and release points
\end{itemize}

We scale both \(R_{t}(\bm{S})\) and \(R_{n}(\bm{S})\) to ensure that \(R(\bm{S}, \bm{a}) \leq 100\). This results in rewards with magnitude often corresponding to task completion in percentage, where the highest achievable reward is \(R(\bm{S}, \bm{a}) = 100\). We place a larger penalty on covering a person's face with the blanket \(R_{h}(\bm{S})\) as this outcome is unfavorable to people. The final reward term, \(R_{d}(\bm{S}, \bm{a})\), is a penalty applied when the Euclidean distance between grasp and release points is greater than 1.5m, which discourages policies from selecting actions that traverse from one corner of the bed to the opposite corner. 
% We apply a fixed penalty when the distance exceeds a threshold rather than scaling the penalty based on distance since the goal is not to minimize distance between grasp and release points.

\subsection{Policy Learning}
	We introduce two formulations for learning policies that enable a robot to manipulate bedding around people, one of which uses reinforcement learning and the other self-supervised learning. \new{Due to the high degree of variation across people and deformable cloth, we leverage learning in simulation to more efficiently model the wide distribution of cloth dynamics and human-cloth states.}
	
	We train policies for uncovering six body parts including the right lower leg (shin and foot), left arm, both lower legs, upper body, lower body, and whole body. Both formulations receive a single observation of the human at the start of a simulation rollout, execute a single action to manipulate the blanket (traversing a linear trajectory from $\bm{a}_g$ to $\bm{a}_r$), and receive a single reward at the terminal state once the simulation is complete.
    
	First, we use proximal policy optimization (PPO)~\cite{schulman2017proximal}, a deep reinforcement learning technique, to learn policies represented by a fully connected neural network with two hidden layers of 50 nodes and tanh activations. We train six policies, one for each target limb, with random variation in human pose (discussed in Section~\ref{sec:sim_env}) to facilitate better transfer of the learned policies to the real world. Each policy is trained for 5,000 simulation rollouts using 32 concurrent simulation actors with a 32 vCPU machine. We use data from each batch of 32 simulation rollouts to perform 50 SGD updates to the policy with a learning rate of 5e-5. Training time for each PPO policy took roughly 26 hours.
    
\begin{figure*}
\centering
 \includegraphics[width=0.16\textwidth, trim={9cm 0cm 8cm 1cm}, clip]{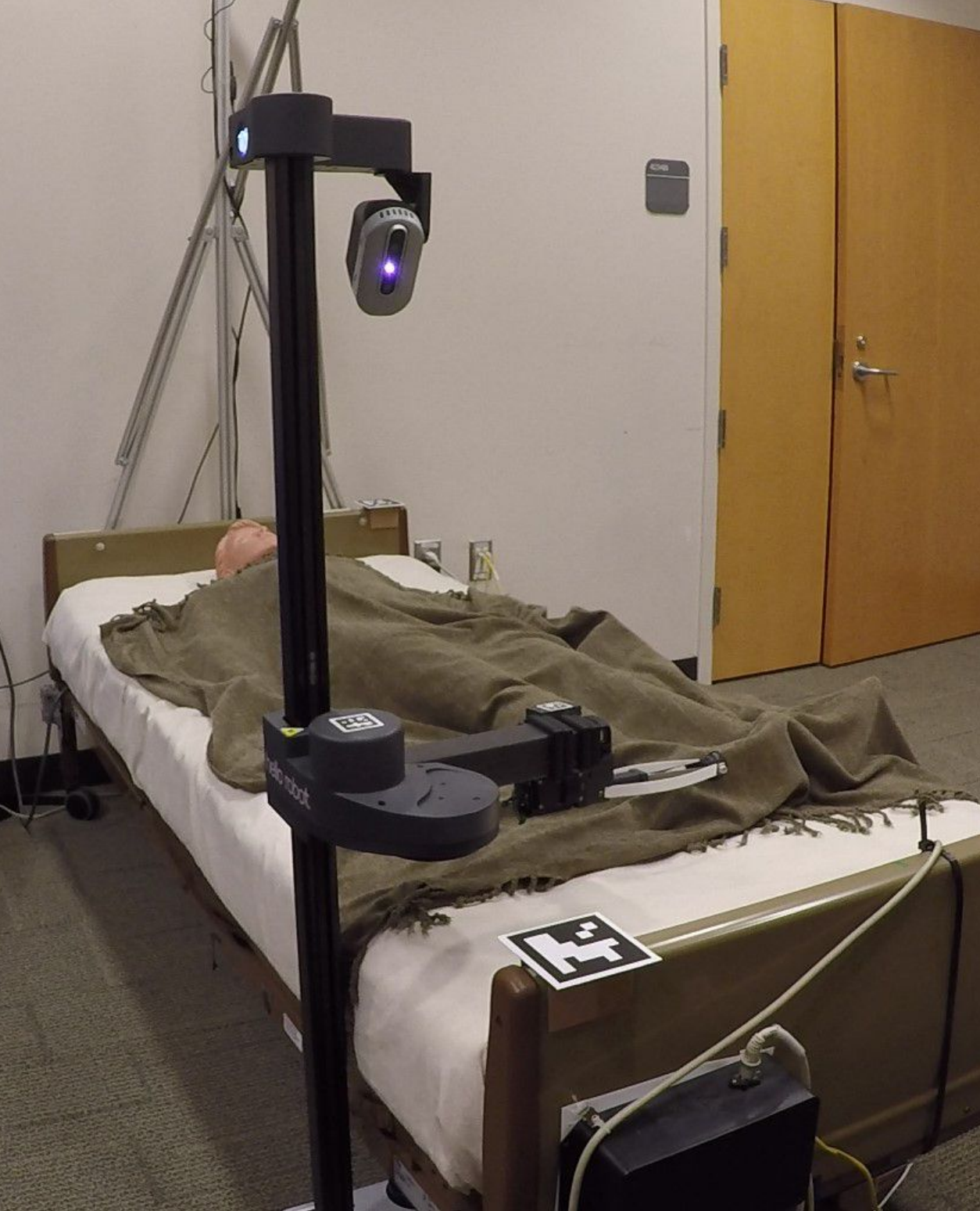}
 \includegraphics[width=0.16\textwidth, trim={9cm 0cm 8cm 1cm}, clip]{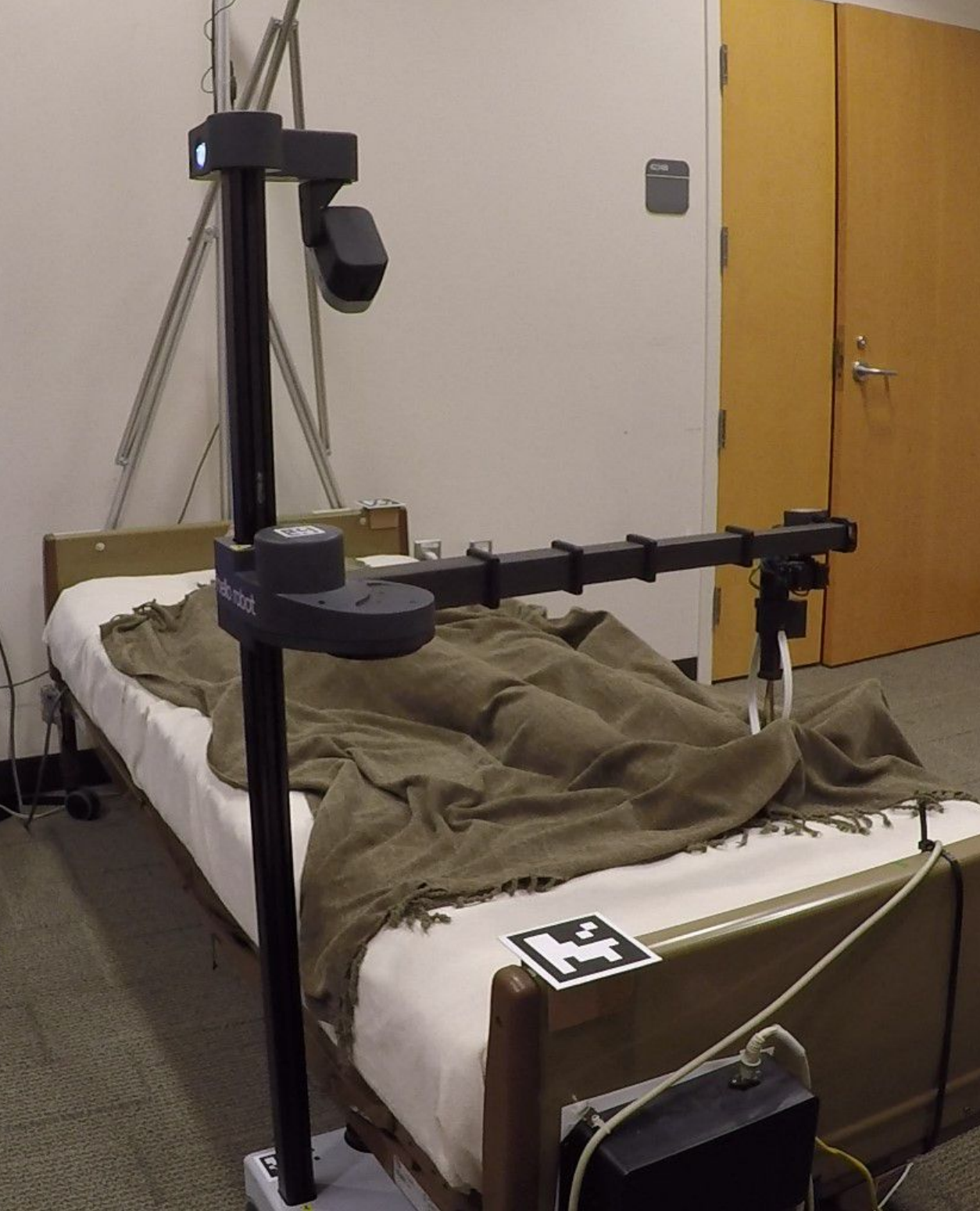}
 \includegraphics[width=0.16\textwidth, trim={9cm 0cm 8cm 1cm}, clip]{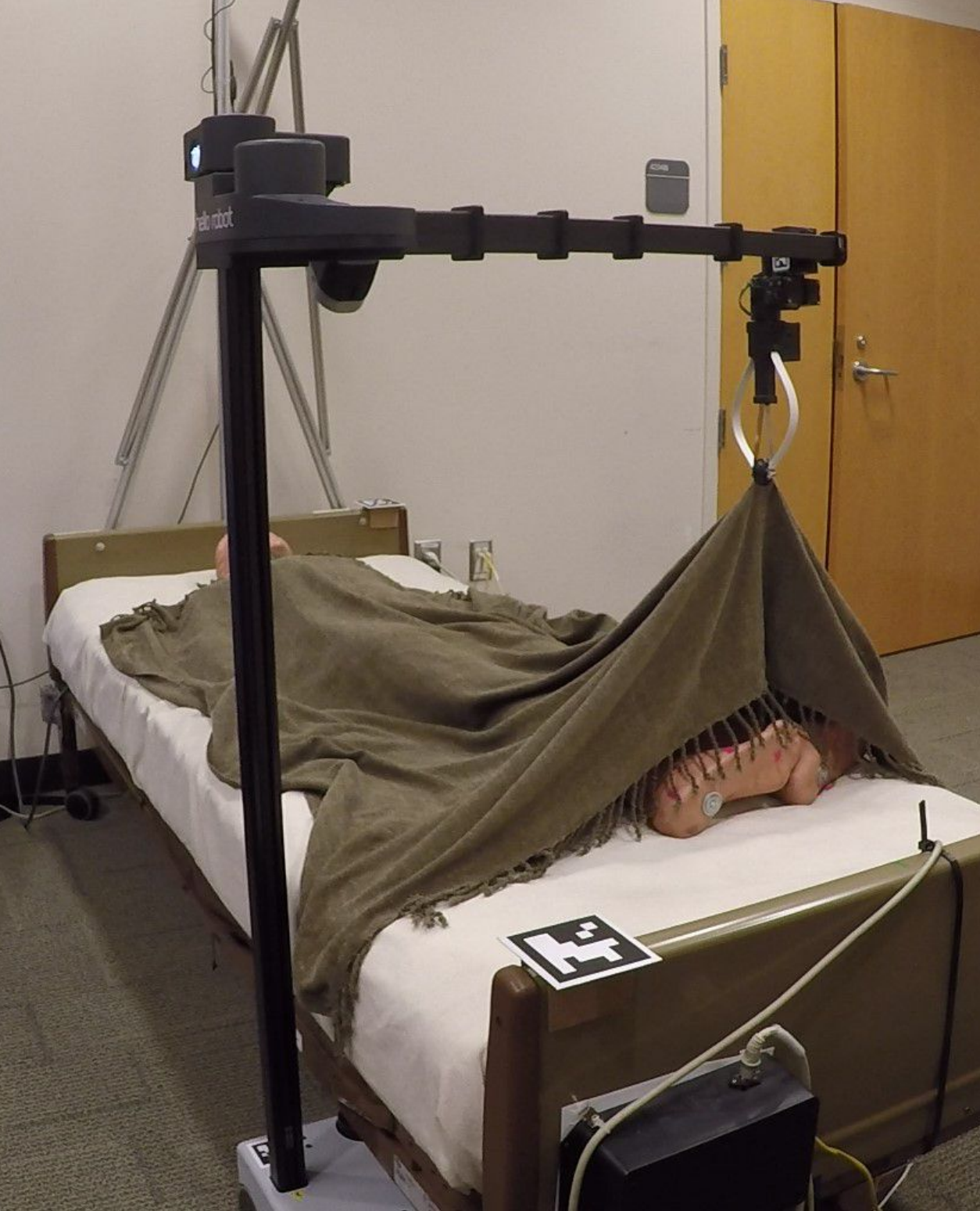}
 \includegraphics[width=0.16\textwidth, trim={9cm 0cm 8cm 1cm}, clip]{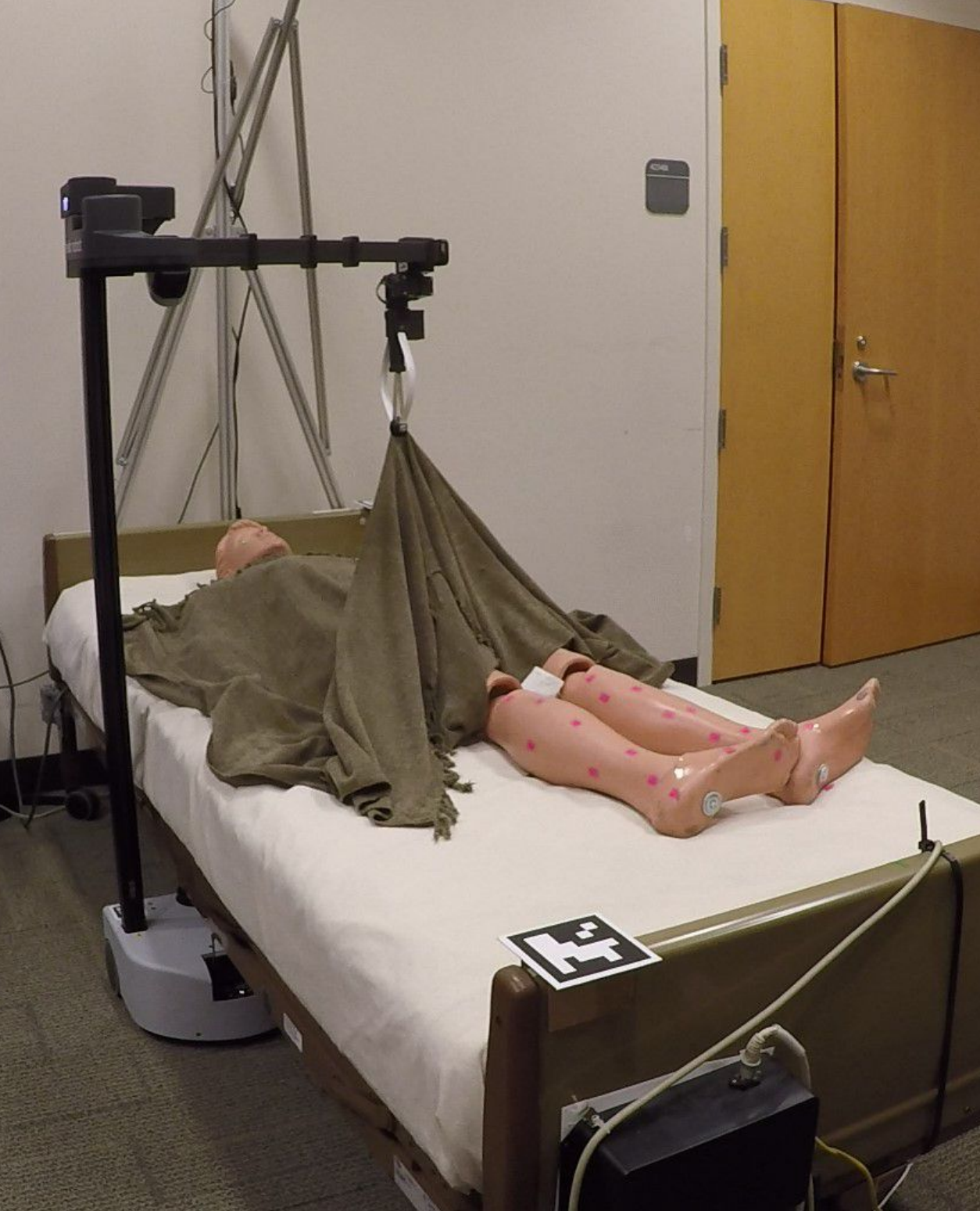}
 \includegraphics[width=0.16\textwidth, trim={9cm 0cm 8cm 1cm}, clip]{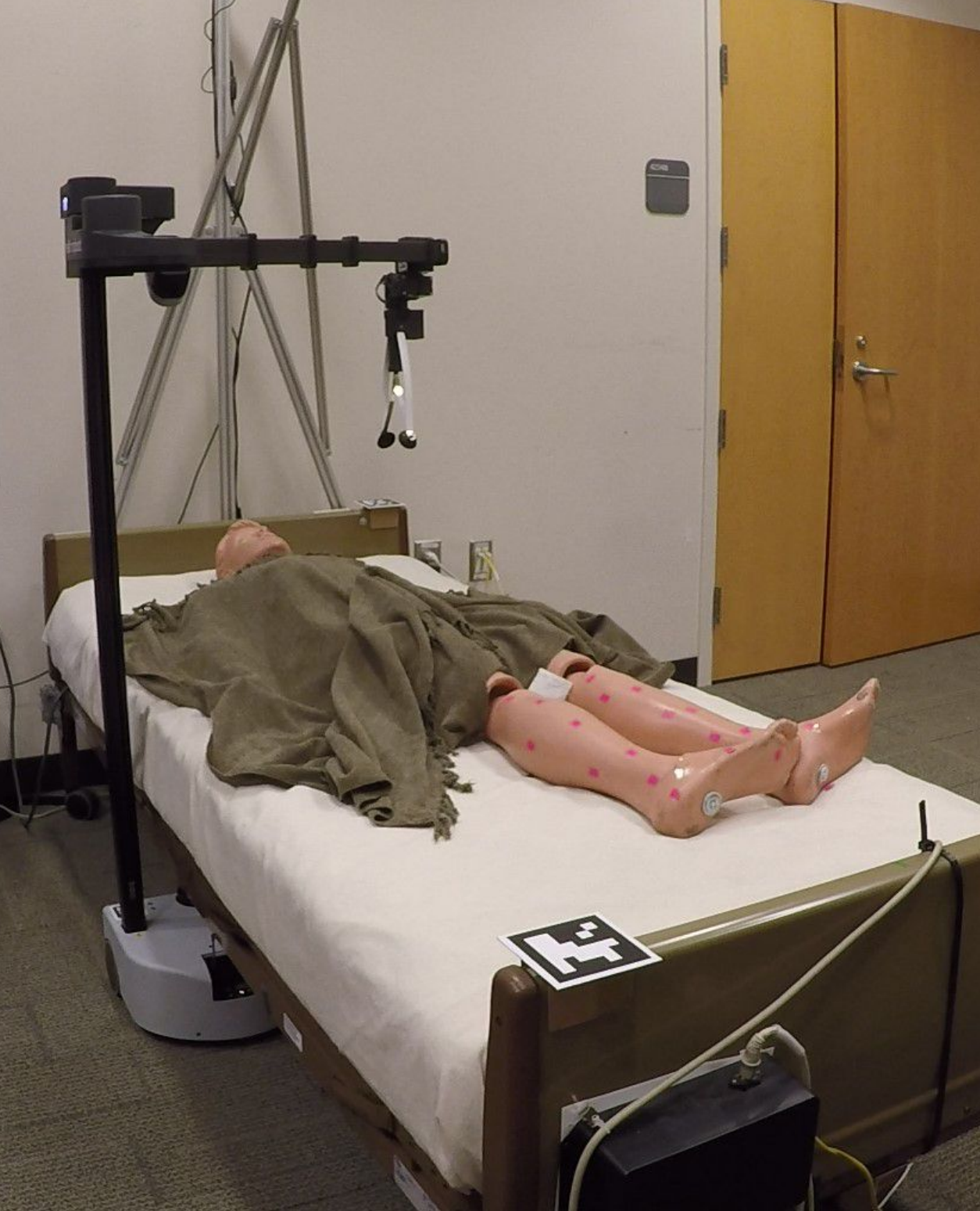}\\
\vspace{-0.3cm}
\caption{\label{fig:full_page1} Image sequence from executing a simulation-trained PPO policy to uncover both lower legs in the real world. The robot previously observed the pose of the manikin, as shown in Fig.~\ref{fig:obs}. Sequence from left to right: 1) locate fiducial tags on opposite ends of the bed, 2) move end effector to the grasp location and grasp the blanket, 3) raise the blanket 40cm up from the top of the bed, 4) move linearly to the release location, 5) release the blanket.}
\vspace{-0.4cm}
\end{figure*}

    In the second approach, we use covariance matrix adaptation evolution strategy (CMA-ES)~\cite{hansen2003reducing, hansen2019pycma} to find observation-action pairs that maximize rewards for uncovering a target body part and use those solutions to train a neural network that maps observations to actions. To collect training data, we randomly sample a human pose and use CMA-ES to find an action that optimizes the reward function for this fixed human pose. We sampled a new random human pose whenever CMA-ES discovered an action that achieved a reward value of at least 95 for the previous pose or when no such action was found after 300 simulation rollouts. We performed a total of 5,000 simulation rollouts using this CMA-ES procedure for each of the target body parts. 
    % Performing 5,000 simulation rollouts with CMA-ES took roughly 29 hours.
    
    Observation-action pairs that achieve rewards greater than 90 are then used to train a feedforward neural network that maps an observation of the human pose $\bm{s}$ to an action $\bm{a}$ for uncovering a target body part. These networks have two hidden layers of 32 nodes with ReLU activations and a four node output layer with tanh activation. We train each of the six models, one for each target body part, using observation-action pairs from the associated 5,000 simulation rollouts collected via CMA-ES. We trained each model for 100 epochs using the Adam optimizer with a mean squared error loss function, batch size of 8, and a learning rate of 0.001.

\subsection{Implementing Trained Models in the Real World}
\label{sec:real_world_setup}
	We demonstrate transfer of PPO policies trained in simulation to the real world using the Stretch RE1 mobile manipulator from Hello Robot. The bedding manipulation environment is emulated in the real world with a manikin laying supine in a hospital bed. To evaluate the performance of our policies in the real world, we placed magenta and green markers over the manikin according to the target and non-target body parts, respectively. Fig.~\ref{fig:obs} depicts this setup with the manikin. To localize Stretch's end effector with respect to the center of the real bed, we use the robot’s RGB-D head camera to detect fiducial markers on opposite diagonal ends of the hospital bed.

	Before covering the manikin with a blanket, we captured the distribution of target and non-target markers along the manikin's body using an RGB camera positioned vertically above the center of the bed. We use these markers only to evaluate performance of the real robot in Section~\ref{sec:real_world}. We place fiducial tags on the manikin’s elbows and knees, oriented to the forearms and shins respectively. The robot captured the 2D position and yaw orientation of these tags relative to the bed frame to recreate the observation $\bm{s}$ needed for our trained policies to compute grasp and release points. We used fiducial tags to capture human pose in our real-world demonstration as they provide an accurate ground truth representation of human pose unaffected by estimation error. In future work, this could be replaced with methods for pose estimation of a human covered by blankets in bed using pressure images from a pressure mat on a bed~\cite{clever2018pressurepose} or depth images from an overhead camera~\cite{clever2021depthpose}.

    A blanket similar to that used in simulation is draped over the manikin before the robot executes a rollout. At the beginning of each rollout, the robot first moves its pan-tilt camera to observe fiducial markers on the corners of the bed to define a planar coordinate system comparable to the coordinate system used in simulation. Grasp and release points are defined with respect to this coordinate system. The robot then moves its end effector to the grasp location with its gripper closed. The robot lowers its end effector until it detects contact with the bed via effort sensing in its actuators. It lifts up by 2cm to open its gripper before completing the grasp by lowering to the cloth while closing its gripper. After grasping the blanket, the robot raises its end effector to 40cm above the bed, moves linearly to the release location, and then releases the blanket. This manipulation sequence is visualized in Fig.~\ref{fig:full_page1}. Finally, we capture another image from the above-bed camera to determine the distribution of exposed target and non-target markers.
    % After manipulation, we capture another image from the above-bed camera to determine the distribution of exposed target and non-target markers.

\new{
\subsection{Simplifying Assumptions}
\label{sec:key_assumptions}
    In order to reasonably reduce the complexity of the bedding manipulation task, we make simplifying assumptions including the following: 
    1) human pose information is considered known in both simulation and the real-world, 
    2) models are trained on a human of fixed body shape similar to that of the medical manikin used in our real world evaluation, and
    3) a robot's end effector follows a linear trajectory between a single pair of grasp and release points that are bounded to the dimensions of the bed.
}

\subsection{Future Research Directions}

In our simulation environment, we drop a blanket of fixed size and configuration onto a human model in bed. Future work could consider training policies with variation in the fabric properties or in the configuration of the blanket over a person. In Section~\ref{sec:generalize}, we evaluate how our policies generalize when we randomize the configuration of a blanket over a person in bed and vary human body shape. In addition, while we consider humans lying supine in bed, there is an opportunity to explore greater human pose variation, including prone, lateral recumbent, arms crossed, and others. Training policies that account for these different variations would require that more detailed human pose and cloth state information are incorporated into a robot's observation.
\new{Prior work has used linear trajectories for quasistatic cloth manipulation \cite{seita2019bedmaking, hoque2021visuospatial, hoque2021visuospatial2}. The potential costs and benefits of more complex trajectories would be an interesting topic for future study.}

\begin{figure*}
\centering
\hspace{-0.3cm}
\begingroup
\fontsize{7}{11}\selectfont
\setlength{\tabcolsep}{5pt}
\begin{tabular}{ccc}
 Uncovered & Initial State & Final State \\
\end{tabular}
\endgroup
\hspace{0.1cm}
\begingroup
\fontsize{7}{11}\selectfont
\setlength{\tabcolsep}{5pt}
\begin{tabular}{ccc}
 Uncovered & Initial State & Final State \\
\end{tabular}
\endgroup
\hspace{0cm}
\begingroup
\fontsize{7}{11}\selectfont
\setlength{\tabcolsep}{5pt}
\begin{tabular}{ccc}
 Uncovered & Initial State & Final State \\
\end{tabular}\\
%  \vspace{-0.2cm}
\endgroup
 \includegraphics[width=0.08 \textwidth, trim={0cm 0cm 0cm 0cm}, clip]{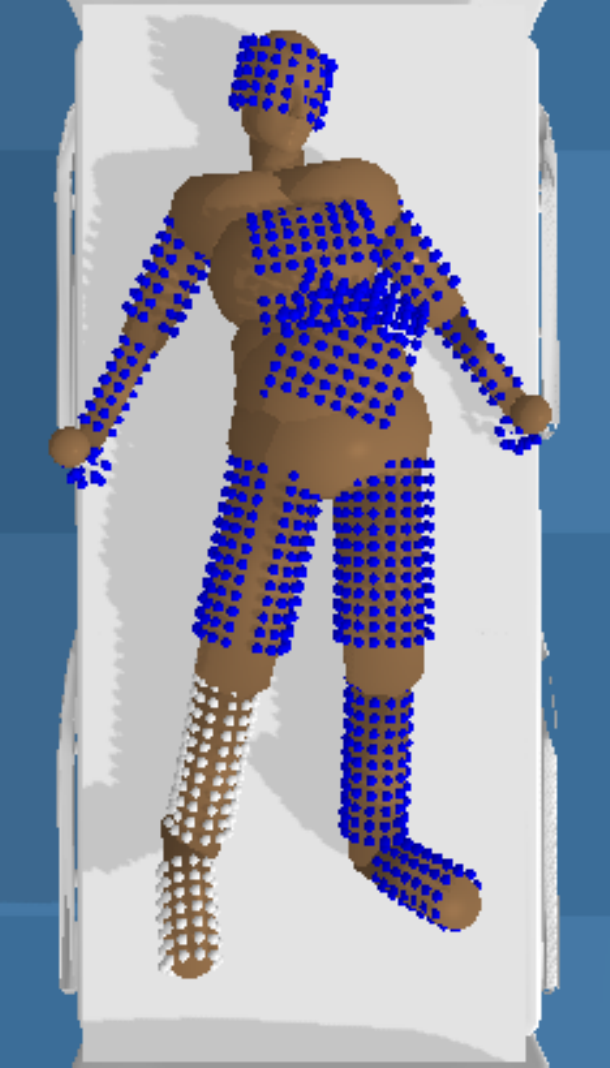}
 \hspace{-0.2 cm}
 \includegraphics[width=0.08\textwidth, trim={0cm 0cm 0cm 0cm}, clip]{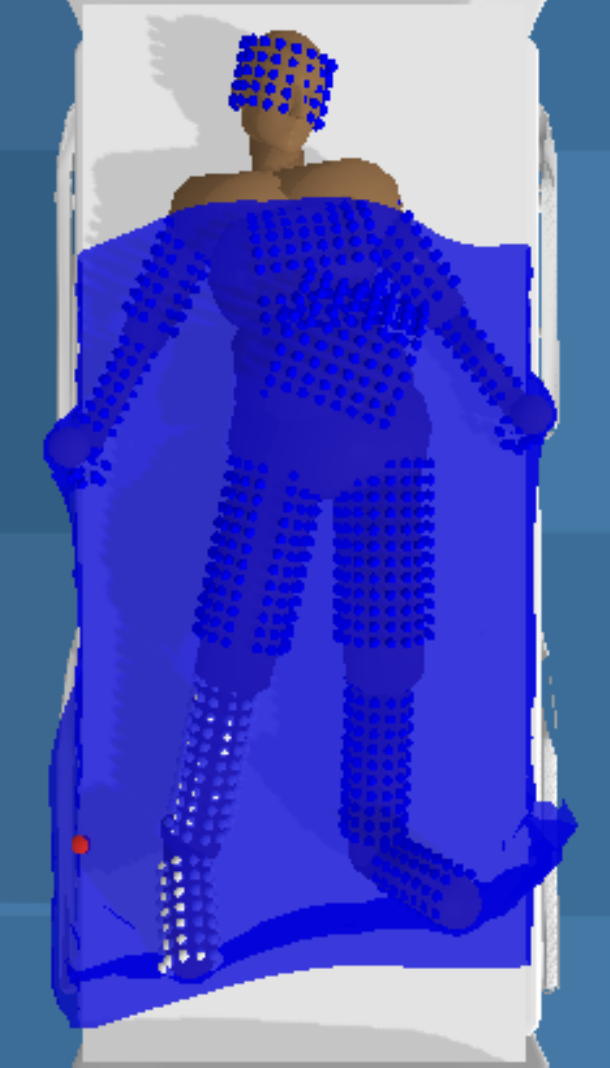}
 \hspace{-0.2cm}
 \includegraphics[width=0.08\textwidth, trim={0cm 0cm 0cm 0cm}, clip]{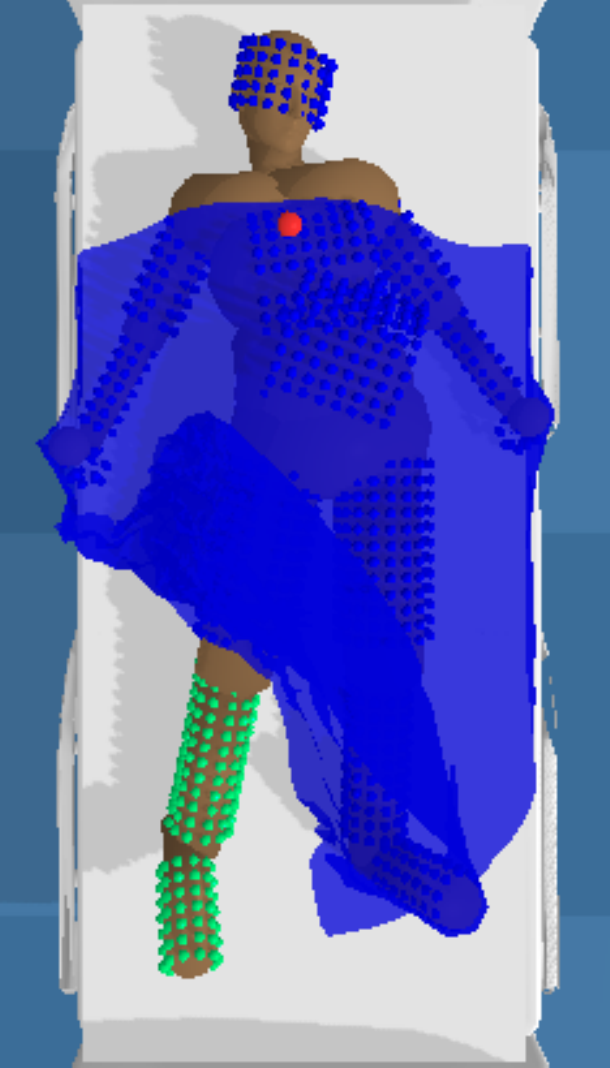}
 \hspace{0.1 cm}
 \includegraphics[width=0.08\textwidth, trim={0cm 0cm 0cm 0cm}, clip]{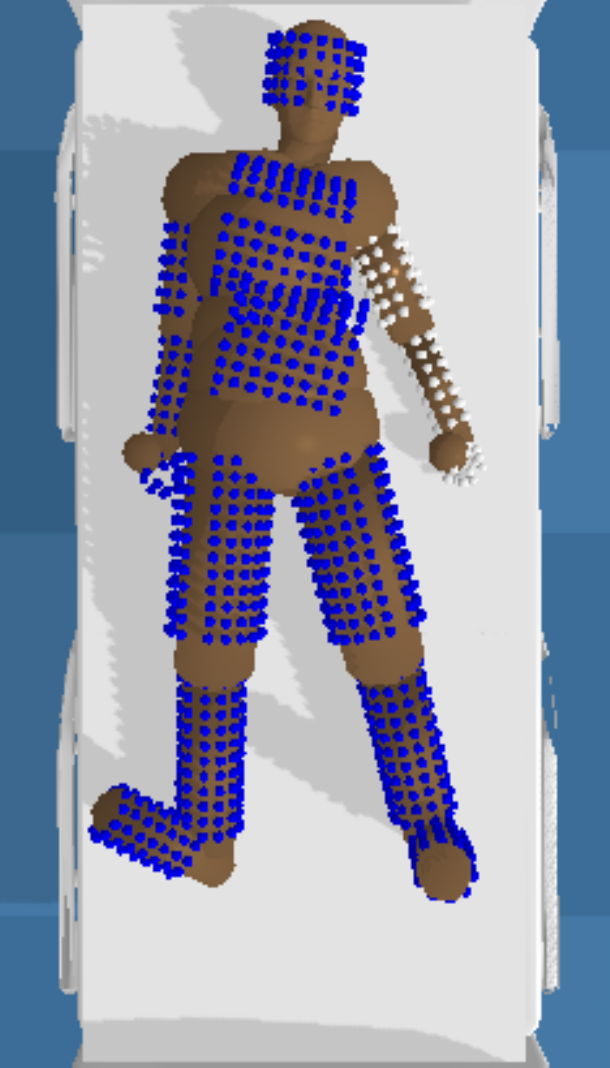}
  \hspace{-0.2 cm}
 \includegraphics[width=0.08\textwidth, trim={0cm 0cm 0cm 0cm}, clip]{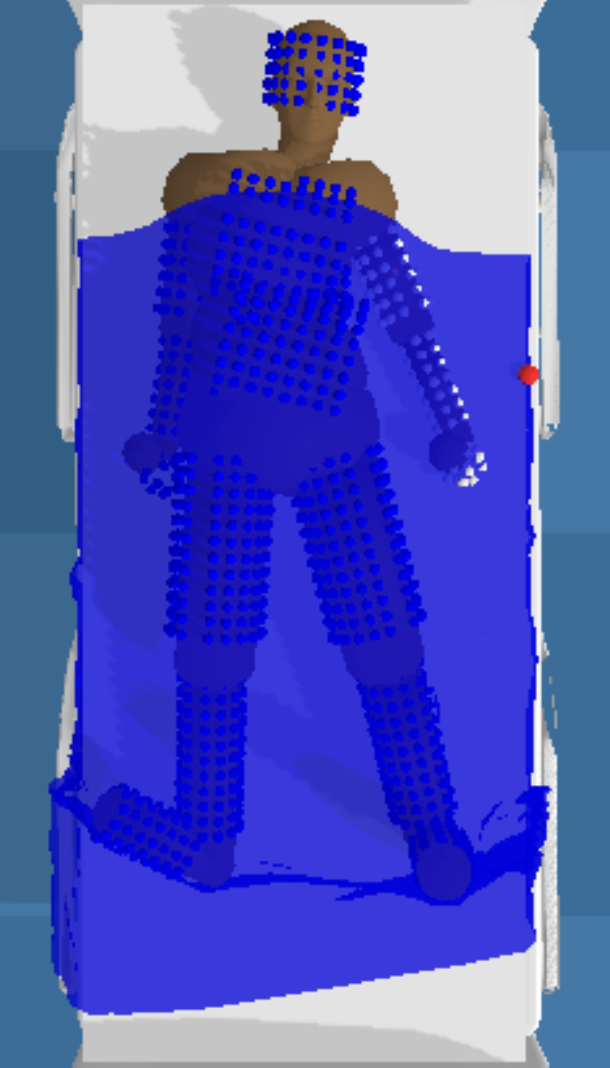}
  \hspace{-0.2 cm}
 \includegraphics[width=0.08\textwidth, trim={0cm 0cm 0cm 0cm}, clip]{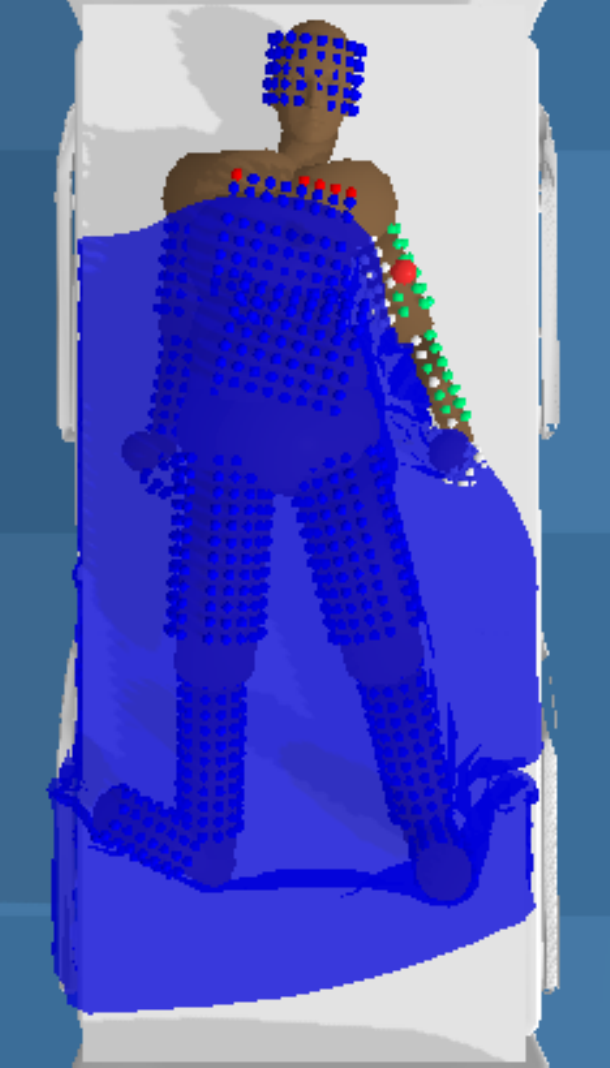}
 \hspace{0.1 cm}
 \includegraphics[width=0.08\textwidth, trim={0cm 0cm 0cm 0cm}, clip]{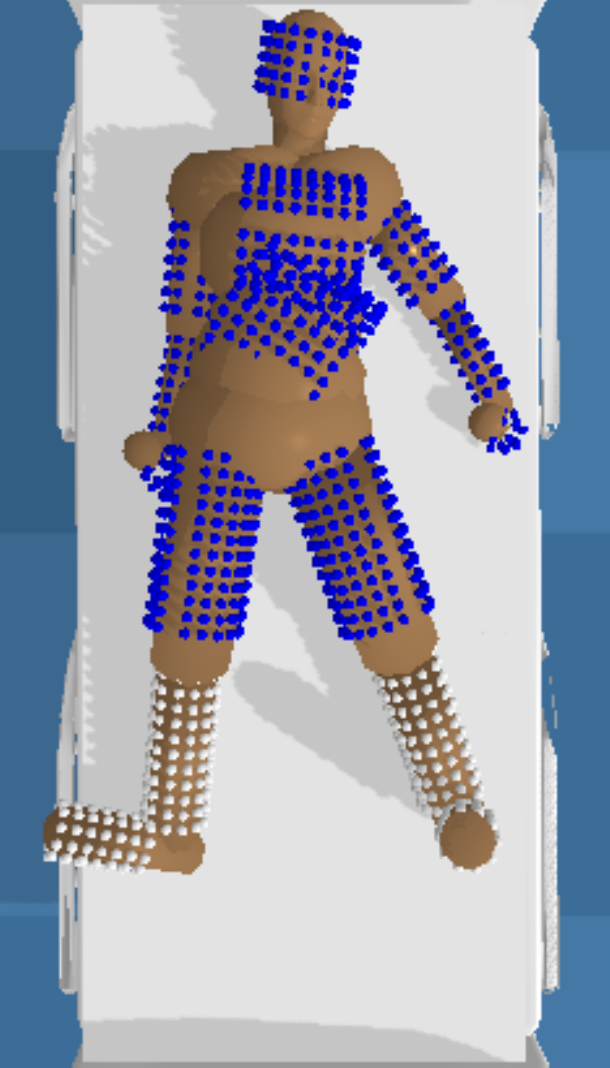}
  \hspace{-0.2 cm}
  \includegraphics[width=0.08\textwidth, trim={0cm 0cm 0cm 0cm}, clip]{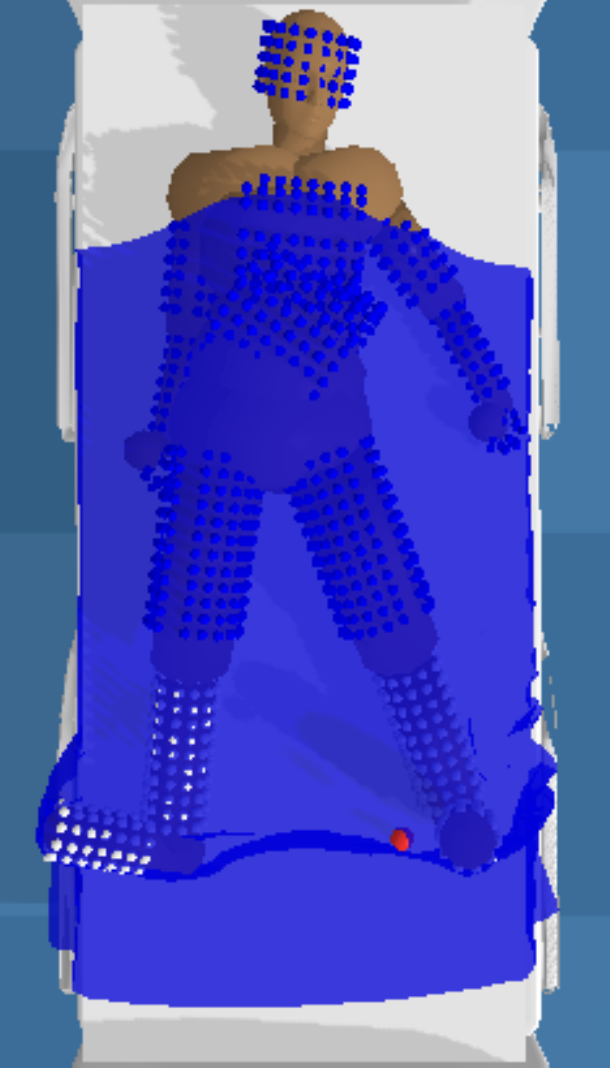}
   \hspace{-0.2 cm}
 \includegraphics[width=0.08\textwidth, trim={0cm 0cm 0cm 0cm}, clip]{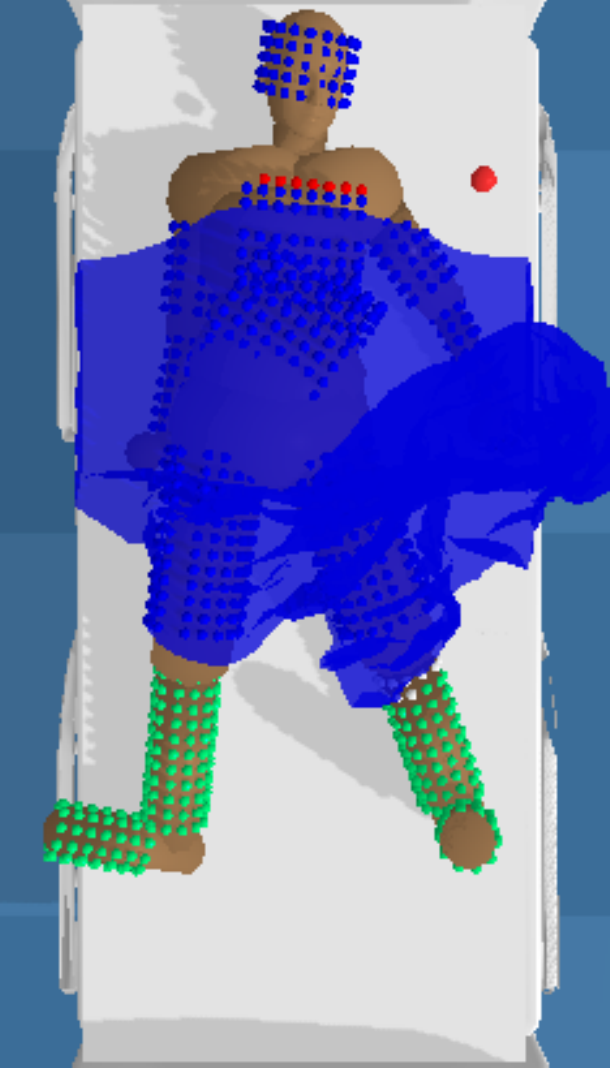}\\
 \vspace{-0.1cm}
\begingroup
\setlength{\tabcolsep}{19pt}
\fontsize{7}{11}\selectfont
 \begin{tabular}{ccc}
   Right Lower Leg, $R(\bm{S}, \bm{a}) = 97.7$ & Left Arm, $R(\bm{S}, \bm{a}) = 36.6$ &  Both Lower Legs, $R(\bm{S}, \bm{a}) = 81.8$ \\
\end{tabular}\\
\endgroup
 \vspace{0.1 cm}
 \includegraphics[width=0.08\textwidth, trim={0cm 0cm 0cm 0cm}, clip]{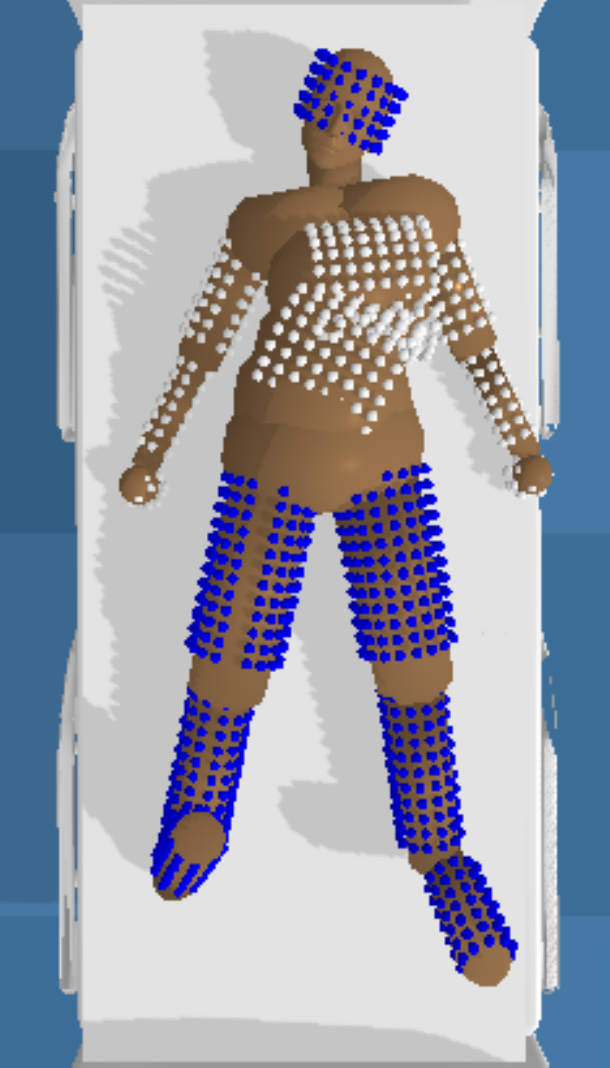}
  \hspace{-0.2 cm}
  \includegraphics[width=0.08\textwidth, trim={0cm 0cm 0cm 0cm}, clip]{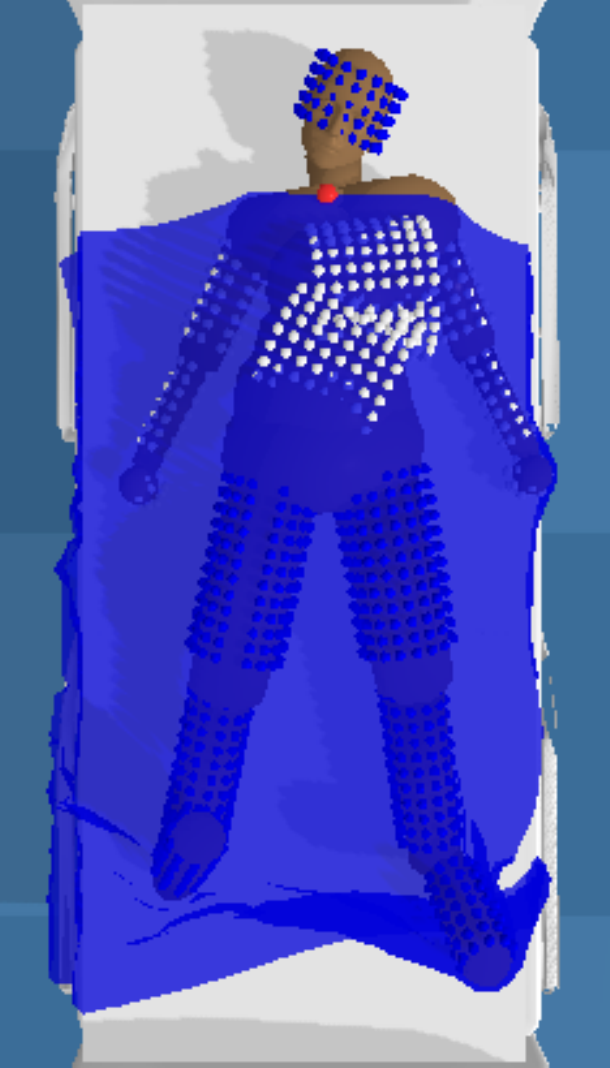}
   \hspace{-0.2 cm}
 \includegraphics[width=0.08\textwidth, trim={0cm 0cm 0cm 0cm}, clip]{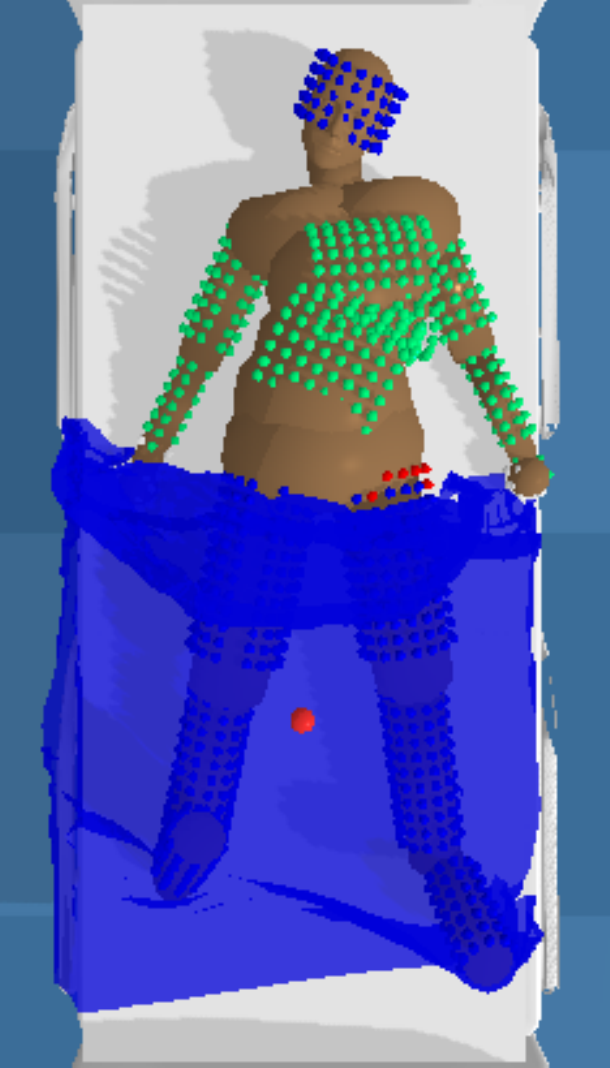}
 \hspace{0.1 cm}
 \includegraphics[width=0.08\textwidth, trim={0cm 0cm 0cm 0cm}, clip]{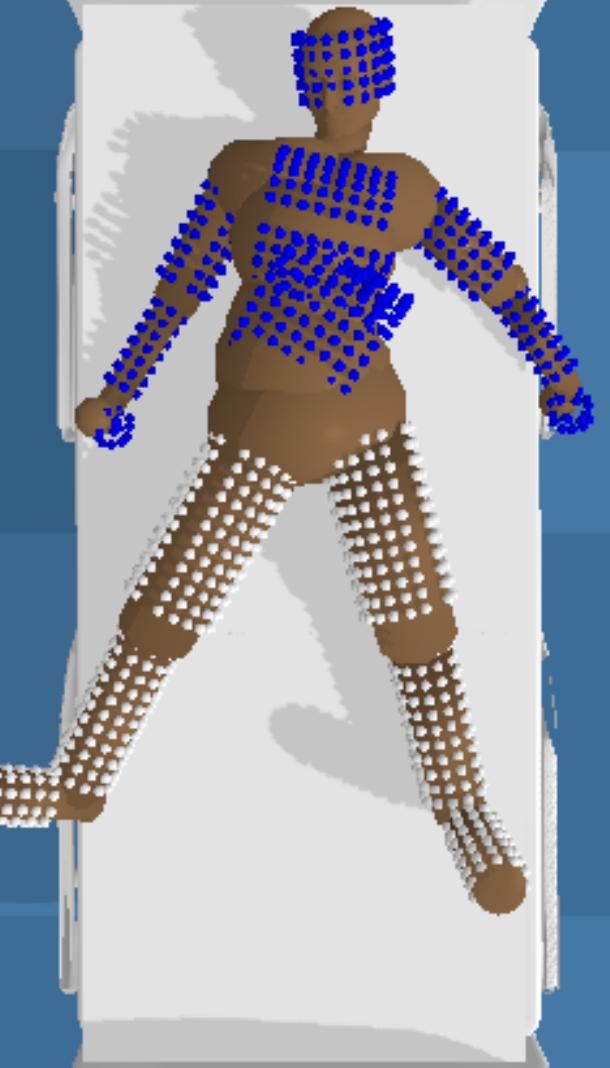}
  \hspace{-0.2 cm}
  \includegraphics[width=0.08\textwidth, trim={0cm 0cm 0cm 0cm}, clip]{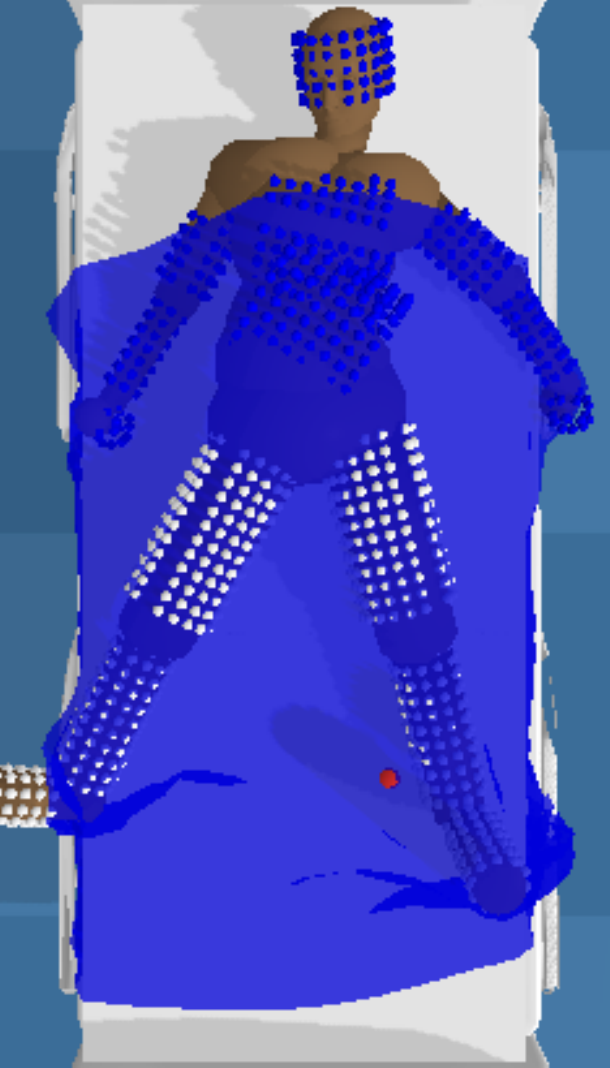}
   \hspace{-0.2 cm}
 \includegraphics[width=0.08\textwidth, trim={0cm 0cm 0cm 0cm}, clip]{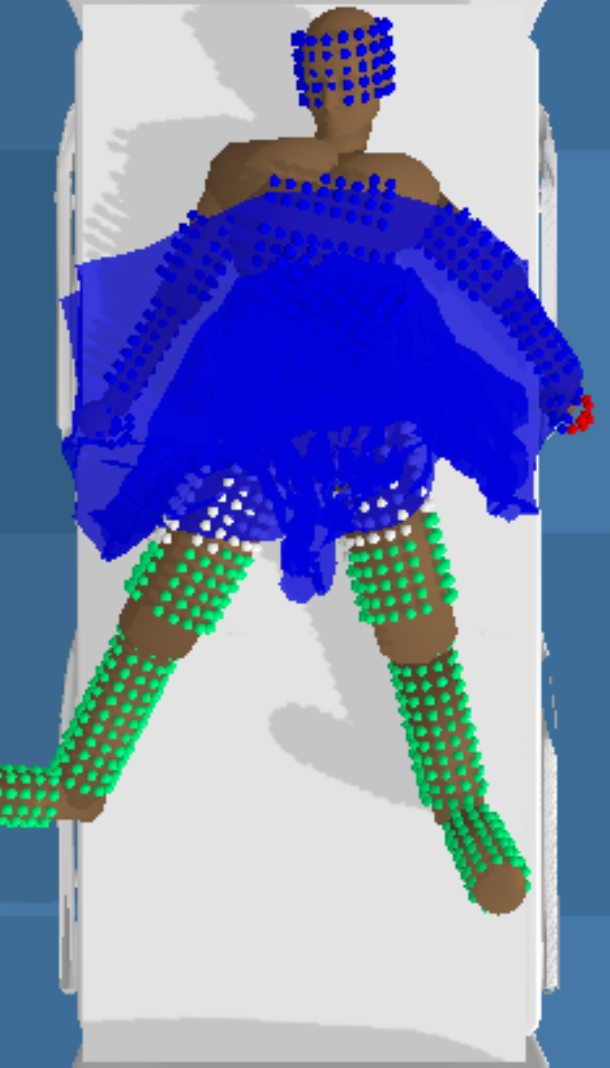}
 \hspace{0.1 cm}
 \includegraphics[width=0.08\textwidth, trim={0cm 0cm 0cm 0cm}, clip]{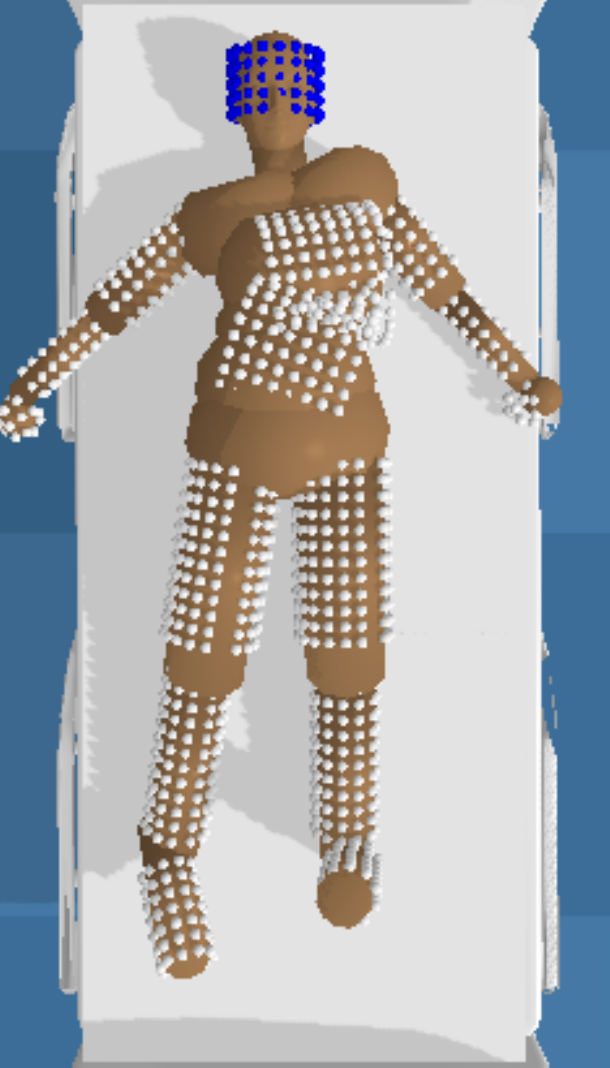}
  \hspace{-0.2 cm}
  \includegraphics[width=0.08\textwidth, trim={0cm 0cm 0cm 0cm}, clip]{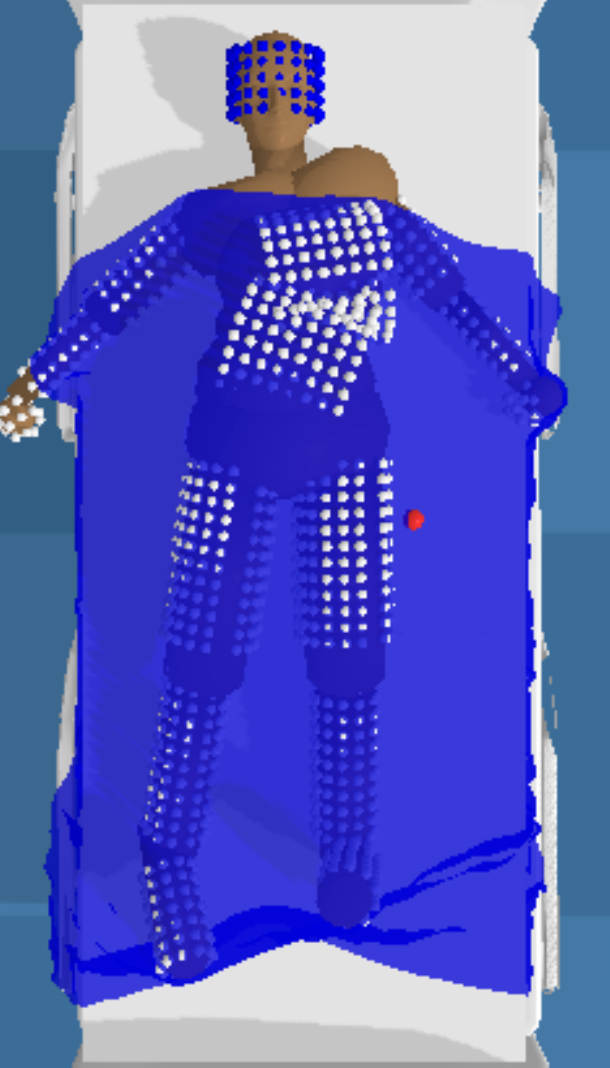}
   \hspace{-0.2 cm}
 \includegraphics[width=0.08\textwidth, trim={0cm 0cm 0cm 0cm}, clip]{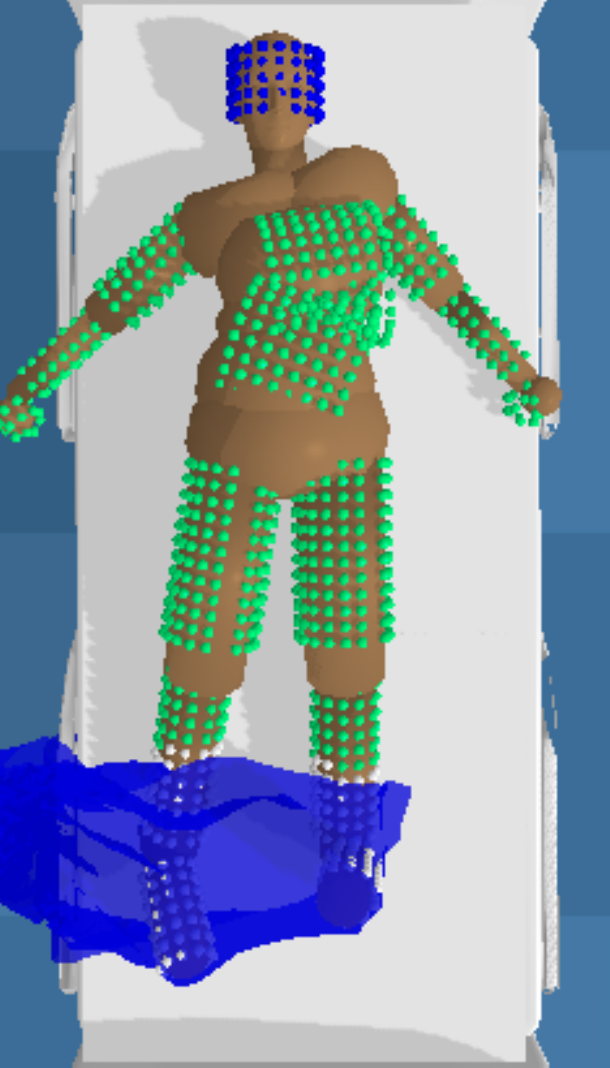}\\
 \begingroup
 \vspace{-0.1cm}
\setlength{\tabcolsep}{22pt}
\fontsize{7}{11}\selectfont
 \begin{tabular}{ccc}
   Upper Body, $R(\bm{S}, \bm{a}) = 92.3$ & Lower Body, $R(\bm{S}, \bm{a}) = 70.5$ &  Entire Body, $R(\bm{S}, \bm{a}) = 83.3$ \\
\end{tabular}\\
\endgroup
\caption{\label{fig:ppo_render} Images from evaluating all six trained PPO policies in simulation. The reward value for each outcome is shown for each target limb. }
\vspace{-0.4cm}
\end{figure*}

\section{Evaluation}
\label{sec:evaluation}
\subsection{Simulation}

	We evaluate and compare the six PPO policies and the six self-supervised learning models trained using data collected by CMA-ES over 100 randomized simulation rollouts. In order to evaluate performance, we define and compute the following metrics:
\begin{itemize}
\item \(\mu_{R}\): mean reward
\item $\text{True positive }(TP) = \rho_{t}$%, target points uncovered
\item $\text{False positive }(FP) = \rho_{n}$%, non-target points uncovered
\item $\text{False negative }(FN) = |\bm{P_{t}}| - \rho_{t}$%, target points not uncovered
% \item $\text{Precision} = TP /(TP+FP) = \rho_{t}/(\rho_{t} + \rho_{n})$ 
% \item $\text{Recall}= TP/(TP+FN) =  \rho_{t}/|\bm{P_{t}}|$
% \item $F_1=  2*\frac{\text{Precision}*\text{Recall}}{\text{Precision}+\text{Recall}} = \frac{TP}{TP + 0.5(FP+FN)}$
\item \new{$\text{F-Score } (F_1) = \frac{TP}{TP + 0.5(FP+FN)}$}
\end{itemize}

% \ck{In a future paper, it would be interesting to provide more analysis of the fitness landscape, which should be easy to visualize in 2D. For example, fix the grasp point and show how the fitness varies with different release points using a color map across the plane with sampled points. Similarly, fix release point and vary the grasp point. It could also be interesting to see how the points a policy provides vary as the observations are varies. I'm less sure how to visualize this, but the low-dimensional 2D structure should present opportunities.}
% \ck{I find the following text a little confusing. For example, "non-target points incorrectly uncovered" seems a bit like a double negative. How about "TP is the number of target points uncovered. FP is the number of non-target points uncovered. FN is the number of target points that remained covered."}
% TP represents the number of target points successfully uncovered while FP represents the number of non-target points incorrectly uncovered. FN refers to the number of target points that were incorrectly left covered.
\new{TP represents the number of target points uncovered while FP represents the number of non-target points uncovered. FN refers to the number of target points that remained covered. F-score is a measure of the robot's accuracy at performing bedding manipulation, where $F_1 \in [0, 1]$. Average F-score and reward ($\pm$ standard deviation) metrics are computed over 100 simulation trials.}

\begin{table}
% \vspace{0.4cm}
\centering
% \vspace{6pt}
\caption{\label{table:ppo+cmaes_eval_sim} Evaluation results in simulation for PPO policies and policies trained with data collected using CMA-ES.}
\begin{tabular}{ccccc} \toprule
    & \multicolumn{2}{c}{PPO} & \multicolumn{2}{c}{CMA-ES}\\
    \cmidrule(lr){2-3} \cmidrule(lr){4-5}
    Target & \(F_{1}\) &\(\mu_{R}\) &\(F_{1}\) & \(\mu_{R}\) \\ \midrule\midrule
    Right Lower Leg & 0.72 &54.6 $(\pm36.5)$ & 0.78 & 44.4 $(\pm75.4)$ \\
    Left Arm & 0.35 &4.4 $(\pm76.7)$ & 0.20 & -10.5 $(\pm70.7)$ \\
    Lower Legs & 0.86 &69.7 $(\pm55.2)$ & 0.93  & 84.8 $(\pm11.9)$ \\
    Upper Body & 0.96 &92.9 $(\pm3.4)$ & 0.96 & 91.5 $(\pm4.6)$ \\
    Lower Body & 0.89 &72.0 $(\pm37.6)$ & 0.91 & 65.1 $(\pm45.5)$ \\
    Entire Body & 0.94 &87.2 $(\pm17.4)$ & 0.94 & 88.7 $(\pm7.5)$ \\
	\bottomrule
\end{tabular}
% \vspace{0cm}
\vspace{-0.6cm}
\end{table}

% \ck{"performed best" confused me here. it made me think the comparison was with the CMA-ES policy. maybe you could clarify that performance varied with respect to body parts with examples of best and worst. is there a standard way of creating a single figure of merit (performance number) using recall and precision? how do people typically present them? are average recall and average precision the de facto standards? }
% \zackory{We could use F1 scores as a single figure of merit. F1 = 2*precision*recall / (precision+recall)}

\new{As shown in Table~\ref{table:ppo+cmaes_eval_sim}, we find that our policies trained with PPO achieved F-scores greater than 0.85 for all but two target limbs. Fig.~\ref{fig:ppo_render} visualizes the before and after states in simulation with the PPO-trained policies for uncovering each of the target body parts.} While there is a trivial solution to uncovering the entire body by grabbing the blanket and pulling it off the bed, our policies instead find a solution where much of the body is uncovered while ensuring the blanket remains on the bed, as seen in Fig.~\ref{fig:ppo_render} (bottom, right).

\new{Performance of the reinforcement learning and self-supervised learning (CMA-ES) formulations were similar, as shown in Table~\ref{table:ppo+cmaes_eval_sim}. Both the policies trained via PPO and the models trained with data from CMA-ES had comparable values for both F-score and reward metrics.

Based on the mean F-score and reward values, uncovering the left arm was a particularly challenging task for both methods}. This is due to the proximity of the arm to the torso, which often results in a large number of non-target torso points being uncovered when attempting to uncover the arm. One potential solution could be models that output nonlinear end effector trajectories that can make small adjustments to the cloth in these tightly constrained scenarios.

\begin{figure}
\centering
\begingroup
\tiny
\begin{tabular}{cccccc}
 \hspace{-0.1cm} Uncovered & \hspace{0.15cm} Initial State  & \hspace{0.1cm} Final State& \hspace{0.2cm}  Uncovered & \hspace{0.15cm}  Initial State & \hspace{0.1cm}  Final State\\
\end{tabular}
\endgroup\\
\vspace{0cm}
\includegraphics[width=0.076\textwidth, trim={0.6cm 0.5cm 0.5cm 0cm}, clip]{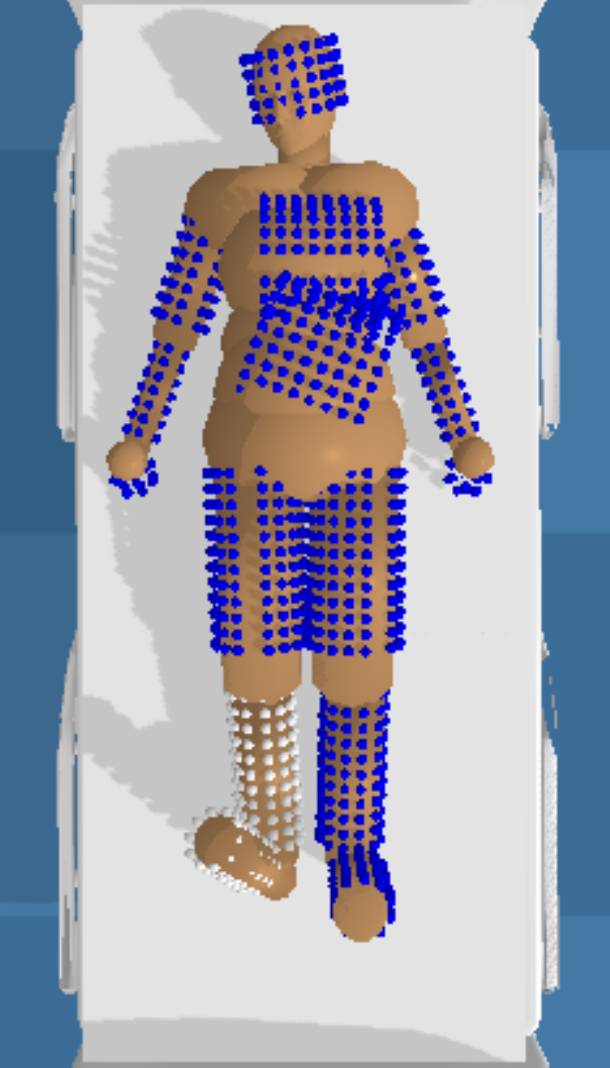} \hspace{-0.2cm}
\includegraphics[width=0.076\textwidth, trim={0.6cm 0.5cm 0.5cm 0cm}, clip]{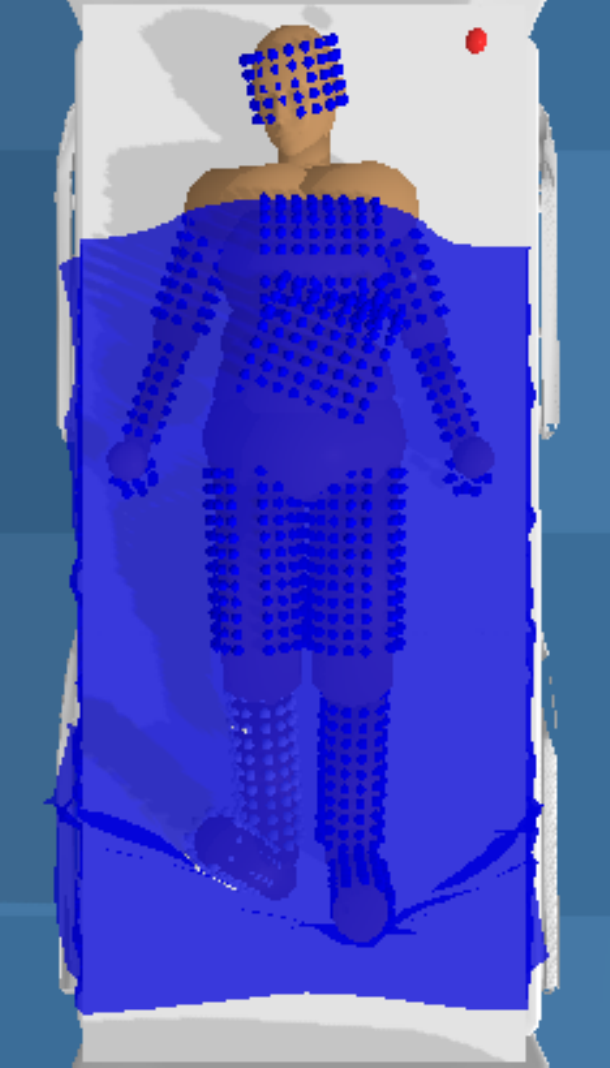} \hspace{-0.2cm}
\includegraphics[width=0.076\textwidth, trim={0.6cm 0.5cm 0.5cm 0cm}, clip]{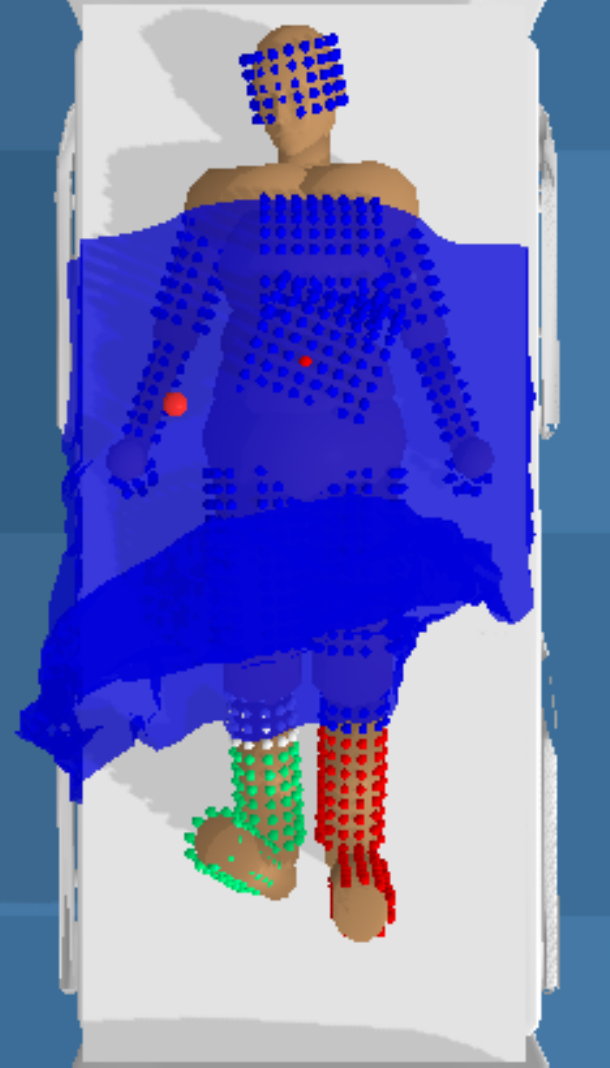} \hspace{-0.2cm}
\hspace{-0.05cm}
\includegraphics[width=0.076\textwidth, trim={0.6cm 0.5cm 0.5cm 0cm}, clip]{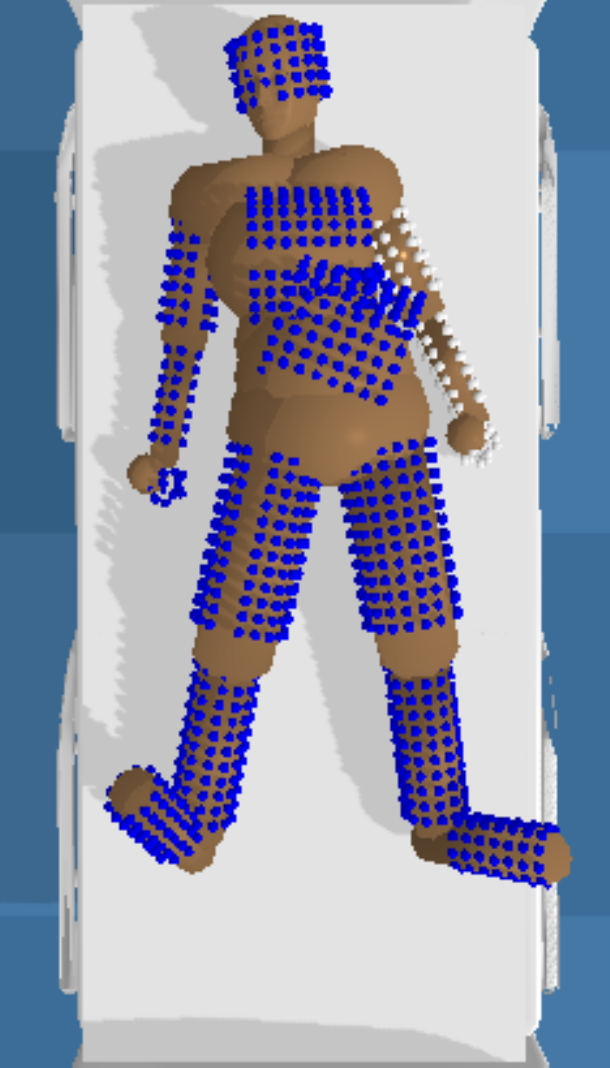} \hspace{-0.2cm}
\includegraphics[width=0.076\textwidth, trim={0.6cm 0.5cm 0.5cm 0cm}, clip]{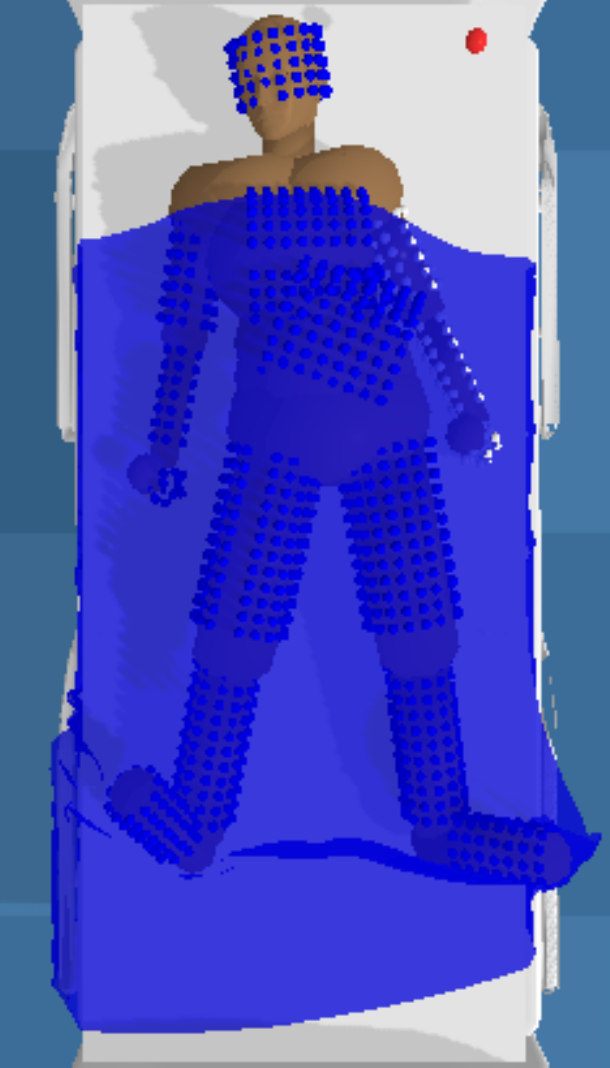} \hspace{-0.2cm}
\includegraphics[width=0.076\textwidth, trim={0.6cm 0.5cm 0.5cm 0cm}, clip]{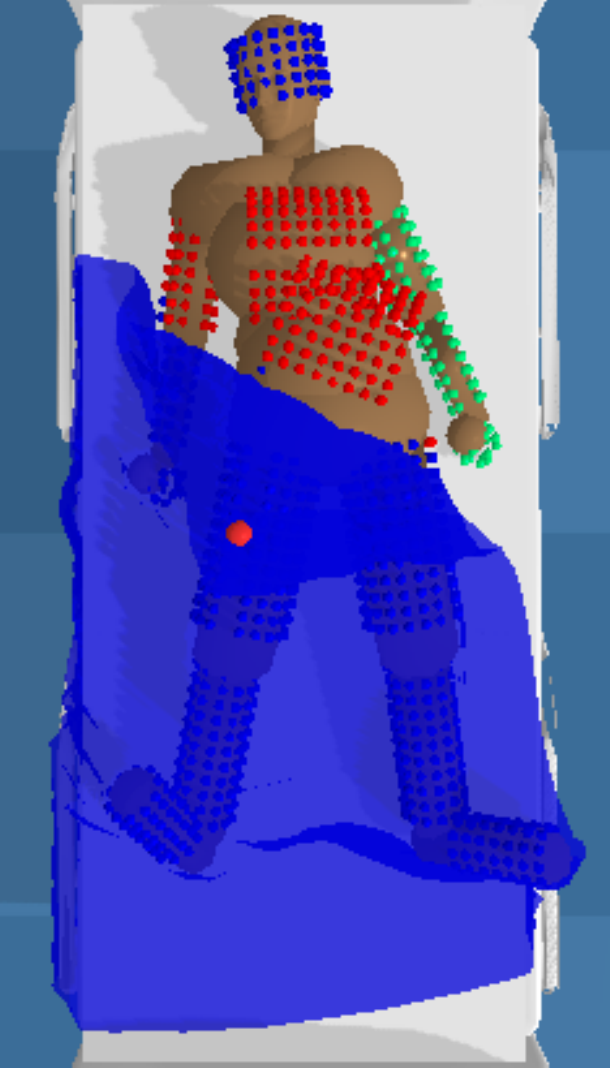} \hspace{-0.2cm}\\
\vspace{-0.1cm}
\begingroup
\fontsize{7}{11}\selectfont
\begin{tabular}{cc}
 \hspace{-0.4cm} Right Lower Leg, $R(\bm{S}, \bm{a}) = -3.7$  & \hspace{0.3cm} Left Arm, $R(\bm{S}, \bm{a}) = -316$ \\
\end{tabular}
\endgroup\\

\vspace{-0.2cm}

\caption{\label{fig:failures} Failure cases: uncovering the right lower leg and the left arm.}
\vspace{-0.5cm}
\end{figure}

Uncovering the right lower leg was another challenging task. The closer the legs were to one another, the more challenging it became for policies to find solutions that uncovered just a single leg. Fig.~\ref{fig:failures} depicts failure cases when attempting to uncover the left arm and the right lower leg.
% Despite this difficulty, performance was substantially greater for uncovering the right lower leg as compared to the left arm, as shown by all three metrics.

% \begin{table}
% % \vspace{0.4cm}
% \centering
% % \vspace{6pt}
% \caption{\label{table:cmaes_eval_sim} Evaluation results in simulation for policies trained with data collected using CMA-ES.}
% \begin{tabular}{cccc} \toprule
%     Target & Recall & Precision & \(\mu_{R}\) \\ \midrule\midrule
%     Right Lower Leg & 0.86 & 0.68 & 44.4 $(\pm75.4)$ \\
%     Left Arm & 0.21 & 0.40 & -10.5 $(\pm70.7)$ \\
%     Lower Legs & 0.97 & 0.89 & 84.8 $(\pm11.9)$ \\
%     Upper Body & 0.96 & 0.95 & 91.5 $(\pm4.6)$ \\
%     Lower Body & 0.91 & 0.91 & 65.1 $(\pm45.5)$ \\
%     Entire Body & 0.89 & 1.00 & 88.7 $(\pm7.5)$ \\
% 	\bottomrule
% \end{tabular}
% % \vspace{0cm}
% \vspace{-0.6cm}
% \end{table}

% \begin{table}
% % \vspace{0.4cm}
% \centering
% % \vspace{6pt}
% \caption{\label{table:cmaes_eval_sim} Evaluation results in simulation for policies trained with data collected using CMA-ES.}
% \begin{tabular}{ccc} \toprule
%     Target &  \(F_{1}\) & \(\mu_{R}\) \\ \midrule\midrule
%     Right Lower Leg & 0.78 & 44.4 $(\pm75.4)$ \\
%     Left Arm & 0.20 & -10.5 $(\pm70.7)$ \\
%     Lower Legs & 0.93  & 84.8 $(\pm11.9)$ \\
%     Upper Body & 0.96 & 91.5 $(\pm4.6)$ \\
%     Lower Body & 0.91 & 65.1 $(\pm45.5)$ \\
%     Entire Body & 0.94 & 88.7 $(\pm7.5)$ \\
% 	\bottomrule
% \end{tabular}
% % \vspace{0cm}
% \vspace{-0.6cm}
% \end{table}

\subsection{Generalizing to Novel Human Bodies and Blanket Configurations}
\label{sec:generalize}

In this section, we break the assumptions discussed in Section~\ref{sec:generalize} and evaluate how well learned policies generalize to varying the initial blanket state and to varying human body size. In order to evaluate generalization to initial blanket configurations, we modify the original simulation environment described in Section~\ref{sec:sim_env} to randomize the initial blanket pose before being dropped on a human by introducing variation uniformly sampled between \(\pm\)2cm to the initial $x$ position, \new{[-25, 5]cm} to the initial $y$ position, and \(\pm\)\new{45} degrees to the initial yaw $\theta_z$ orientation. This variation is defined such that the human body, excluding the head, remains completely covered at the start of a trial.

\begin{figure}
\centering
 \includegraphics[width=0.085\textwidth, trim={0.5cm 0.5cm 0.4cm 0cm}, clip]{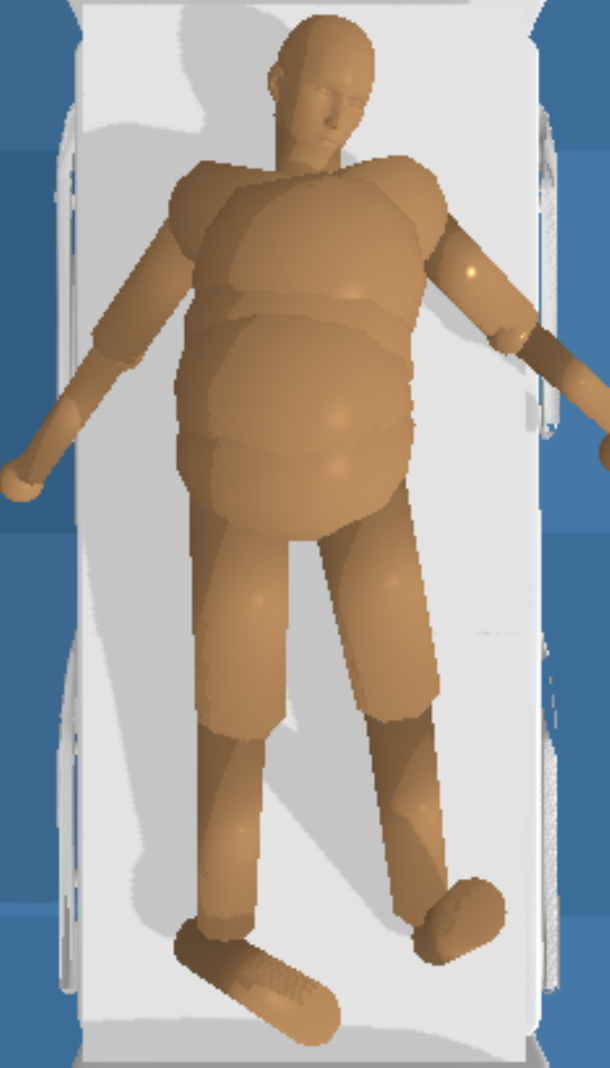} \hspace{-0.2cm}
 \includegraphics[width=0.085\textwidth, trim={0.5cm 0.5cm 0.4cm 0cm}, clip]{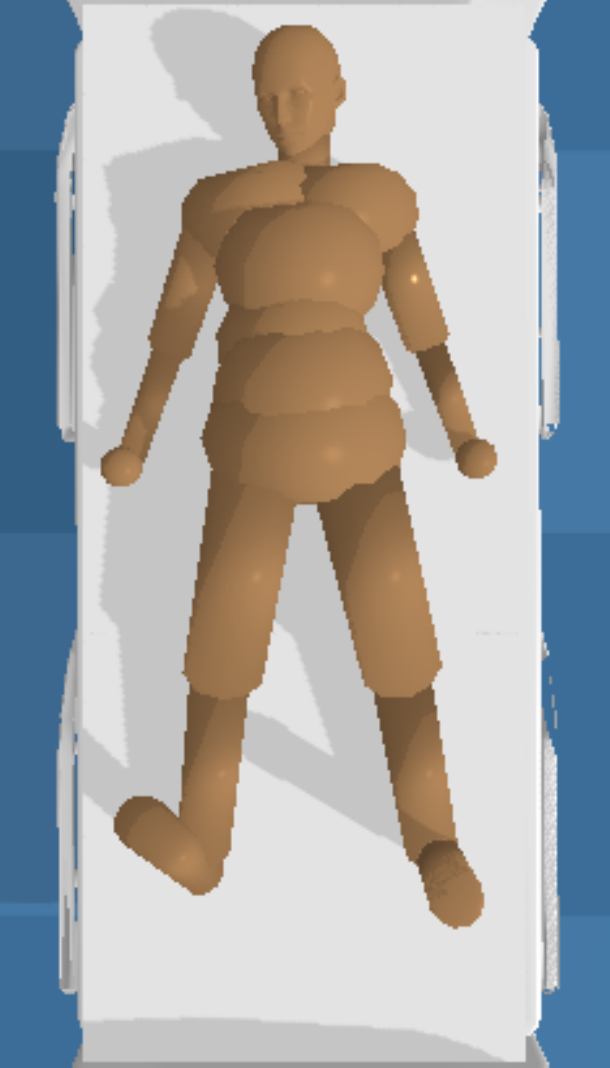} \hspace{0cm}
 \includegraphics[width=0.085\textwidth, trim={0.5cm 0.5cm 0.4cm 0cm}, clip]{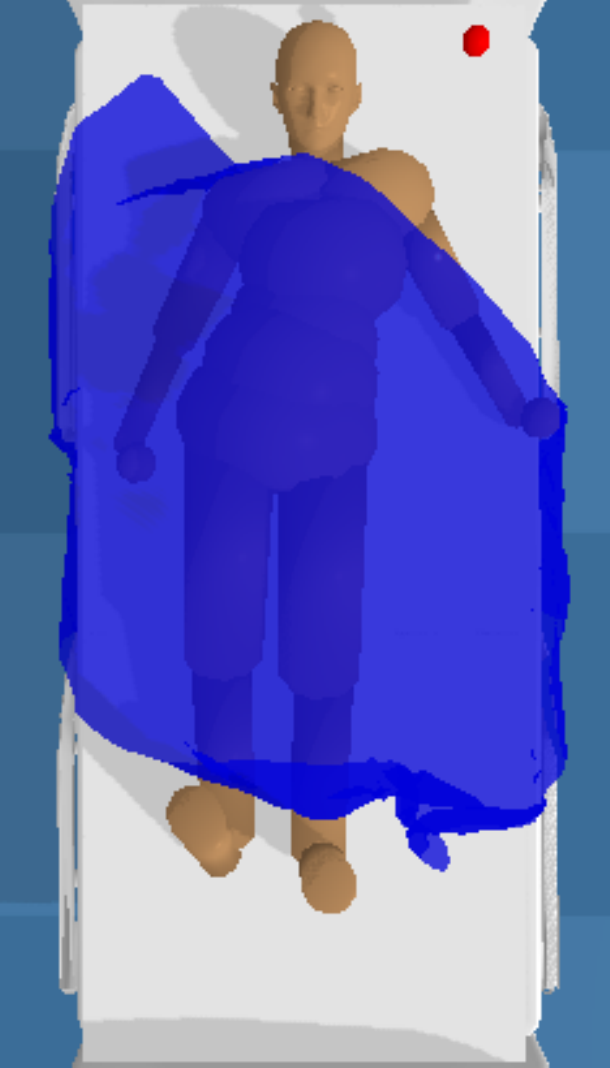} \hspace{-0.2cm}
 \includegraphics[width=0.085\textwidth, trim={0.5cm 0.5cm 0.4cm 0cm}, clip]{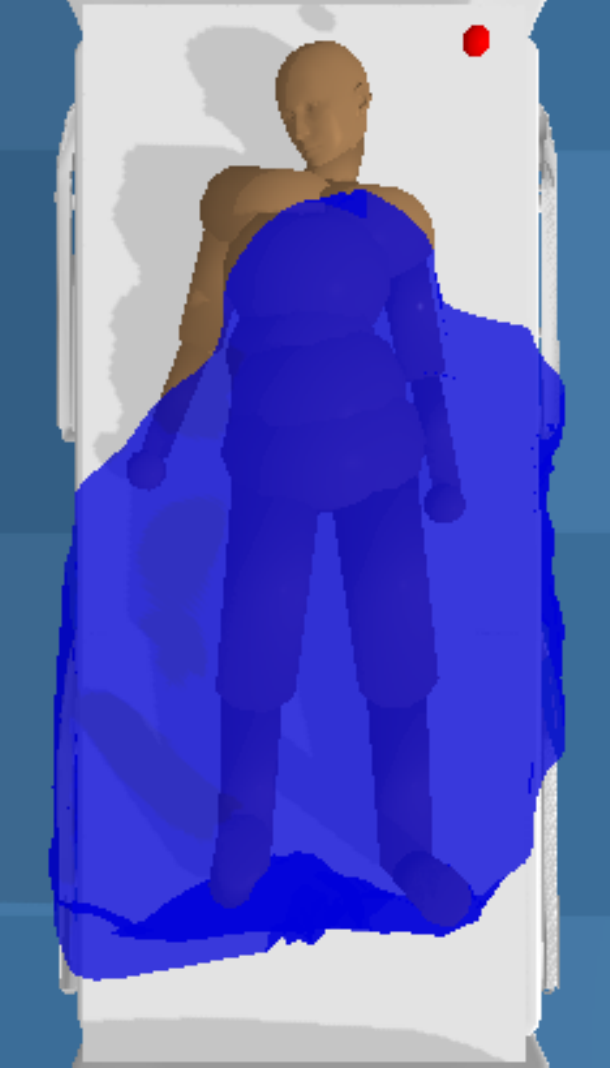}
 \vspace{-0.3cm}
\caption{\label{fig:generalize_examples} \new{Variation examples: human body shape and initial blanket state.}}
\vspace{-0.2cm}
% \vspace{0.2cm}
\end{figure}

To evaluate generalization of learned policies to varied human body size, we first generate an SMPL-X body mesh~\cite{pavlakos2019expressive} defined using 10 uniformly sampled body shape parameters $\bm{\beta} \in U(0, 4)$. We optimize the body parts of the capsulized human model in Assistive Gym to fit the randomized SMPL-X human mesh and we then drop this new capsulized human model with joints onto the bed as described in~\ref{sec:sim_env}. This method of varying human body size produced human models ranging between 160cm to 185cm in height. \new{Examples of the variation of human body size and initial blanket configuration introduced in the environment are shown in Fig.~\ref{fig:generalize_examples}}.

\begin{figure}
\centering
\includegraphics[width=0.48 \textwidth, trim={0.16cm 0.32cm 2.2cm 1.68cm}, clip]{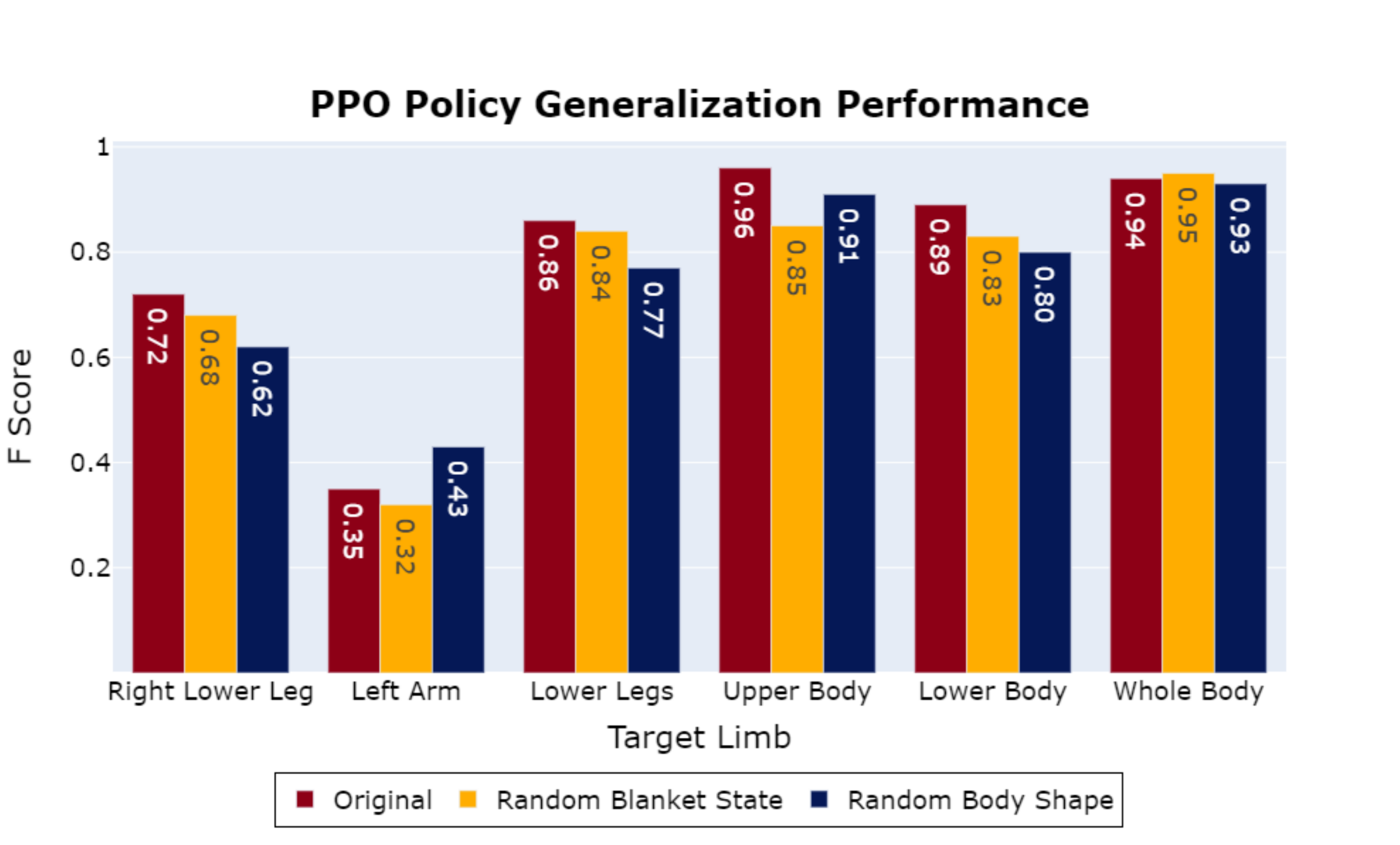}
\vspace{0cm}
\caption{\label{fig:generalize} \new{Comparison of F-score when the trained PPO policies uncovered target body parts in the original simulation conditions (fixed initial blanket state and fixed body shape) and in the two generalization cases (random initial blanket state and random body shape).}}
\vspace{-0.5cm}
\end{figure}

\new{We then evaluated how the PPO-trained policies for each target limb generalized to these novel blanket configurations and human body sizes. Fig.~\ref{fig:generalize} visualizes the F-score averaged over 100 simulation rollouts for the original environments (see Table~\ref{table:ppo+cmaes_eval_sim}) compared to environments with random blanket states or human body shapes. For a few target body parts, F-scores remained consistent across the generalization cases. However, there are scenarios, like uncovering the upper body with a randomized blanket state, where performance drops noticeably.} These results suggest that there remains room for future advances towards robust bedding manipulation that is adaptable to changes in human bodies and blanket states.

% We then evaluated how the policies trained with PPO for uncovering the manikin’s right lower leg and upper body generalized to these novel blanket configurations and human body sizes. Fig.~\ref{fig:generalize} visualizes the recall and precision averaged over 100 simulation rollouts for the original environments (see Table~\ref{table:ppo_eval_sim}) compared to environments with random blanket states or body shapes. For uncovering the upper body, this amount of human and blanket variation did not significantly impact performance, but variation in human body sizes did slightly lower precision. However, for uncovering the right lower leg, our models made substantially more errors in uncovering non-target points, resulting in noticeable decreases in precision. These results suggest that there remains room for future advances towards robust bedding manipulation that is adaptable to changes in human bodies and blanket states.

% \begin{figure}
% \centering
% \includegraphics[width=0.48\textwidth, trim={5cm 10.3cm 4.5cm 10.5cm}, clip]{images/generalization.pdf}
% \vspace{-0.6cm}
% \caption{\label{fig:generalize} Comparison of evaluation metrics for PPO policies to uncover the right lower leg and upper in the original simulation conditions (fixed initial blanket state and fixed body shape) and in the two generalization cases (random initial blanket state and random body shape).}
% \vspace{-0.5cm}
% \end{figure}

\subsection{Real World}
\label{sec:real_world}

\begin{figure}
\centering
\begingroup
\tiny
\begin{tabular}{cccccc}
 \hspace{-0.15cm} Uncovered & \hspace{0.15cm} Initial State  & \hspace{0.1cm} Final State& \hspace{0.4cm}  Uncovered & \hspace{0.15cm}  Initial State & \hspace{0.1cm}  Final State\\
\end{tabular}
\endgroup\\
\vspace{-0.3cm}
 %\rotatebox[x=-0cm,y=0.cm]{0}{Right Lower Leg}
 \includegraphics[width=0.17\textwidth, angle=-90, trim={1cm 1cm 0.6cm 2.8cm}, clip]{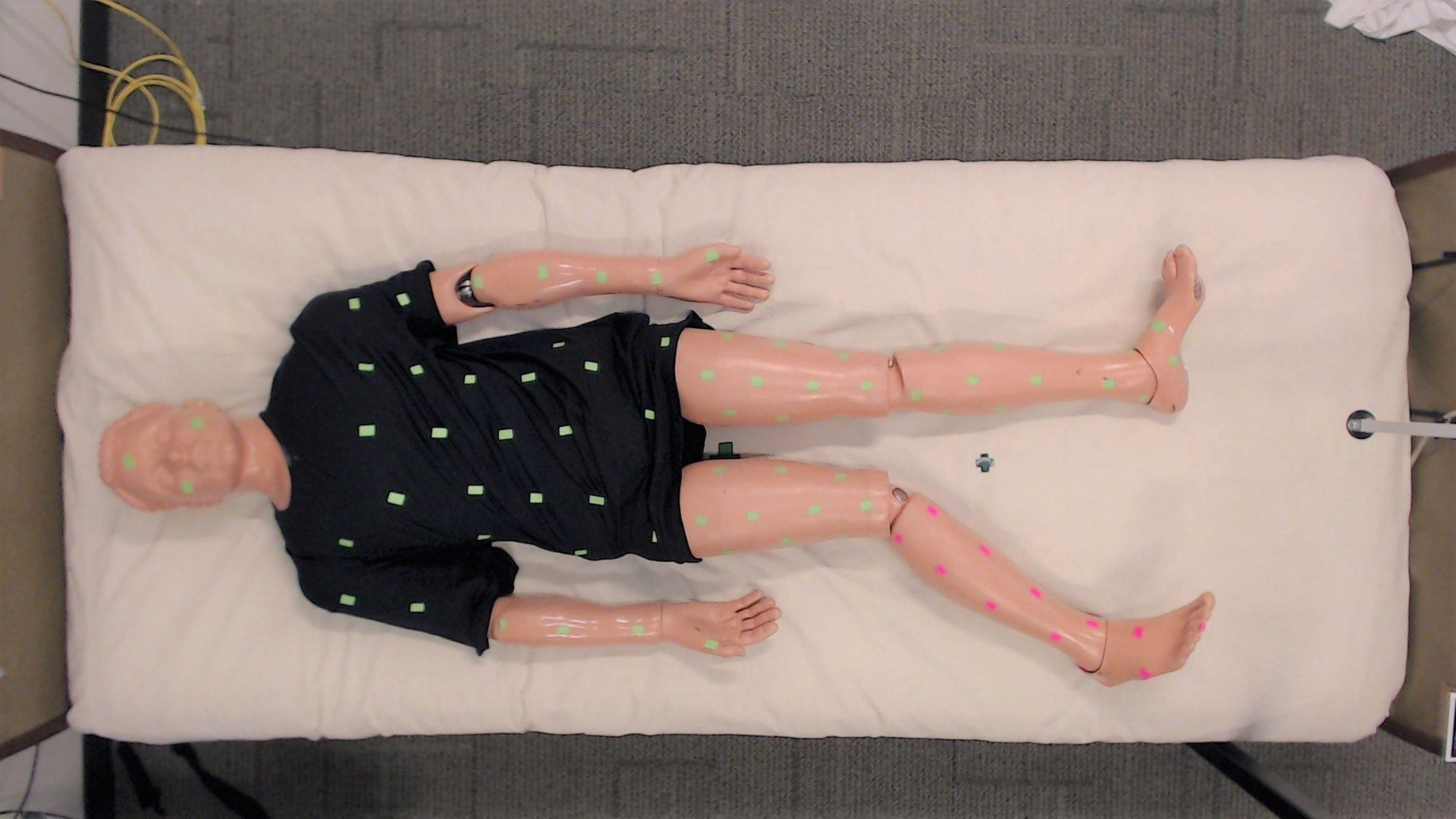}
 \hspace{-0.2 cm}
 \includegraphics[width=0.17\textwidth, angle=-90, trim={1cm 1cm 0.6cm 2.8cm}, clip]{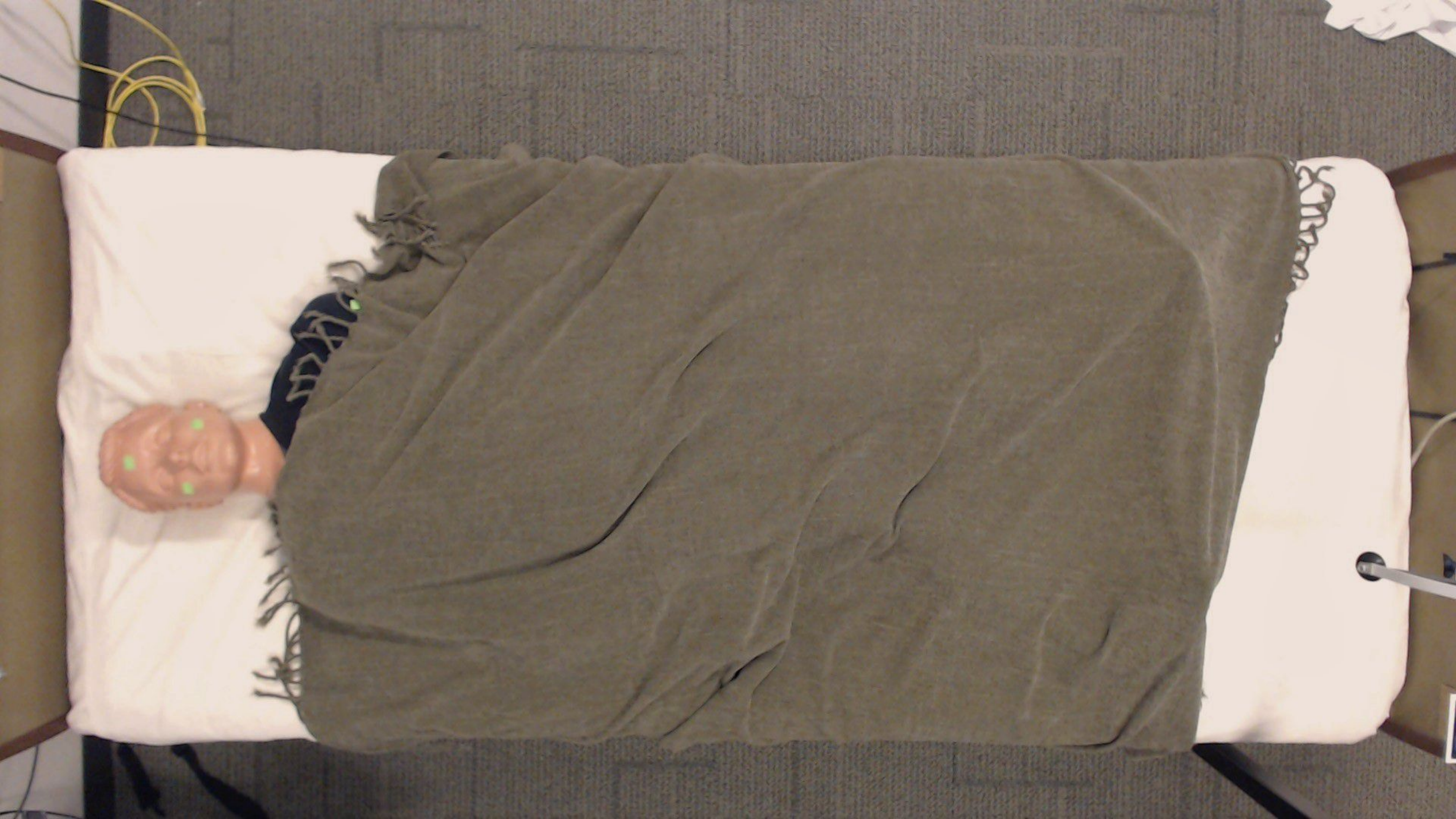}
 \hspace{-0.2 cm}
 \includegraphics[width=0.17\textwidth, angle=-90, trim={1cm 1cm 0.6cm 2.8cm}, clip]{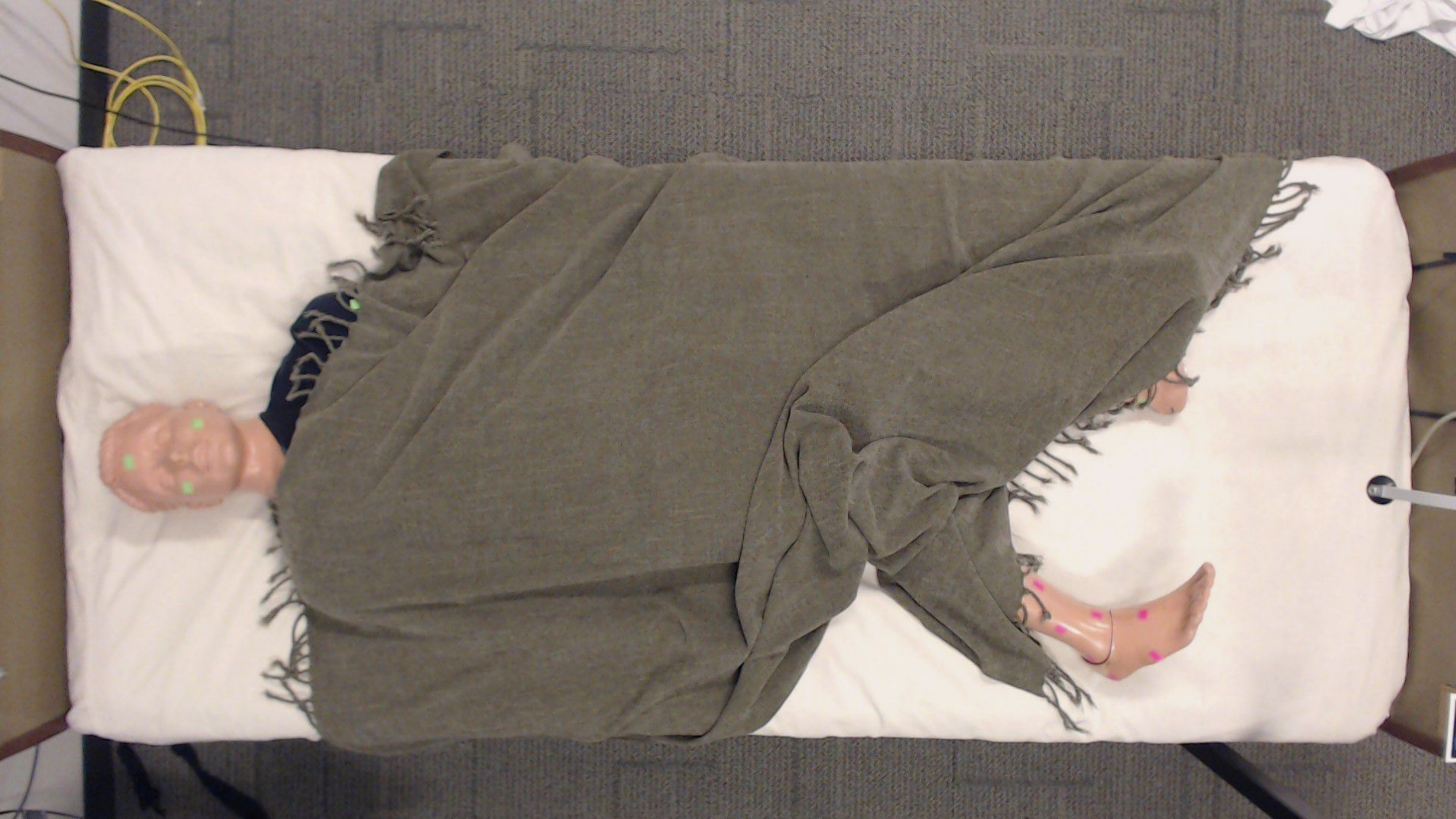}
 %\rotatebox[x=0cm,y=0cm]{90}{Left Arm}
 \hspace{0cm}
 \includegraphics[width=0.17\textwidth, angle=-90, trim={1cm 1cm 0.6cm 2.8cm}, clip]{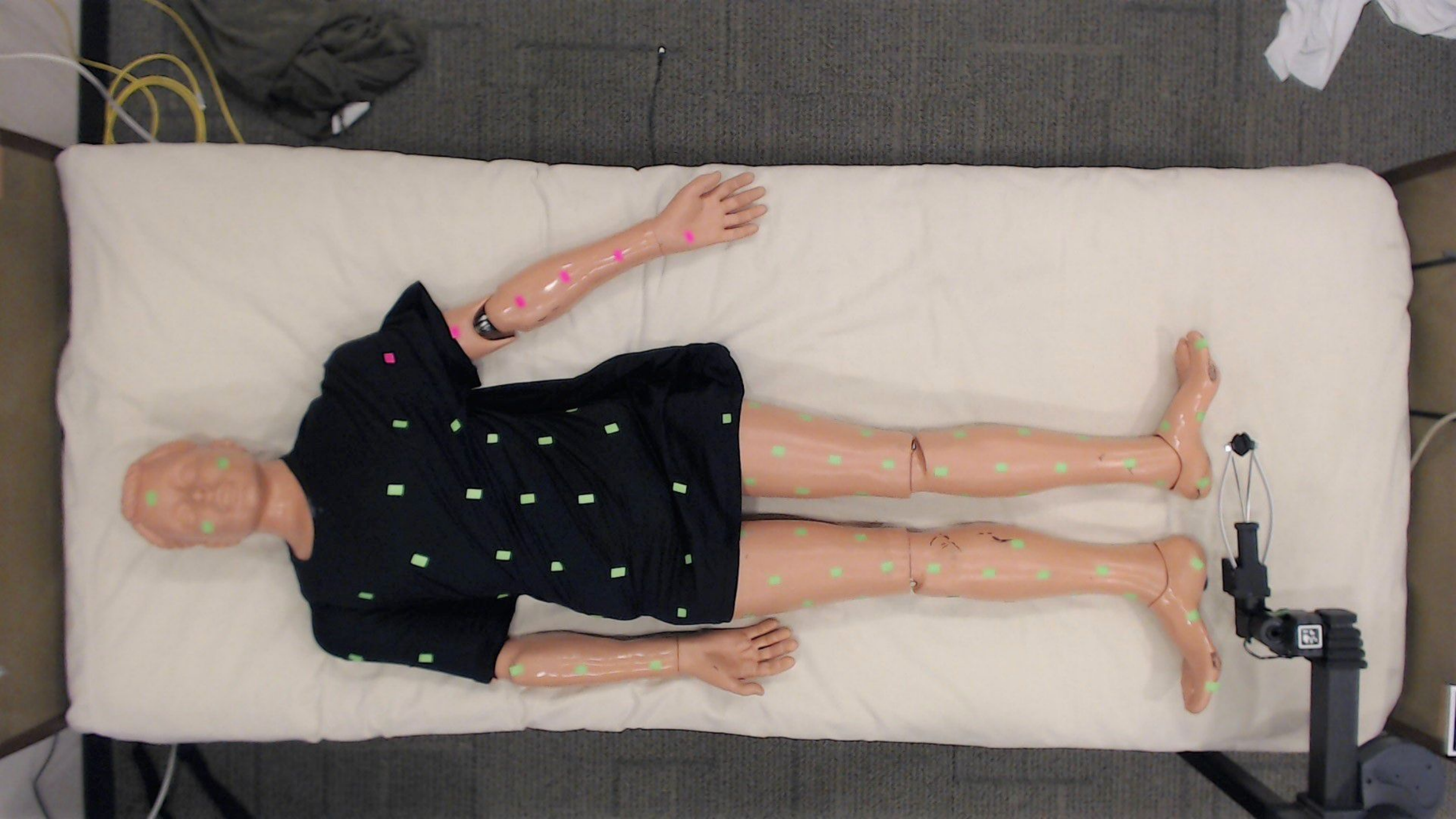}
 \hspace{-0.2 cm}
 \includegraphics[width=0.17\textwidth, angle=-90, trim={1cm 1cm 0.6cm 2.8cm}, clip]{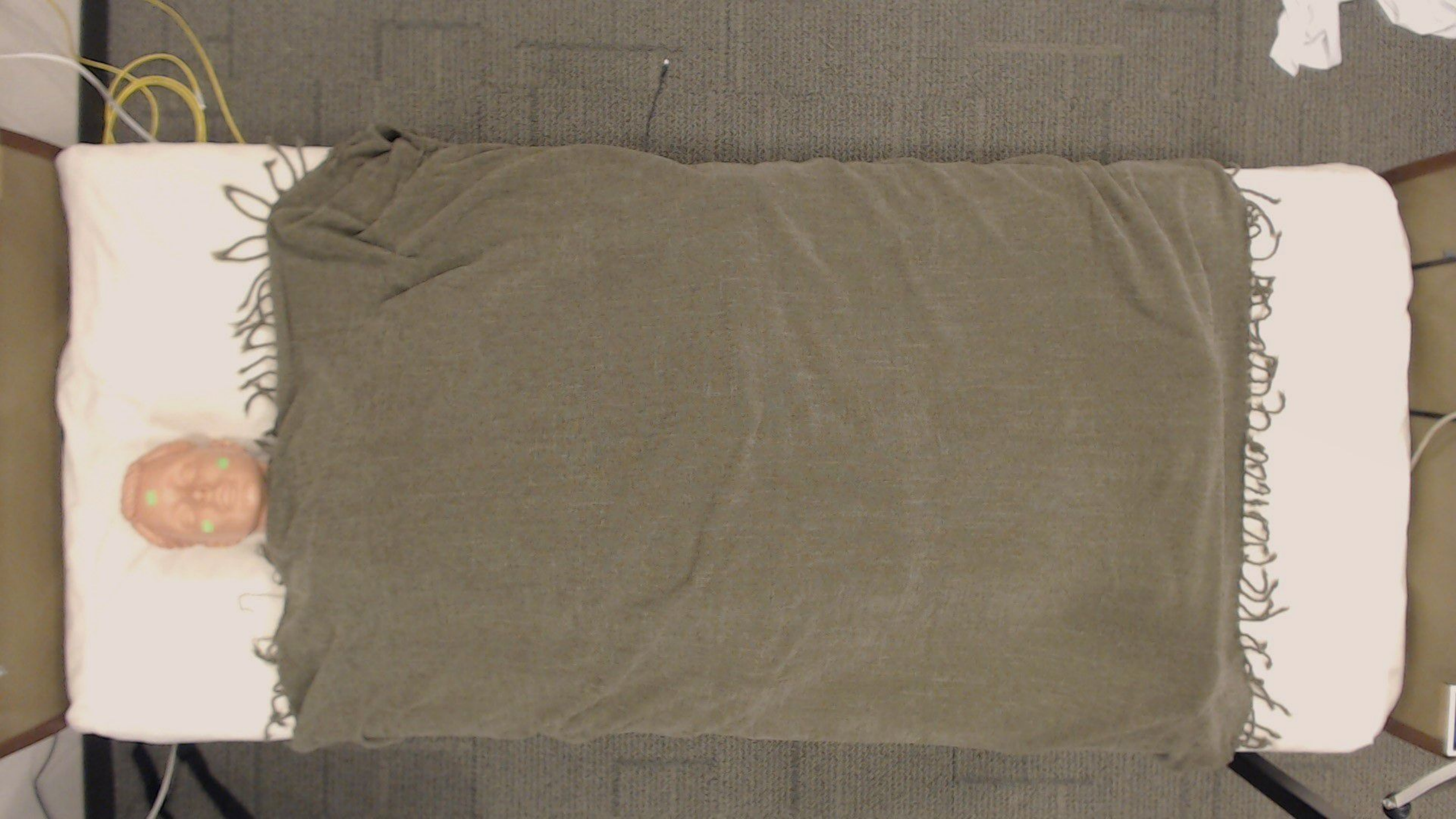}
 \hspace{-0.2 cm}
 \includegraphics[width=0.17\textwidth, angle=-90, trim={1cm 1cm 0.6cm 2.8cm}, clip]{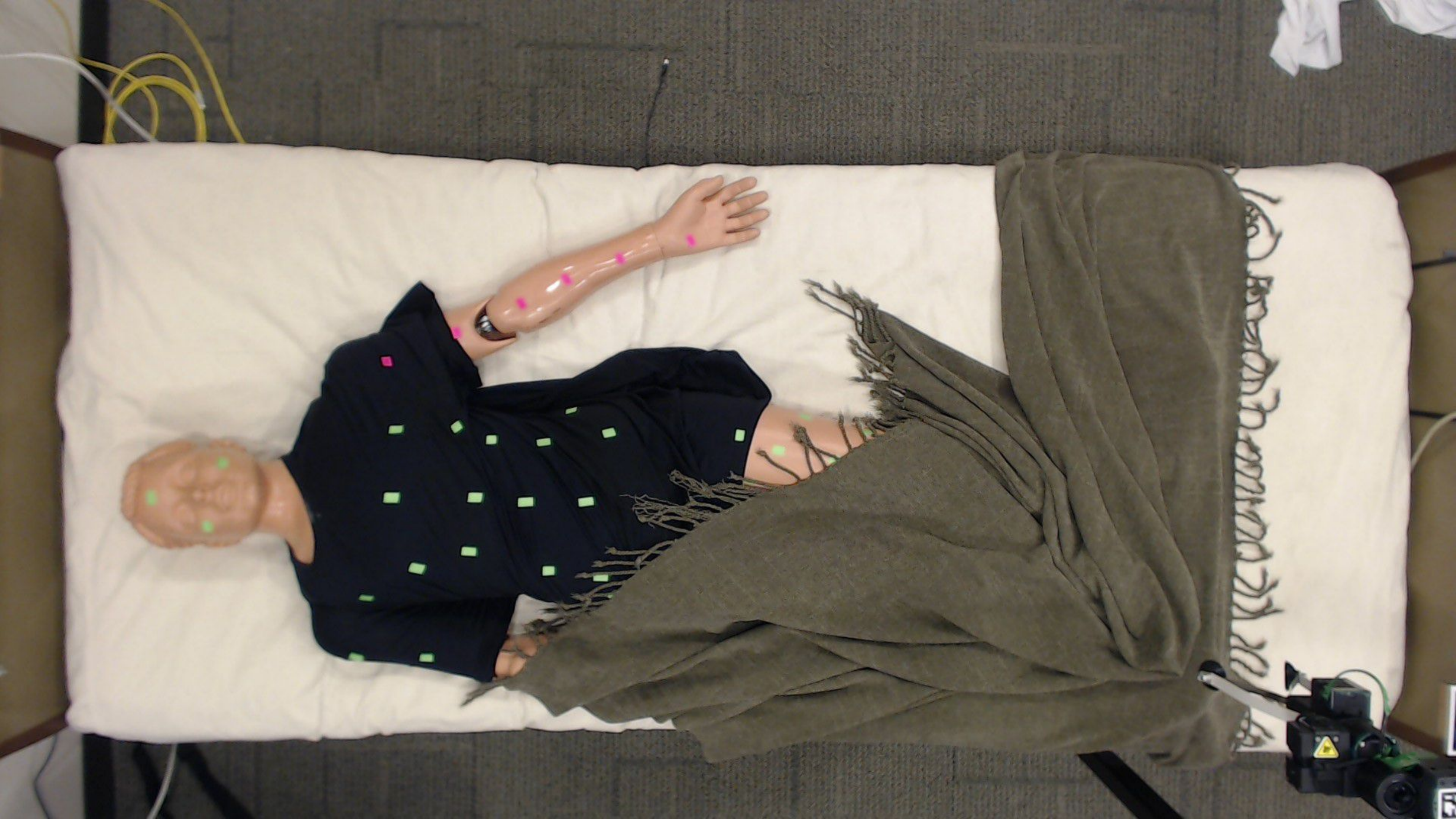} \\

\begingroup
\fontsize{7}{11}\selectfont
\begin{tabular}{cc}
 \hspace{-0.4cm} Right Lower Leg, $R(\bm{S}, \bm{a}) = 69.7$  & \hspace{0.8cm} Left Arm, $R(\bm{S}, \bm{a}) = -222$ \\
\end{tabular}
\endgroup\\
\vspace{-0.2cm}
 
\vspace{0.1cm}
 \includegraphics[width=0.17\textwidth, angle=-90, trim={1cm 1cm 0.6cm 2.8cm}, clip]{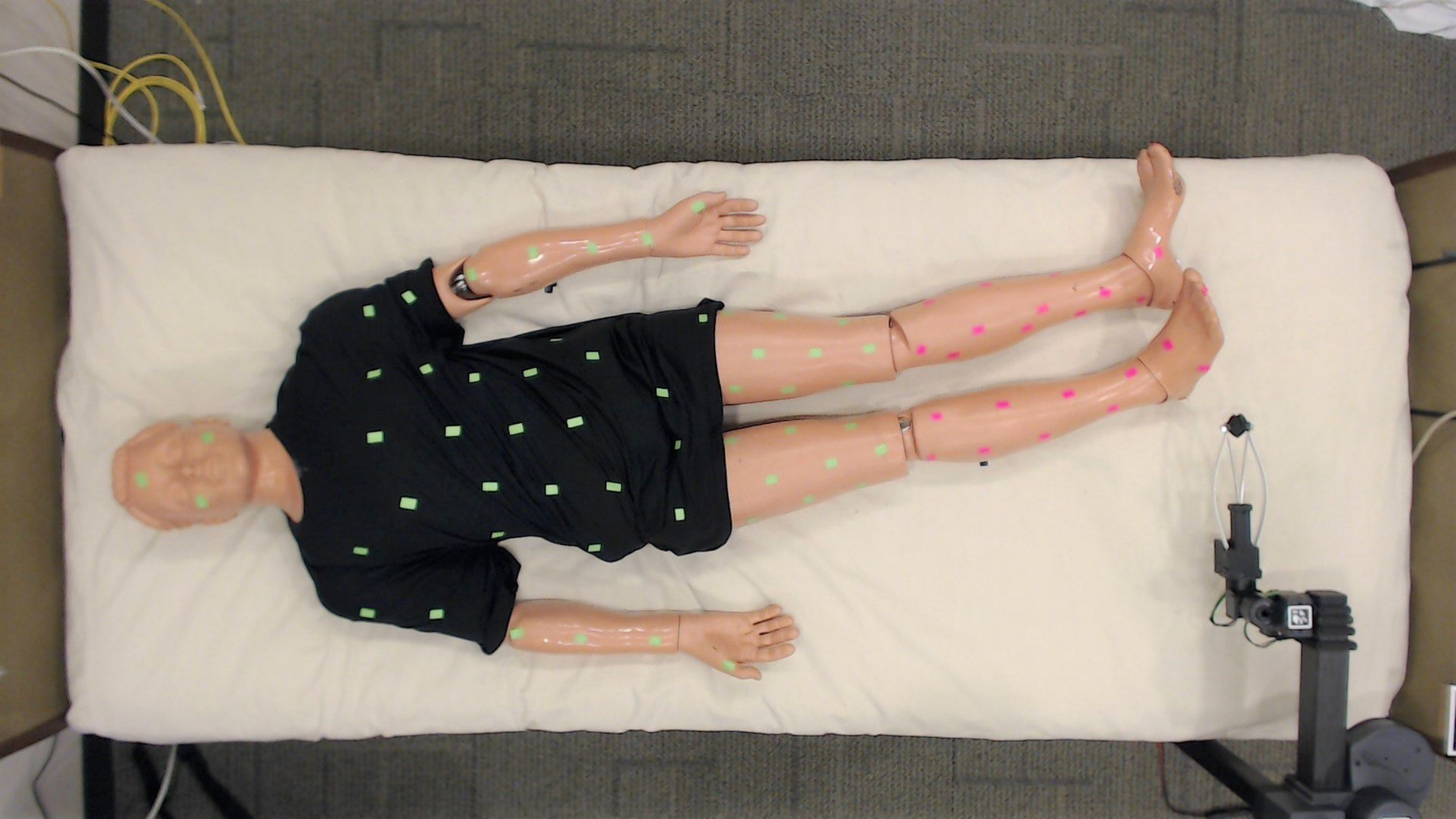}
 \hspace{-0.2 cm}
 \includegraphics[width=0.17\textwidth, angle=-90, trim={1cm 1cm 0.6cm 2.8cm}, clip]{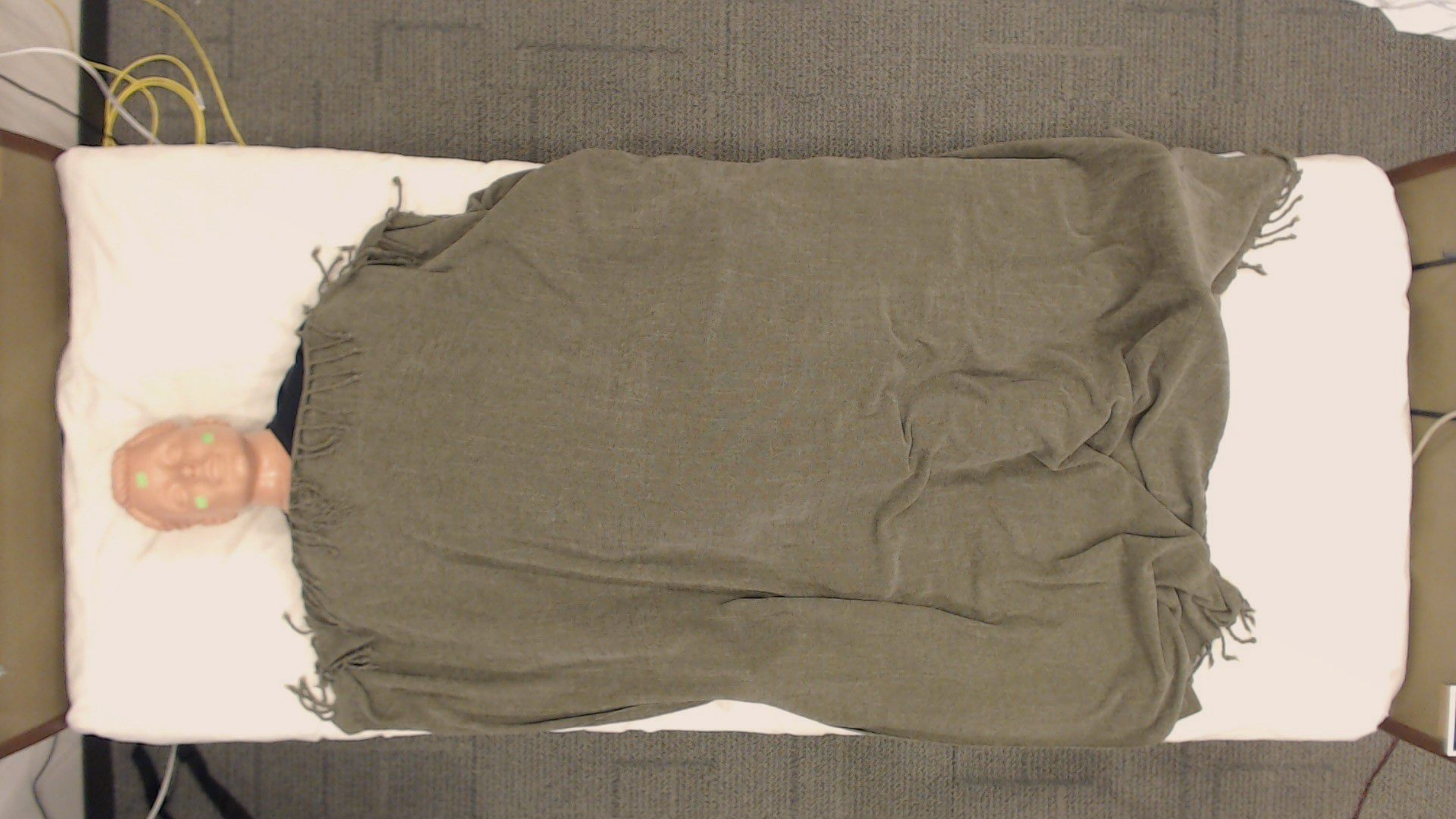}
 \hspace{-0.2 cm}
 \includegraphics[width=0.17 \textwidth, angle=-90, trim={1cm 1cm 0.6cm 2.8cm}, clip]{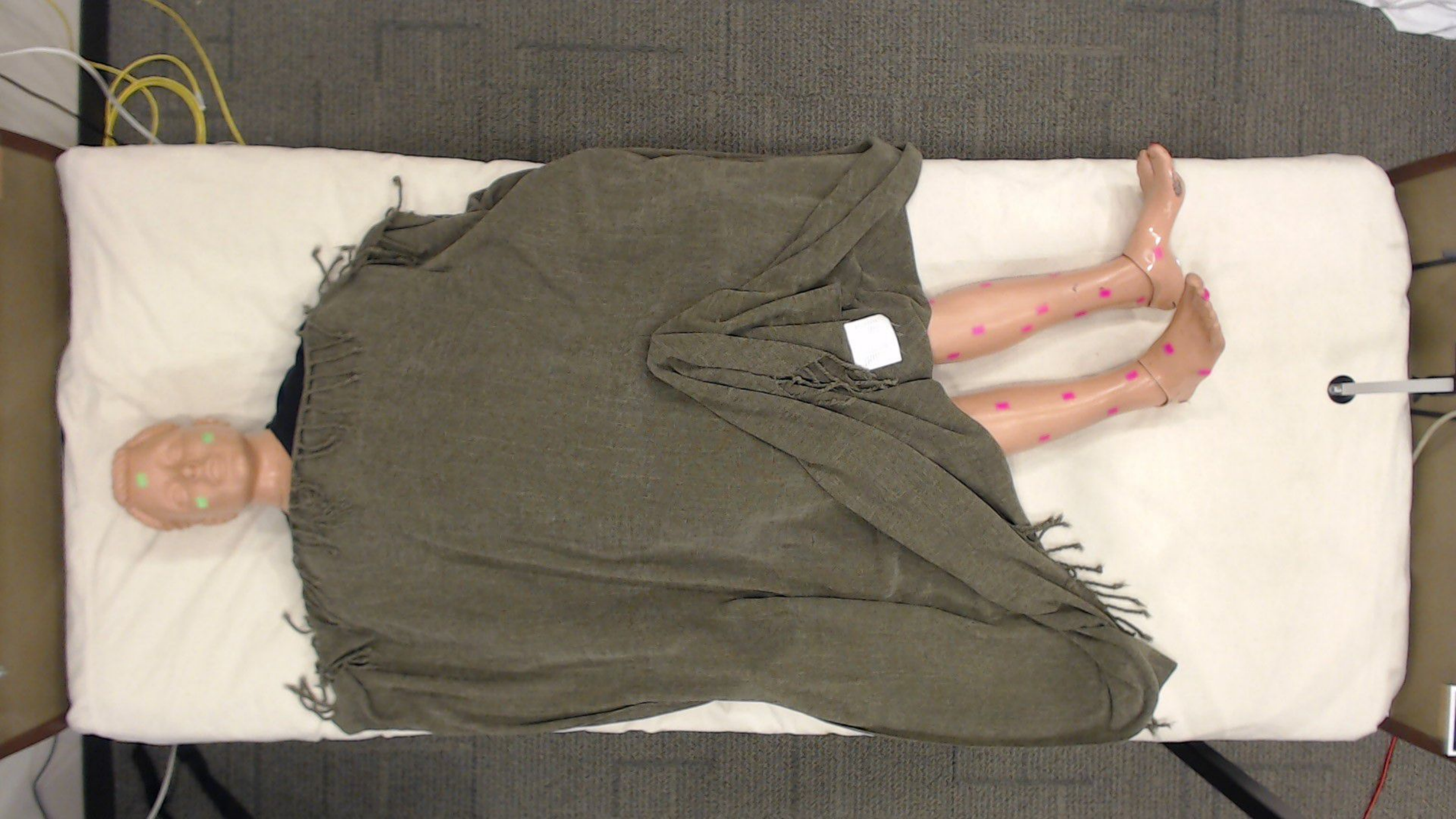}
 %\rotatebox[x=0cm,y=0.3cm]{90}{Upper Body}
 \hspace{0cm}
 \includegraphics[width=0.17\textwidth, angle=-90, trim={1cm 1cm 0.6cm 2.8cm}, clip]{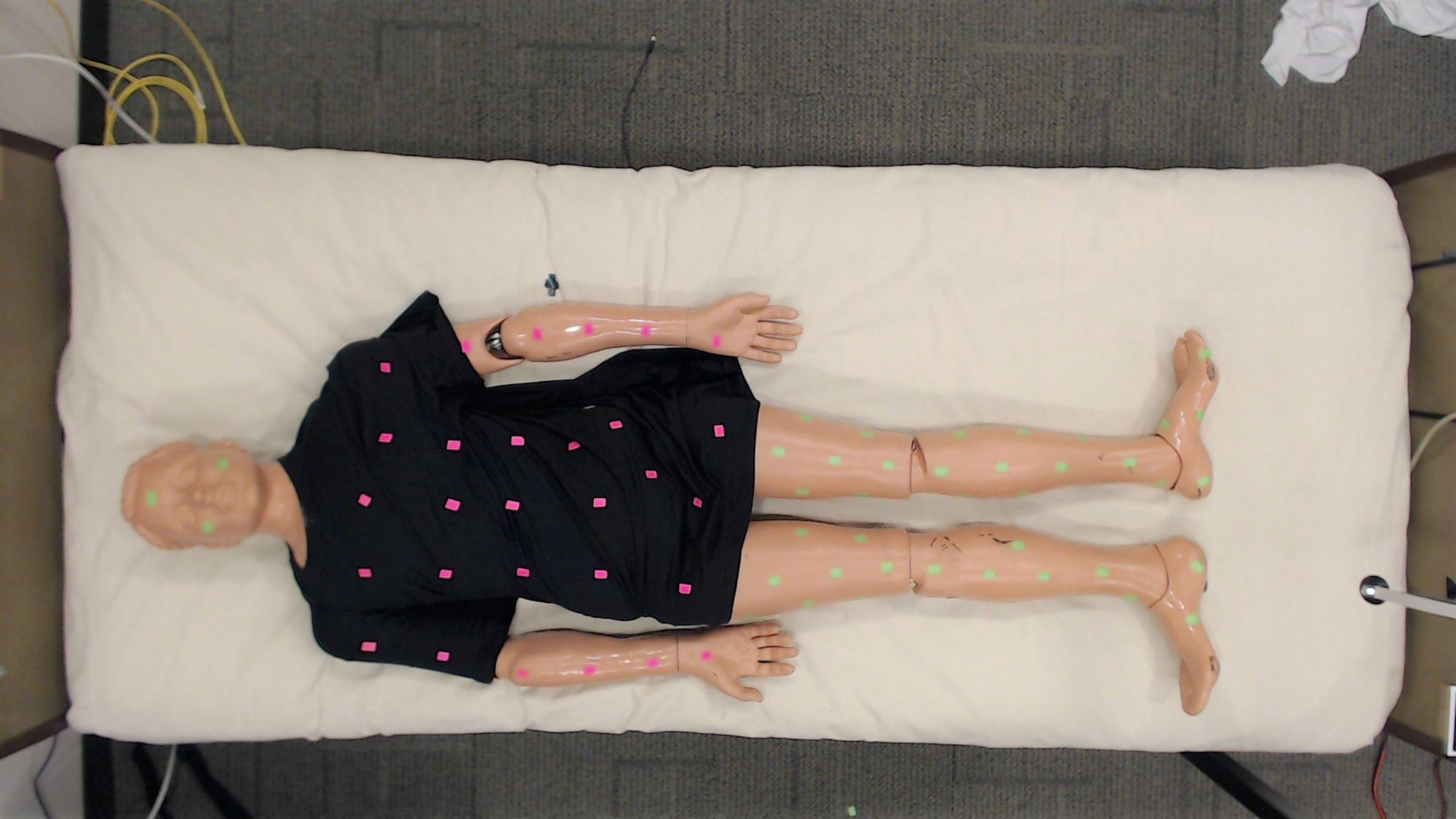}
 \hspace{-0.2 cm}
 \includegraphics[width=0.17\textwidth, angle=-90, trim={1cm 1cm 0.6cm 2.8cm}, clip]{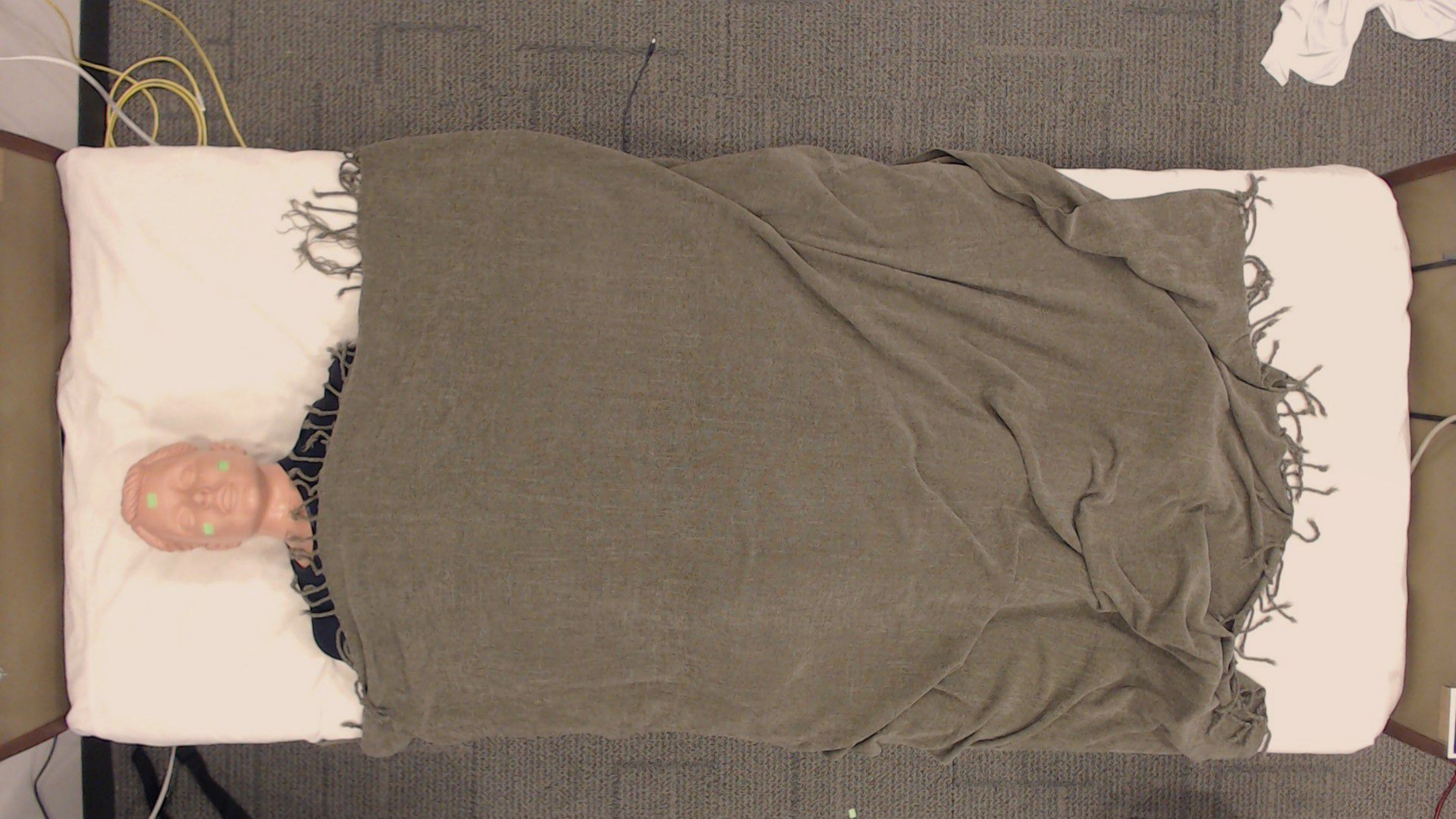}
 \hspace{-0.2 cm}
 \includegraphics[width=0.17\textwidth, angle=-90, trim={1cm 1cm 0.6cm 2.8cm}, clip]{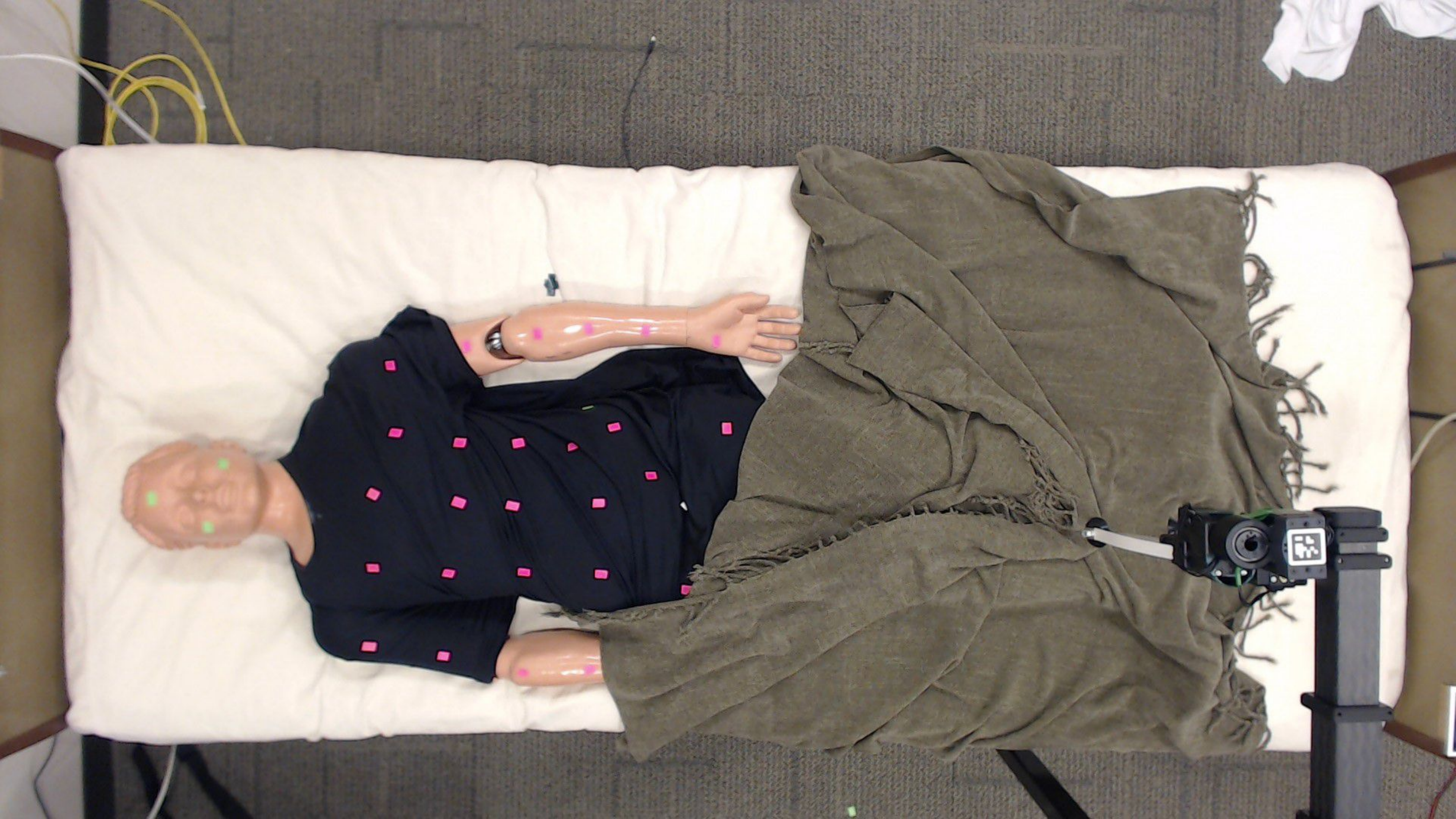} \\

\begingroup
\fontsize{7}{11}\selectfont
\begin{tabular}{cc}
 \hspace{-0.4cm} Both Lower Legs, $R(\bm{S}, \bm{a}) = 90.7$  & \hspace{0.6cm} Upper Body, $R(\bm{S}, \bm{a}) = 84.4$ \\
\end{tabular}
\endgroup\\
\vspace{-0.2cm}
 
\caption{\label{fig:full_page3} Images from evaluating PPO policies for uncovering the right lower leg, left arm, both lower legs, and upper body in the real world. From left to right, columns show the human pose and distribution of target and non-target markers on the body, the initial state of the blanket, and the markers exposed after the robot performs an action from the policy.}
\vspace{-0.2cm}
\end{figure}

We further evaluate this method for bedding manipulation on a real-world Stretch robot for uncovering target body parts of a medical manikin lying supine in a hospital bed. We setup the environment and experiments as discussed in Section~\ref{sec:real_world_setup}. Given the similarity in performance between policies trained with PPO versus with data from CMA-ES, we demonstrate only the PPO-trained policies in the real world. Specifically, we investigate how the policies trained with PPO performed for uncovering the manikin's right lower leg, leg arm, (both) lower legs, and upper body. Uncovering these target body parts represented a range of difficulties for policies in simulation, with uncovering the upper body being the most successful of the target body parts, and uncovering the left arm being the most challenging scenario. For each target body part to uncover, we evaluate the corresponding policy on a single manikin pose over three trials, where the manikin's pose remains constant across the three trials. In total, we perform 12 evaluation trials on the real robot, \new{three trials for each of the four attempted target body parts}. Fig.~\ref{fig:full_page3} shows the before and after results from evaluating our policies on a real robot (Stretch RE1) for uncovering each of the four target body parts. Table~\ref{table:ppo_eval_real} presents the average \new{F-score} and rewards resulting from the real-world trials. Demonstrations of both simulated and real-world bedding manipulation trials can be found in the supplementary video.

\begin{table}
% \vspace{0cm}
\centering
% \vspace{6pt}
\caption{\label{table:ppo_eval_real} Evaluation results for PPO policies on the real robot.}
\begin{tabular}{ccc} \toprule
    Target & \(F_{1}\)  & $\mu_{R}$ \\ \midrule\midrule
    Right Lower Leg & 0.83 & 69.7 \\
    Left Arm & 0.38 & -222 \\
    Lower Legs & 0.95 &  90.7 \\
    Upper Body & 0.92 & 84.4 \\
	\bottomrule
\end{tabular}
\vspace{-0.6cm}
\end{table}

% of the robot uncovering each of the four target body parts. Table~\ref{table:ppo_eval_real} presents the average recall, precision, and rewards that resulted from evaluating our policies on a real robot (Stretch RE1) for uncovering each of the four target body parts.

    \new{Overall, the F-score and rewards achieved in the real world for the given pose for each target limb are consistent with the metrics reported in Table~\ref{table:ppo+cmaes_eval_sim} where the PPO policies were evaluated in simulation. These results indicate that transfer of these simulation-trained policies to real-world robots for bedding manipulation around people may be viable. One noticeable difference between policy performance in simulation and in the real world was when attempting to uncover the left arm. For the given human pose, the real robot also uncovered the majority of the upper body, resulting in a lower mean reward. This failure case in the real world is, however, consistent with results from similar poses in simulation.}
    
    % \new{Overall, the F-score and rewards achieved in the real world are consistent with the metrics reported in Table~\ref{table:ppo+cmaes_eval_sim} where the PPO policies were evaluated in simulation. As a result, transfer of these simulation-trained policies to real-world robots for bedding manipulation around people remains promising. One noticeable difference between policy performance in simulation and in the real world was when attempting to uncover the left arm. For the given human pose, the real robot also uncovered the majority of the upper body, resulting in a lower mean reward. This failure case in the real world is, however, consistent with those observed in simulation.}

% \begin{table}
% \vspace{0cm}
% \centering
% % \vspace{6pt}
% \caption{\label{table:ppo_eval_real} Evaluation results for PPO policies on the real robot.}
% \begin{tabular}{cccc} \toprule
%     Target & Recall & Precision & $\mu_{R}$ \\ \midrule\midrule
%     Right Lower Leg & 0.73 & 0.96 & 69.7 \\
%     Left Arm & 1.00 & 0.24 & -222 \\
%     Lower Legs & 0.97 & 0.93 & 90.7 \\
%     Upper Body & 1.00 & 0.84 & 84.4 \\
% 	\bottomrule
% \end{tabular}
% \vspace{-0.6cm}
% \end{table}

\section{Conclusion}

We introduced a formulation for robotic bedding manipulation around people that aims to uncover a blanket from only target body parts while keeping the rest of the body covered. We presented and compared both reinforcement learning and self-supervised learning approaches to this task. For each approach, we trained policies in simulation to uncover one of six target body parts. Through an evaluation in simulation, we found that these policies were able to uncover cloth from target body parts with reasonable performance across several metrics. We also investigated how well these policies generalize to scenarios outside of the training distribution with novel human body sizes and blanket configurations. Lastly, we demonstrated these simulation-trained policies on a real-world robot for manipulating and uncovering a blanket from select body parts of a manikin lying in bed.

% \section*{Acknowledgment}

% \small
% \textit{Text} 

\bibliographystyle{IEEEtran}
\bibliography{bibliography}

\begin{thebibliography}{10}
\providecommand{\url}[1]{#1}
\csname url@rmstyle\endcsname
\providecommand{\newblock}{\relax}
\providecommand{\bibinfo}[2]{#2}
\providecommand\BIBentrySTDinterwordspacing{\spaceskip=0pt\relax}
\providecommand\BIBentryALTinterwordstretchfactor{4}
\providecommand\BIBentryALTinterwordspacing{\spaceskip=\fontdimen2\font plus
\BIBentryALTinterwordstretchfactor\fontdimen3\font minus
  \fontdimen4\font\relax}
\providecommand\BIBforeignlanguage[2]{{%
\expandafter\ifx\csname l@#1\endcsname\relax
\typeout{** WARNING: IEEEtran.bst: No hyphenation pattern has been}%
\typeout{** loaded for the language `#1'. Using the pattern for}%
\typeout{** the default language instead.}%
\else
\language=\csname l@#1\endcsname
\fi
#2}}

\bibitem{lynn2011bedbath}
P.~Lynn and C.~Lynn, \emph{Hygiene: Giving a Bed Bath}, 3rd~ed.\hskip 1em plus
  0.5em minus 0.4em\relax Wolters Kluwer/Lippincott Williams \& Wilkins Health,
  2011, p. 316–325.

\bibitem{tanaka2018emd}
D.~Tanaka, S.~Arnold, and K.~Yamazaki, ``Emd net: An
  encode–manipulate–decode network for cloth manipulation,'' \emph{IEEE
  Robotics and Automation Letters}, vol.~3, no.~3, pp. 1771--1778, 2018.

\bibitem{ganapathi2021learning}
A.~Ganapathi, P.~Sundaresan, B.~Thananjeyan, A.~Balakrishna, D.~Seita,
  J.~Grannen, M.~Hwang, R.~Hoque, J.~E. Gonzalez, N.~Jamali, \emph{et~al.},
  ``Learning dense visual correspondences in simulation to smooth and fold real
  fabrics,'' in \emph{2021 IEEE International Conference on Robotics and
  Automation (ICRA)}.\hskip 1em plus 0.5em minus 0.4em\relax IEEE, 2021, pp.
  11\,515--11\,522.

\bibitem{tsurumine2019drl}
Y.~Tsurumine, Y.~Cui, E.~Uchibe, and T.~Matsubara, ``Deep reinforcement
  learning with smooth policy update: Application to robotic cloth
  manipulation,'' \emph{Robotics and Autonomous Systems}, 2019.

\bibitem{sun2015wrinkles}
L.~Sun, G.~Aragon-Camarasa, S.~Rogers, and J.~P. Siebert, ``Accurate garment
  surface analysis using an active stereo robot head with application to
  dual-arm flattening,'' in \emph{2015 IEEE International Conference on
  Robotics and Automation (ICRA)}, 2015, pp. 185--192.

\bibitem{ramisa2012wrinkles}
A.~Ramisa, G.~Alenyà, F.~Moreno-Noguer, and C.~Torras, ``Using depth and
  appearance features for informed robot grasping of highly wrinkled clothes,''
  in \emph{2012 IEEE International Conference on Robotics and Automation},
  2012, pp. 1703--1708.

\bibitem{yamazaki2014hems}
K.~Yamazaki, ``Grasping point selection on an item of crumpled clothing based
  on relational shape description,'' in \emph{2014 IEEE/RSJ International
  Conference on Intelligent Robots and Systems}, 2014.

\bibitem{seita2019bedmaking}
D.~Seita, N.~Jamali, M.~Laskey, A.~K. Tanwani, R.~Berenstein, P.~Baskaran,
  S.~Iba, J.~Canny, and K.~Goldberg, ``{Deep Transfer Learning of Pick Points
  on Fabric for Robot Bed-Making},'' in \emph{International Symposium on
  Robotics Research (ISRR)}, 2019.

\bibitem{yuba2017stateest}
H.~Yuba, S.~Arnold, and K.~Yamazaki, ``Unfolding of a rectangular cloth from
  unarranged starting shapes by a dual-armed robot with a mechanism for
  managing recognition error and uncertainty,'' \emph{Advanced Robotics},
  vol.~31, no.~10, pp. 544--556, 2017.

\bibitem{qian2020segmentation}
J.~Qian, T.~Weng, L.~Zhang, B.~Okorn, and D.~Held, ``Cloth region segmentation
  for robust grasp selection,'' \emph{2020 IEEE/RSJ International Conference on
  Intelligent Robots and Systems (IROS)}, Oct 2020.

\bibitem{shepard2010towel}
J.~Maitin-Shepard, M.~Cusumano-Towner, J.~Lei, and P.~Abbeel, ``Cloth grasp
  point detection based on multiple-view geometric cues with application to
  robotic towel folding,'' in \emph{2010 IEEE International Conference on
  Robotics and Automation}, 2010, pp. 2308--2315.

\bibitem{matsubara2013topology}
T.~Matsubara, D.~Shinohara, and M.~Kidode, ``Reinforcement learning of a motor
  skill for wearing a t-shirt using topology coordinates,'' \emph{Advanced
  Robotics}, vol.~27, no.~7, pp. 513--524, 2013.

\bibitem{hoque2021visuospatial}
R.~Hoque, D.~Seita, A.~Balakrishna, A.~Ganapathi, A.~K. Tanwani, N.~Jamali,
  K.~Yamane, S.~Iba, and K.~Goldberg, ``Visuospatial foresight for multi-step,
  multi-task fabric manipulation,'' \emph{arXiv preprint arXiv:2003.09044},
  2020.

\bibitem{garcia2020benchmarking}
I.~Garcia-Camacho, M.~Lippi, M.~C. Welle, H.~Yin, R.~Antonova, A.~Varava,
  J.~Borras, C.~Torras, A.~Marino, G.~Alenya, \emph{et~al.}, ``Benchmarking
  bimanual cloth manipulation,'' \emph{IEEE Robotics and Automation Letters},
  vol.~5, no.~2, pp. 1111--1118, 2020.

\bibitem{ebert2018visual}
F.~Ebert, C.~Finn, S.~Dasari, A.~Xie, A.~Lee, and S.~Levine, ``Visual
  foresight: Model-based deep reinforcement learning for vision-based robotic
  control,'' \emph{arXiv preprint arXiv:1812.00568}, 2018.

\bibitem{lin2020softgym}
X.~Lin, Y.~Wang, J.~Olkin, and D.~Held, ``Softgym: Benchmarking deep
  reinforcement learning for deformable object manipulation,'' \emph{arXiv
  preprint arXiv:2011.07215}, 2020.

\bibitem{matas2018sim}
J.~Matas, S.~James, and A.~J. Davison, ``Sim-to-real reinforcement learning for
  deformable object manipulation,'' in \emph{Conference on Robot
  Learning}.\hskip 1em plus 0.5em minus 0.4em\relax PMLR, 2018, pp. 734--743.

\bibitem{wu2020learning}
Y.~Wu, W.~Yan, T.~Kurutach, L.~Pinto, and P.~Abbeel, ``Learning to manipulate
  deformable objects without demonstrations,'' \emph{arXiv preprint
  arXiv:1910.13439}, 2019.

\bibitem{jimenez2020perception}
P.~Jiménez and C.~Torras, ``Perception of cloth in assistive robotic
  manipulation tasks,'' \emph{Natural Computing}, vol.~19, no.~2, 2020.

\bibitem{koganti2015cloth}
N.~Koganti, J.~G. Ngeo, T.~Tomoya, K.~Ikeda, and T.~Shibata, ``Cloth dynamics
  modeling in latent spaces and its application to robotic clothing
  assistance,'' in \emph{IEEE/RSJ IROS}, 2015, pp. 3464--3469.

\bibitem{twardon2018learning}
L.~Twardon and H.~Ritter, ``Learning to put on a knit cap in a head-centric
  policy space,'' \emph{IEEE Robotics and Automation Letters}, 2018.

\bibitem{yamazaki2014bottom}
K.~Yamazaki, R.~Oya, K.~Nagahama, K.~Okada, and M.~Inaba, ``Bottom dressing by
  a life-sized humanoid robot provided failure detection and recovery
  functions,'' in \emph{2014 IEEE/SICE International Symposium on System
  Integration}, 2014, pp. 564--570.

\bibitem{canal2018joining}
G.~Canal, E.~Pignat, G.~Alenyà, S.~Calinon, and C.~Torras, ``Joining
  high-level symbolic planning with low-level motion primitives in adaptive
  hri: Application to dressing assistance,'' in \emph{IEEE ICRA}, 2018.

\bibitem{pignat2017learning}
E.~Pignat and S.~Calinon, ``Learning adaptive dressing assistance from human
  demonstration,'' \emph{Robotics and Autonomous Systems}, 2017.

\bibitem{gao2016iterative}
Y.~Gao, H.~J. Chang, and Y.~Demiris, ``Iterative path optimisation for
  personalised dressing assistance using vision and force information,'' in
  \emph{International Conference on Intelligent Robots and Systems}, 2016.

\bibitem{erickson2018hapticpredictive}
Z.~Erickson, H.~M. Clever, G.~Turk, C.~K. Liu, and C.~C. Kemp, ``Deep haptic
  model predictive control for robot-assisted dressing,'' \emph{2018 IEEE
  International Conference on Robotics and Automation (ICRA)}, 2018.

\bibitem{zhang2017personalized}
F.~Zhang, A.~Cully, and Y.~Demiris, ``Personalized robot-assisted dressing
  using user modeling in latent spaces,'' in \emph{IEEE/RSJ International
  Conference on Intelligent Robots and Systems (IROS)}, 2017.

\bibitem{kapusta2016data}
A.~Kapusta, W.~Yu, T.~Bhattacharjee, C.~K. Liu, G.~Turk, and C.~C. Kemp,
  ``Data-driven haptic perception for robot-assisted dressing,'' in \emph{2016
  25th IEEE International Symposium on Robot and Human Interactive
  Communication (RO-MAN)}, 2016, pp. 451--458.

\bibitem{yu2017haptic}
W.~Yu, A.~Kapusta, J.~Tan, C.~C. Kemp, G.~Turk, and C.~K. Liu, ``Haptic
  simulation for robot-assisted dressing,'' in \emph{2017 IEEE International
  Conference on Robotics and Automation (ICRA)}, 2017.

\bibitem{erickson2019multidimensional}
Z.~Erickson, H.~M. Clever, V.~Gangaram, G.~Turk, C.~K. Liu, and C.~C. Kemp,
  ``Multidimensional capacitive sensing for robot-assisted dressing and
  bathing,'' in \emph{2019 IEEE 16th International Conference on Rehabilitation
  Robotics (ICORR)}.\hskip 1em plus 0.5em minus 0.4em\relax IEEE, 2019, pp.
  224--231.

\bibitem{kapusta2019personalized}
A.~Kapusta, Z.~Erickson, H.~M. Clever, W.~Yu, C.~K. Liu, G.~Turk, and C.~C.
  Kemp, ``Personalized collaborative plans for robot-assisted dressing via
  optimization and simulation,'' \emph{Autonomous Robots}, 2019.

\bibitem{zhang2020learningGP}
F.~Zhang and Y.~Demiris, ``Learning grasping points for garment manipulation in
  robot-assisted dressing,'' \emph{2020 IEEE International Conference on
  Robotics and Automation (ICRA)}, 2020.

\bibitem{saxena2019garment}
K.~Saxena and T.~Shibata, ``Garment recognition and grasping point detection
  for clothing assistance task using deep learning*,'' in \emph{IEEE/SICE
  International Symposium on System Integration}, 2019.

\bibitem{king2010bedbath}
C.-H. King, T.~L. Chen, A.~Jain, and C.~C. Kemp, ``Towards an assistive robot
  that autonomously performs bed baths for patient hygiene,'' in
  \emph{International Conference on Intelligent Robots and Systems}, 2010.

\bibitem{kapusta2019system}
A.~S. Kapusta, P.~M. Grice, H.~M. Clever, Y.~Chitalia, D.~Park, and C.~C. Kemp,
  ``A system for bedside assistance that integrates a robotic bed and a mobile
  manipulator,'' \emph{Plos one}, vol.~14, no.~10, 2019.

\bibitem{erickson2020assistivegym}
Z.~Erickson, V.~Gangaram, A.~Kapusta, C.~K. Liu, and C.~C. Kemp, ``Assistive
  gym: A physics simulation framework for assistive robotics,'' in \emph{2020
  IEEE International Conference on Robotics and Automation (ICRA)}.\hskip 1em
  plus 0.5em minus 0.4em\relax IEEE, 2020, pp. 10\,169--10\,176.

\bibitem{schulman2017proximal}
J.~Schulman, F.~Wolski, P.~Dhariwal, A.~Radford, and O.~Klimov, ``Proximal
  policy optimization algorithms,'' \emph{arXiv:1707.06347}, 2017.

\bibitem{hansen2003reducing}
N.~Hansen, S.~D. M{\"u}ller, and P.~Koumoutsakos, ``Reducing the time
  complexity of the derandomized evolution strategy with covariance matrix
  adaptation (cma-es),'' \emph{Evolutionary computation}, 2003.

\bibitem{hansen2019pycma}
N.~Hansen, Y.~Akimoto, and P.~Baudis, ``{CMA-ES/pycma} on {G}ithub,'' Zenodo,
  DOI:10.5281/zenodo.2559634, Feb. 2019.

\bibitem{clever2018pressurepose}
H.~M. Clever, A.~Kapusta, D.~Park, Z.~Erickson, Y.~Chitalia, and C.~C. Kemp,
  ``3d human pose estimation on a configurable bed from a pressure image,'' in
  \emph{2018 IEEE/RSJ International Conference on Intelligent Robots and
  Systems (IROS)}, 2018, pp. 54--61.

\bibitem{clever2021depthpose}
H.~M. Clever, P.~Grady, G.~Turk, and C.~C. Kemp, ``Bodypressure--inferring body
  pose and contact pressure from a depth image,'' \emph{arXiv preprint
  arXiv:2105.09936}, 2021.

\bibitem{hoque2021visuospatial2}
R.~Hoque, D.~Seita, A.~Balakrishna, A.~Ganapathi, A.~K. Tanwani, N.~Jamali,
  K.~Yamane, S.~Iba, and K.~Goldberg, ``Visuospatial foresight for physical
  sequential fabric manipulation,'' \emph{arXiv preprint arXiv:2102.09754},
  2021.

\bibitem{pavlakos2019expressive}
G.~Pavlakos, V.~Choutas, N.~Ghorbani, T.~Bolkart, A.~A. Osman, D.~Tzionas, and
  M.~J. Black, ``Expressive body capture: 3d hands, face, and body from a
  single image,'' in \emph{Proceedings of the IEEE/CVF Conference on Computer
  Vision and Pattern Recognition}, 2019.

\end{thebibliography}

% \balance

\end{document}